\RequirePackage{xcolor}
\documentclass[journal,twoside,web]{ieeecolor}
\usepackage[pdftex]{graphicx}
\usepackage{generic}
\usepackage{cite}
\usepackage{amssymb,amsfonts}
\usepackage{cite}
\usepackage{amsmath,amssymb,amsfonts}
\usepackage[utf8]{inputenc}
\usepackage{changes}
\usepackage{mathtools}
\usepackage{hyperref}
\usepackage{amssymb}
\usepackage{bm}
\usepackage{graphicx}
\usepackage{float}
\usepackage{amsmath}
\usepackage{tikz}
\usepackage{lmodern}
\usepackage{bbm}
\usepackage{multirow}
\usetikzlibrary{positioning, backgrounds, fit, shapes.arrows}
\usetikzlibrary{decorations.markings}
\usepackage{cuted}
\let\labelindent\relax
\usepackage{enumitem}
\usepackage{caption}
\usepackage{subcaption}

\newtheorem{theorem}{Theorem}[section]
\newtheorem{corollary}{Corollary}[theorem]
\newtheorem{lemma}[theorem]{Lemma}

\newtheorem{problem}{Problem}
\newtheorem{assumption}{Assumption}[section]

\usepackage{algorithm}
\usepackage{algpseudocode}
\tikzset{block/.style = {draw, fill=white, rectangle,
		minimum height=3em, minimum width=2cm},
	input/.style = {coordinate},
	output/.style = {coordinate},
	pinstyle/.style = {pin edge={to-,t,black}}
	radiation/.style={{decorate,decoration={expanding waves,angle=90,segment   length=4pt}}}
	
}
\usepackage{smartdiagram}

\tikzstyle{block} = [draw, rectangle, minimum height=2em, minimum width=2em]
\tikzstyle{sum} = [draw, circle,minimum width=0.1 cm]
\tikzstyle{input} = [coordinate]
\tikzstyle{output} = [coordinate]
\tikzstyle{dummy} = [coordinate]
\tikzstyle{pinstyle} = [pin edge={to-,thin,black}]
\usetikzlibrary{positioning, fit, arrows.meta}
\usetikzlibrary{positioning}
\usetikzlibrary{shapes,arrows}
\tikzstyle{frame_cyan} = [thick, draw=blue, solid,inner sep=0.3em]
\tikzstyle{frame_red} = [thick, draw=red, solid,inner sep=0.3em]
\tikzstyle{frame_green} = [thick, draw=green, solid,inner sep=0.3em]

\usepackage{tikz}
\usepackage{xcolor}
\definecolor{fc}{HTML}{1E90FF}
\definecolor{h}{HTML}{228B22}
\definecolor{bias}{HTML}{87CEFA}
\definecolor{noise}{HTML}{8B008B}
\definecolor{conv}{HTML}{FFA500}
\definecolor{pool}{HTML}{B22222}
\definecolor{up}{HTML}{B22222}
\definecolor{view}{HTML}{FFFFFF}
\definecolor{bn}{HTML}{FFD700}
\tikzset{fc/.style={black,draw=black,fill=fc,rectangle,minimum height=1cm}}
\tikzset{h/.style={black,draw=black,fill=h,rectangle,minimum height=1cm}}
\tikzset{bias/.style={black,draw=black,fill=bias,rectangle,minimum height=1cm}}
\tikzset{noise/.style={black,draw=black,fill=noise,rectangle,minimum height=1cm}}
\tikzset{conv/.style={black,draw=black,fill=conv,rectangle,minimum height=1cm}}
\tikzset{pool/.style={black,draw=black,fill=pool,rectangle,minimum height=1cm}}
\tikzset{up/.style={black,draw=black,fill=up,rectangle,minimum height=1cm}}
\tikzset{view/.style={black,draw=black,fill=view,rectangle,minimum height=1cm}}
\tikzset{bn/.style={black,draw=black,fill=bn,rectangle,minimum height=1cm}}
 
\usepackage{xspace}

\tikzstyle{dummy} = [coordinate]
\pgfkeys{/pgf/.cd,
  parallelepiped offset x/.initial=2mm,
  parallelepiped offset y/.initial=2mm
}
\pgfdeclareshape{parallelepiped}
{
  \inheritsavedanchors[from=rectangle] 
  \inheritanchorborder[from=rectangle]
  \inheritanchor[from=rectangle]{north}
  \inheritanchor[from=rectangle]{north west}
  \inheritanchor[from=rectangle]{north east}
  \inheritanchor[from=rectangle]{center}
  \inheritanchor[from=rectangle]{west}
  \inheritanchor[from=rectangle]{east}
  \inheritanchor[from=rectangle]{mid}
  \inheritanchor[from=rectangle]{mid west}
  \inheritanchor[from=rectangle]{mid east}
  \inheritanchor[from=rectangle]{base}
  \inheritanchor[from=rectangle]{base west}
  \inheritanchor[from=rectangle]{base east}
  \inheritanchor[from=rectangle]{south}
  \inheritanchor[from=rectangle]{south west}
  \inheritanchor[from=rectangle]{south east}
  \backgroundpath{
    \southwest \pgf@xa=\pgf@x \pgf@ya=\pgf@y
    \northeast \pgf@xb=\pgf@x \pgf@yb=\pgf@y
    \pgfmathsetlength\pgfutil@tempdima{\pgfkeysvalueof{/pgf/parallelepiped offset x}}
    \pgfmathsetlength\pgfutil@tempdimb{\pgfkeysvalueof{/pgf/parallelepiped offset y}}
    \def\ppd@offset{\pgfpoint{\pgfutil@tempdima}{\pgfutil@tempdimb}}
    \pgfpathmoveto{\pgfqpoint{\pgf@xa}{\pgf@ya}}
    \pgfpathlineto{\pgfqpoint{\pgf@xb}{\pgf@ya}}
    \pgfpathlineto{\pgfqpoint{\pgf@xb}{\pgf@yb}}
    \pgfpathlineto{\pgfqpoint{\pgf@xa}{\pgf@yb}}
    \pgfpathclose
    \pgfpathmoveto{\pgfqpoint{\pgf@xb}{\pgf@ya}}
    \pgfpathlineto{\pgfpointadd{\pgfpoint{\pgf@xb}{\pgf@ya}}{\ppd@offset}}
    \pgfpathlineto{\pgfpointadd{\pgfpoint{\pgf@xb}{\pgf@yb}}{\ppd@offset}}
    \pgfpathlineto{\pgfpointadd{\pgfpoint{\pgf@xa}{\pgf@yb}}{\ppd@offset}}
    \pgfpathlineto{\pgfqpoint{\pgf@xa}{\pgf@yb}}
    \pgfpathmoveto{\pgfqpoint{\pgf@xb}{\pgf@yb}}
    \pgfpathlineto{\pgfpointadd{\pgfpoint{\pgf@xb}{\pgf@yb}}{\ppd@offset}}
  }
}
\pgfdeclareshape{document}{
\inheritsavedanchors[from=rectangle] 
\inheritanchorborder[from=rectangle]
\inheritanchor[from=rectangle]{center}
\inheritanchor[from=rectangle]{north}
\inheritanchor[from=rectangle]{north east}
\inheritanchor[from=rectangle]{north west}
\inheritanchor[from=rectangle]{south}
\inheritanchor[from=rectangle]{south east}
\inheritanchor[from=rectangle]{south west}
\inheritanchor[from=rectangle]{west}
\inheritanchor[from=rectangle]{east}
\backgroundpath{%
\southwest \pgf@xa=\pgf@x \pgf@ya=\pgf@y
\northeast \pgf@xb=\pgf@x \pgf@yb=\pgf@y
\pgf@xc=\pgf@xb \advance\pgf@xc by-5pt 
\pgf@yc=\pgf@ya \advance\pgf@yc by5pt
\pgfpathmoveto{\pgfpoint{\pgf@xa}{\pgf@ya}}
\pgfpathlineto{\pgfpoint{\pgf@xa}{\pgf@yb}}
\pgfpathlineto{\pgfpoint{\pgf@xb}{\pgf@yb}}
\pgfpathlineto{\pgfpoint{\pgf@xb}{\pgf@yc}}
\pgfpathlineto{\pgfpoint{\pgf@xc}{\pgf@ya}}
\pgfpathclose
\pgfpathmoveto{\pgfpoint{\pgf@xc}{\pgf@ya}}
\pgfpathlineto{\pgfpoint{\pgf@xc}{\pgf@yc}}
\pgfpathlineto{\pgfpoint{\pgf@xb}{\pgf@yc}}
\pgfpathclose
}
}
\tikzstyle{block} = [draw, fill=white, rectangle, 
    minimum height=3em, minimum width=6em]
    
\usetikzlibrary{backgrounds}
\usepackage{pifont}
\tikzstyle{startstop} = [rectangle, rounded corners, minimum width=2cm, minimum height=0.7cm,text centered, draw=black, fill=lime!30]
\newcommand\MakeUppercaseGreek[1]{
  \begingroup
    \let\psi\Psi
    \let\omega\Omega
    \let\gamma\Gamma
    \MakeUppercase{#1}
  \endgroup}
\newcommand{\vectorsym}[1]{\bm{#1}}
\newcommand{\expectation}[2]{\mathbb{E}_{#2}\left[#1\right]}

\newcommand{\normaldis}[2]{\mathcal{N}(#1,#2)}
\newcommand{\brackets}[1]{\left(#1\right)}
\newcommand{\abs}[1]{\left|#1\right|}
\newcommand{\norm}[1]{\left\lVert#1\right\rVert}
\newcommand{\normf}[1]{\left\lVert#1\right\rVert_{\mathrm{F}}}

\newcommand{\matsym}[1]{\mathbf{#1}}
\newcommand{\squareb}[1]{\left[{#1}\right]}
\newcommand{\prob}[1]{p_{\MakeUppercaseGreek{#1}}\left(#1\right)}
\newcommand{\trace}[1]{\mathrm{Tr}\left(#1\right)}
\newcommand{\probt}[2]{p_{#2}\left(#1\right)}

\newcommand{\divc}[2]{\frac{\partial#1}{\partial#2}}

\newcommand{\ceil}[1]{\left\lceil#1\right\rceil}
\newcommand{\idemat}{\matsym{I}}
\newcommand{\at}[2]{\left.#1\right\vert_{#2}}

\newcommand{\vecop}[1]{\mathrm{Vec}\brackets{#1}}

\newcommand{\R}[0]{\mathrm{R}}
\newcommand{\RGamma}[0]{\Gamma}
\newcommand{\Ps}[0]{\Theta}

\newcommand{\G}[0]{\mathrm{G}}
\newcommand{\GI}[0]{\nu}

\newcommand{\F}[0]{\mathrm{F}}
\newcommand{\NLL}[0]{\mathrm{L}}
\newcommand{\crb}[0]{\mathrm{CRB}}

\newcommand{\p}[0]{\vectorsym{\theta}}
\newcommand{\g}[0]{\vectorsym{\gamma}}
\newcommand{\z}[0]{\vectorsym{z}}
\newcommand{\Z}[0]{\vectorsym{Z}}
\newcommand{\vr}[0]{\vectorsym{r}}

\newcommand{\method}{GCRB }

\begin{document}
\raggedbottom

\begingroup
\let\clearpage\relax

\title{Learning to Bound: A Generative Cram\'er-Rao Bound}

\author{Hai Victor Habi, Hagit~Messer,~Life~Fellow,~IEEE and Yoram~Bresler,~Life~Fellow,~IEEE
\thanks{H.V. Habi and H. Meseer are with the School of Electrical Engineering, Tel Aviv University, Tel Aviv 6139001, Israel (e-mail: haivictorh@mail.tau.ac.il; messer@eng.tau.ac.il).}
\thanks{Y.Bresler is with the Department of Electrical and Computer Engineering, University of Illinois, Urbana-Champaign, USA (e-mail: ybresler@illinois.edu). His work was supported in part by ARO grant W911NF-15-1-0479.
}}

\markboth{October 2022}%
{Habi, Messer, and Bresler:  Learning to Bound: A Generative Cram\'er-Rao Bound}


\maketitle

\begin{abstract}
The Cram\'er-Rao bound (CRB), a well-known lower bound on the performance of any unbiased parameter estimator, has been used to study a wide variety of problems. However, to obtain the CRB,  requires an analytical expression for the likelihood of the measurements given the parameters, or equivalently a precise and explicit statistical model for the data. In many applications, such a model is not available.  Instead, this work introduces a novel approach to approximate the CRB using data-driven methods, which removes the requirement for an analytical statistical model. This approach is based on the recent success of deep generative models in modeling complex, high-dimensional distributions. Using a learned normalizing flow model, we model the distribution of the measurements and obtain an approximation of the CRB, which we call Generative Cram\'er-Rao Bound (GCRB). Numerical experiments on simple problems validate this approach,  and experiments on two image processing tasks of image denoising and edge detection with a learned camera noise model demonstrate its power and benefits.
\end{abstract}

\begin{IEEEkeywords}
Generative Models, Normalizing Flows, CRB, Parameter Estimation.
\end{IEEEkeywords}
\section{Introduction}
The Cram\'er-Rao Bound (CRB) is a lower bound on the variance of any unbiased parameter estimator \cite{rao1945information,kay1993fundamentals,van2004detection}.
It  has been used in a wide variety of estimation problems such as DOA\cite{van2004optimum}, TDOA\cite{catovic2004cramer}, etc. The CRB enables to understand the fundamental limits in a given parameter estimation problem, regardless of the algorithm used. However, to obtain an applicable CRB, it is required to have an analytical expression for the likelihood of the measurements given the parameters, or equivalently a precise and explicit statistical model for the measurements. In many applications, such a model is not available. Examples include device-specific noise statistics, such as in image sensors \cite{abdelhamed2019noise}, or radio frequency communications with jamming \cite{carmack2021neural} or unknown channel characteristics  \cite{yang2019generative}. 

Recently, generative models have shown state-of-the-art results in modeling complex, high-dimensional data distribution from images \cite{goodfellow2014generative,NEURIPS2018_d139db6a},  voice  \cite{oord2018parallel}, image noise \cite{abdelhamed2019noise} and communications channels \cite{8663987}. In this work, we suggest to use generative models to learn the measurement distribution from data. Then, using this generative model, we obtain an approximation to the CRB. We call this approach a \emph{Generative Cram\'er-Rao Bound (GCRB)} and show conditions under which the GCRB accurately approximates the CRB. Specifically, we use a normalizing flow \cite{kobyzev2020normalizing,papamakarios2021normalizing} to learn a generative model for the measurement distribution. This is used, in turn, to generate samples of the gradient of the log-likelihood and obtain, as an empirical mean,  an estimate of the Fisher Information Matrix (FIM).  We refer to this estimate as a \emph{Generative Fisher Information Matrix (GFIM)}. Finally, by inverting the GFIM, we obtain the GCRB.

{The GCRB enables the approximation of CRB in cases when the measurement distribution is completely unknown.}\footnote{{Our approach is somewhat related to the misspecified Cram\'er -Rao bound (MSCRB) \cite{fortunati2017performance} in that the MSCRB too can be evaluated without knowledge of the underlying true distribution by using data samples. However, the MSCRB provides a bound on accuracy of estimating parameters in an assumed (mispecified) model, using measurement taken from an actual unknown distribution. Instead, we aim to determine, from data, the true model and the bound on parameter estimates in the true model.} } To asses the {approximation quality we provide three theoretical bounds: i) a bound on the GFIM error due to imperfect learning in terms of two well-known measures of discrepancy between probability distributions (Total Variation Distance and Fisher Relative Information);  ii)} a bound on the error in the GCRB due to the use of an empirical mean to estimate the GFIM from a finite number of samples generated by the normalizing flow model; {; and iii) a bound on the relative error of the GCRB, combining the effects learning and sampling errors. }


To validate the GCRB, we examine two simple examples of parameter estimation with a Gaussian and non-Gaussian measurement distributions, respectively. First, we show analytically that the GCRB and the CRB produce the same results under optimal conditions, i.e., assuming an invertible   generative model that produces the exact measurement distribution.  Second, we illustrate a realistic setup where we train a standard normalizing flow on each of the two measurement distributions to evaluate its GCRB and compare it to the corresponding CRB. 

Then, to demonstrate the value of GCRB, we use two examples from image processing: image denoising, and edge position detection, in the presence of realistic, camera-specific noise.
We model the camera noise using a recently published normalizing flow model NoiseFlow \cite{abdelhamed2019noise}. With these examples, we show two main benefits of the GCRB: (1)  a lower bound for image denoising for several cameras, which provides a device-specific lower bound; and (2) we compare the GCRB lower bound on the estimation of the  position {and width} parameters of an edge in an image corrupted by camera noise to the CRB that would be obtained using two popular noise models: white Gaussian, and Noise Level Function (NLF) noises. This experiment demonstrates that the analytical CRB with specific assumed noise models (such as the white Gaussian or even the refined NLF  noise models) cannot capture the complex actual noise of image sensors and its effect on image processing performance, which is however successfully captured by the proposed GCRB.

The main contributions of this paper are the following.
\begin{itemize}
    \item We introduce a Generative Cram\'er-Rao Bound - a data-driven approach to approximate the CRB, eliminating the need for an analytical statistical data model.
    \item We demonstrate the benefit of the GCRB on two real-world problem of image denoising and edge detection.
    \item We evaluate the approximation quality (between the CRB and the \method) using two simple measurement distributions.
    \item We provide a theoretical bound on the GCRB error due to empirical sampling {and learning error}.


\end{itemize}
In the spirit of reproducible research, we make the code and trained models of the generative Cram\'er-Rao bound available online \cite{gencrb}.

The paper is organized as follows:  the Generative Cram\'er-Rao Bound is developed in Sec.~\ref{sec:gcrb} followed by {an analysis of it's theoretical properties in Sec.~\ref{sec:theory}}. A brief overview of normalizing flows in Sec.~\ref{sec:nf}.  In Sec.~\ref{sec:example_models} we present a set of parameter estimation examples, including simple parameter estimation in Gaussian and Non-Gaussian noise, and image processing with device-specific noise. The experimental results for the \method are described in Sec.~\ref{sec:experimental}, and Sec.~\ref{sec:conclusions} provides discussion and conclusions. {Sec.~\ref{sec:proofs} provides detailed proofs of the theoretical results of this paper.} Appendices are included in the online Supplementary Material.
\section{Generative Cramer-Rao Bound}\label{sec:gcrb}

We introduce the Generative Cram\'er-Rao Bound (GCRB), a data-driven approach to approximate the Cram\'er-Rao Bound (CRB). We begin with the measurements model, the classical CRB,  and problem statement. Then, we introduce our method to obtain the Generative Fisher Information Matrix (GFIM) and the GCRB using an invertiable generative model. 
{
\subsection{Notation}
Lower case italics 
$a$ and boldface $\vectorsym{a}$ indicate a scalar
and a vector, respectively, with $\norm{\vectorsym{a}}_2$ denoting 
the $l_2$ norm. The $i$-th element if vector $\vectorsym{a}$ will be indicated by $\squareb{\vectorsym{a}}_i$. 
Upper case boldface $\matsym{A}$ indicates a matrix, 
with its trace, determinant, transpose, Frobenius norm and spectral norm (largest singular value) denoted by
$\trace{\matsym{A}}$, $\det{\matsym{A}}$, 
$\matsym{A}^T$, $\normf{\matsym{A}}$, and $\norm{\matsym{A}}$,
respectively.  
An identity matrix of size $k\times k$ is denoted by  $\matsym{I}_k$.
For symmetric matrix $\matsym{A}$ the notations $\matsym{A} \succ 0$ (or $\matsym{A} \succeq 0$) mean that $\matsym{A}$ is positive-definite (or positive semi-definite).
For symmetric $\matsym{A}$ and $\matsym{B}$ the inequality $A \succ B$  mean that $A-B \succ 0$.}
\subsection{Data model and Problem Statement}
Consider a  data model described by random mapping, also known as a ”channel,” producing a random measurement $\R\brackets{\p}$ from a deterministic input $\p$. The channel is fully characterized by the probability density function (PDF) $\probt{\vectorsym{r};\p}{\R}$. Formally, let $\p\in\mathbb{R}^k$ be a parameter vector, $\R\in\mathbb{R}^d$  the measurement vector, and $\probt{\cdot;\p}{\R}:\mathbb{R}^d\xrightarrow{}\mathbb{R}^{\added{+}}$ the  probability density function of $\R$ for a given parameter value $\p$.  
The CRB is specified in terms of the negative log-likelihood (NLL) of $\R$ given $\p$
\begin{equation*}
    \NLL_{\R}\brackets{\p} \triangleq -\log\probt{\vectorsym{r};\p}{\R}
\end{equation*}
and the corresponding Fisher information matrix (FIM)
\begin{equation}\label{eq:data_fim}
    \F_{\R}\brackets{\p} \triangleq\expectation{\nabla_{\p}\NLL_{\R}\brackets{\p}\nabla_{\p}\NLL_{\R}\brackets{\p}^T}{\R},
\end{equation}
where $\expectation{}{\R}$ denotes the expectation with respect to $\R$.
For the CRB to apply, we assume that appropriate regularity conditions\cite{kay1993fundamentals,lehmann2006theory} hold. We list below those to which we appeal in this paper explicitly, with the understanding that the remaining regularity conditions hold too. 
\begin{assumption}\label{assum:crb_reg}
\label{assumption1}
$\probt{\vectorsym{r};\p}{\R}$ satisfies the   
following conditions:
\begin{enumerate}[label=A.\arabic*]
\item\label{sas:support} For all $\p \in \Ps$, where $\Ps$ is an open set, the densities $\probt{\vectorsym{r};\p}{\R}$ have a common support $\Upsilon=\{\vectorsym{r}:\probt{\vectorsym{r};\p}{\R}>0\} \subseteq \mathbb{R}^d$ that is independent of $\p$.
\item\label{sas:derivative} For any $\vectorsym{r}\in\Upsilon$ and $\p\in\Ps$ the derivative (gradient) $\nabla_{\p}\probt{\vr;\p}{\R}$ with respect to $\p$  exists and is finite.
\item\label{sas:r_fim_pd} For all $\p \in \Ps$, the FIM is positive definite, $\F_{\R} \succ 0$ .
\end{enumerate}
\end{assumption}
Let $\hat{\p}(\R)$ be an unbiased estimator of $\p$ from the measurement $\R\brackets{\p}$ that satisfies $\expectation{\|\hat{\p}(\R)\|_2^2}{\R} < \infty$. Then the {covariance matrix of the estimation error} 
of any such estimator of $\p$ satisfies the so-called \emph{information inequality}
{
\begin{align}\label{eq:infoineq}
\expectation{\brackets{\hat{\p}(\R)-\p}\brackets{\hat{\p}(\R)-\p}^T }{\R}\succeq    \crb_{\R}\brackets{\p}\triangleq\squareb{\F_{\R}\brackets{\p}}^{-1}.
\end{align}}


We wish to determine $\crb_{\R}\brackets{\p}$ when the channel pdf $\probt{\vectorsym{r};\p}{\R}$ is unknown, and we are instead given representative data samples. We define this problem as follows. 
\begin{problem}\label{problem:crb} Let $\Ps \subseteq \mathbb{R}^k$ be an open set. Assume that $\probt{\vectorsym{r};\p}{\R}$ and $p(\p)$ 
satisfy Assumptions \ref{assumption1} and \ref{ass:well-trained-base}. 
        Given a data set $\mathcal{D}=\{\p_i,\vr_i\}_{i=1}^l$ of $l$ channel input-output samples that are independent and identically-distributed (i.i.d) 
as    $\vr_i \sim  \probt{\vectorsym{r}_i;\p_i}{\R}, \p_i\sim
p(\p)$, obtain an approximation to the Cram\'er-Rao lower bound on the estimation of parameter $\p\in\Ps$ from the measurement $\R(\p)$:
\begin{equation*}
    \crb_{\R}\brackets{\p}\quad\forall\p\in\Ps.
\end{equation*} 
\end{problem}
The additional assumptions indicated above in Problem~\ref{problem:crb} are the following.
\begin{assumption}\label{ass:well-trained-base}  $\,$ 

\begin{enumerate}[label=A.\arabic*]
\setcounter{enumi}{3}
    \item\label{sas:theta_bound} $\Ps$ is bounded set. 
    \item\label{sas:theta_prop} $
    p(\p) >\epsilon_\Ps>0\quad\forall\p\in\Ps$
    \item\label{sas:connected_set} $\Upsilon$ is connected set.
\end{enumerate}
\end{assumption}

Assumptions~\ref{assumption1} are required in Problem~\ref{problem:crb} for the validity of the information inequality. Assumptions~\ref{ass:well-trained-base} facilitate the training of  the normalizing flow and generator. 
Specifically, 
\ref{sas:connected_set}
facilitates the universal approximation by the generator; and \ref{sas:theta_bound} and \ref{sas:theta_prop}
enable all $\p\in\Ps$ to be present in the training set with some non-vanishing probability, and limit the degree of generalization to unseen $\p$ required of the generator. Note that while $\p$ is a deterministic unknown parameter for the purposes of the CRB, $
p(\p)$ describes the \emph{sampling} distribution of the training set $\mathcal{D}$. We will address these assumptions where  relevant.  
\subsection{Method}\label{sec:method}
We address Problem \ref{problem:crb} with a two-stage approach. In the first stage, we train a conditional normalizing flow (invertible neural network) that learns the distribution of the measurements. Training of normalizing flows is a well-studied subject, and  we only provide a short overview in Section \ref{sec:nf}. In the second stage, we obtain an approximation of the CRB from the trained conditional normalizing flow. 

In the rest of this section, we describe how to approximate the CRB using a trained conditional normalizing flow. 
Let $\GI\brackets{\g;\p}$
be a trained conditional normalizing flow with conditioning input $\p$ and data input $\g$. Then $\G\brackets{\z;\p}$, the inverse of $\GI$ with respect to $\g$, is a conditional generator with conditioning input $\p$ and random input $\z$ with known and tractable distribution (usually $\z\sim\normaldis{0}{\idemat}$), producing the output:
\begin{equation} \label{eq:gen-model}
    \RGamma\brackets{\p}=\G\brackets{\z;\p}.
\end{equation}
While $\G$ is usually obtained directly from $\GI$ by a simple transformation and does not require separate training (see Sec.~\ref{sec:nf}), we refer to $\G$ as a \emph{trained generator} because it is obtained  from the trained normalizing flow $\GI$. We assume (in a sense soon to be made precise) that  the trained generator simulates the random measurement process $\R\brackets{\p}$ accurately, i.e. $\RGamma(\p)$ has the same distribution as $\R\brackets{\p}$.

We make the standard assumption that for each $\p$, $\G\brackets{\cdot;\p}: \mathbb{R}^d \mapsto \mathbb{R}^d$ is a bijection, i.e., it has an inverse $\GI\brackets{\cdot;\p}$, and that both are differentiable functions, that is, for each $\p$, the mapping $\G\brackets{\cdot;\p}$ is a diffeomorphism. Furthermore,  for reasons explained below, we strengthen the  differntiability assumption to $\G\in C^2$, that is the first and second order derivatives, including the mixed derivative of $\G$ w.r.t  $\z$ and $\p$ exist and are continuous. 
A trained $\G$ is a deterministic function of $\p$ and $\z$, implemented as a neural network $\G\brackets{\cdot;\p}$ that is invertible in its first parameter. Thanks to the randomness of $\z$, the generative model \eqref{eq:gen-model} is a random mapping from $\p$ to $\RGamma\brackets{\p}$.

{It is important to note that when the measurement distribution is not continuous (e.g., quantized measurement), a different approach is needed for learning the CNF. The most straightforward approach is to add a prepossessing stage such dequantization \cite{kobyzev2020normalizing,hoogeboom2021learning} making the measurement distribution continuous. We used this approach to apply the GCRB to the problem of frequency estimation from quantized measurements \cite{habi2022generative}. An alternative approach  can be to use a CNF built for discrete data distribution \cite{nielsen2020closing}. However, the focus of this work is on continuous measurements, leaving extensions to discrete distributions for  future work.}

It follows, using the standard formula of transformation of random variables, that  the probability density function of $\RGamma\brackets{\p}$ is
\begin{equation}\label{eq:flow_inv}
 \probt{\vectorsym{\gamma};\p}{\RGamma}=\probt{\GI\brackets{\vectorsym{\gamma};\p}}{\z}\abs{\mathrm{det}\matsym{J}_{\GI}\brackets{\vectorsym{\gamma}; \p}},
\end{equation}
where $\matsym{J}_{\GI}\brackets{\g; \p}=\divc{\GI\brackets{\vectorsym{\gamma}; \p}}{\vectorsym{\gamma}}$ is the Jacobian matrix of the transformation $\GI\brackets{\vectorsym{\gamma}; \p}$ with respect to $\vectorsym{\gamma}$. 
Since both $\G$ and $\GI$ are known functions and the pdf of $\z$ is known (standard normal), in principle, the pdf $\probt{\gamma;\p}{\RGamma}$ can be determined. 

Given the trained normalizing flow $\GI$ and the corresponding generator $\G$, we compute the GCRB as follows. First using \eqref{eq:flow_inv} we determine (as detailed in Appendix~\ref{apx:score_vector_hybrid}) the so-called score vector 
\begin{align}
        &\vectorsym{s}_{\p}\brackets{\vectorsym{z}} \triangleq \at{\nabla_{\p}\log\probt{\vectorsym{\gamma};\p}{\RGamma}}{\vectorsym{\gamma}=\G\brackets{\z;\p}} \nonumber\\
&=\at{\nabla_{\p}\log\left[\probt{\GI\brackets{\vectorsym{\gamma};\p}}{\z}\abs{\mathrm{det}\matsym{J}_{\GI}\brackets{\vectorsym{\gamma}; \p}}\right]}{\vectorsym{\gamma}=\G\brackets{\z;\p}} \label{eq:score_vector_hybrid}\\        
&= \at{\divc{\GI\brackets{\vectorsym{\gamma};\p}}{\p}^{\mathrm{T}}}{\vectorsym{\gamma}=\G\brackets{\z;\p}}  \hspace{-1.2cm} \times \nabla_{\z}\log\prob{\z}
+\at{\vectorsym{k}\brackets{\g,\p}}{\g=\G\brackets{\z;\p}}, \label{eq:score_vector}
  \end{align}
  \begin{equation} \label{eq:score_vector-k}
 \text{where} \quad     \squareb{\vectorsym{k}\brackets{\g,\p}}_i=\trace{\matsym{J}_{\GI}^{-1}\brackets{\vectorsym{\gamma}; \p}\divc{\matsym{J}_{\GI}\brackets{\vectorsym{\gamma}; \p}}{\squareb{\p}_i}},
  \end{equation}
{where }
 $\divc{\GI\brackets{\vectorsym{\gamma};\p}}{\p}$ is the Jacobian matrix of $\GI$ w.r.t  $\p$, and in \eqref{eq:score_vector-k} $\divc{\matsym{J}_{\GI}\brackets{\vectorsym{\gamma}; \p}}{\squareb{\p}_i}$ is a derivative matrix of the Jacobian matrix $\matsym{J}_{\GI}$ w.r.t the $i$-th component of $\p$.
 Note that to evaluate the score vector for a given $\z$  both $\G$ and $\GI$ are used. We therefore refer to \eqref{eq:score_vector} as a hybrid score vector. As an alternative, we show in Appendix~\ref{apx:generator_score_vector}
 an equivalent form that only uses the generator $\G$.
 
 To perform the computation in \eqref{eq:score_vector}-\eqref{eq:score_vector-k} we need to be able to evaluate the following derivatives:  $\GI$ w.r.t  $\p$, $\GI$ w.r.t  $\z$, and a mixed second derivative of  $\GI$ w.r.t to $\z$ and $\p$. As elaborated in Appendix~\ref{apx:generator_score_vector}, this motivates the differentiability conditions imposed on  $\G$ above. We also require that the log-likelihood of the base distribution $\prob{\z}$ be differentiable, which is satisfied by the Gaussian distribution.  {Moreover, $\matsym{J}_{\GI}$ should be invertible, which is usually guaranteed by common layer structures in the CNF literate by  combining design of the layer and training loss objective \eqref{eq:likelihood}. The invertibility condition ${\mathrm{det}\matsym{J}_{\GI}\brackets{\vectorsym{\gamma};\p}}\neq 0$ also enables stable training.
 }

As a practical matter, because $\GI$ and $\G$ are implemented as a neural networks, the required derivatives of $\GI$ and $\G$ w.r.t to their respective inputs $\vectorsym{\gamma}$, $\z$ and $\p$ can be easily evaluated in common deep learning frameworks such as PyTorch\cite{paszke2019pytorch}, TensorFlow, etc.

Given the score vector, we compute the Generative Fisher Information Matrix (GFIM):
\begin{equation}\label{eq:generator_fim}
\begin{split}
    \F_{\mathrm{G}}\brackets{\p} \triangleq \expectation{\vectorsym{s}_{\p}\brackets{\Z}\vectorsym{s}_{\p}\brackets{\Z}^T}{\Z}.
\end{split}
\end{equation}
In practice, to avoid integration in \eqref{eq:generator_fim}, the expected value with respect to $\Z$ is estimated as an empirical mean by sampling from $p_{\Z}$. 
The result is an empirical Generative Fisher Information Matrix (eGFIM)
that is computed as
\begin{equation}\label{eq:gfim}
    \overline{\F_{\mathrm{G}}}\brackets{\p} \triangleq\frac{1}{m}\sum_{i=1}^m\vectorsym{s}_{\p}\brackets{\vectorsym{z}_i}\vectorsym{s}_{\p}\brackets{\vectorsym{z}_i}^T,
\end{equation}
using $m$ samples  $\z_i \sim p_{\Z}$.  Finally, 
we approximate the CRB using the empirical estimate of the GCRB (eGCRB) $\overline{\mathrm{GCRB}_{\G}}$ associated with  generator $\G$ by 
\begin{equation} \label{eq:inv-GFIM}
    \overline{\mathrm{GCRB}_{\G}}\brackets{\p}=\overline{\F_{\mathrm{G}}}\brackets{\p}^{-1}.
\end{equation}

Given the trained neural networks $\G$ and $\GI$, the computation of $\overline{\mathrm{GCRB}_{\G}}$ for a given value of $\p$ is illustrated in Fig.~\ref{fig:crb_approximation}. It involves $m$ uses of the neural networks $\G$ (to generate $\vectorsym{\gamma}$) and $\GI$ (to generate the various derivatives) and the simple computations in \eqref{eq:score_vector}-\eqref{eq:score_vector-k}, \eqref{eq:gfim}, and \eqref{eq:inv-GFIM},
so can be computationally cheap.

In the rest of this {subsection, we address  
a modification of the GCRB to improve the learning of the generator in the practical situation of a finite training data set.
Because some regions of the measurement space $\Upsilon$ may have few or no training samples,}
we need to bound the region $\hat{\Upsilon}$ where generated samples can be trusted. We define this region by its assumed properties. 
\begin{assumption}[Trusted Region]\label{assum:trusted_region} $\ $
\begin{enumerate}[label=A.\arabic*]

    \setcounter{enumi}{6}
    \item\label{sas:bounded_and_connected} $\hat{\Upsilon} \subseteq\Upsilon$ is a connected and closed and bounded (hence compact) set.
    \item\label{sas:large_region} $\hat{\Upsilon}$ is large enough  that for some chosen  $\epsilon_r\geq 0$
    \begin{equation*}
        \int_{\vectorsym{r}\not\in\hat{\Upsilon}}\probt{\vectorsym{r};\p}{\R}d\vectorsym{r}\leq\epsilon_r\quad \forall\p\in\Ps .
    \end{equation*}
    \item\label{sas:present_in_training} $\probt{\vectorsym{r};\p}{\R}>\epsilon>0\quad\forall\vectorsym{r}\in\hat{\Upsilon}$.
\end{enumerate}
\end{assumption}

To ensure that the computation of the GCRB is performed using a sample generated on the trusted region, we add an optional 
trimming step that removes un-trusted generated samples $\g = \G(\z) \not\in\hat{\Upsilon}$.
The trimming step ensures that only values of $\z$ that correspond to trusted $\g \in \hat{\Upsilon}$ are used in the computation of GCRB.
By Assumption~\ref{assum:trusted_region} the effect of this trimming on the approximation quality should be a negligible. Furthermore, assumption \ref{sas:present_in_training} enables all $\vectorsym{r}\in\hat{\Upsilon}$ to be present in the training set with some non-vanishing probability.
Algorithm~\ref{alg:sampling} describes the evaluation of the eGCRB, with the trimming step included.
\begin{algorithm}
\caption{eGCRB Sampling }\label{alg:sampling}
\begin{algorithmic}
\Require $\G$, $\GI$, $B$, $\hat{\vectorsym{r}}$, $m$, $\p$
\State $\mathrm{S} \gets \emptyset$
\While
{$|\mathrm{S}|<m$}
\State $\z\sim\prob{\z}$ 
\State $\g=\G\brackets{\z;\p}$ \Comment{Generator Step}
\If{$\g\in\hat{\Upsilon}$} \Comment{Timming Step}
    \State $\hat{\vectorsym{s}}=\vectorsym{s}_{\p}\brackets{\vectorsym{z}}$\Comment{Compute score vector \eqref{eq:score_vector_hybrid}}.
    \State $\mathrm{S}\gets\mathrm{S}\cup\{\hat{\vectorsym{s}}\}$  \Comment{Append to $\hat{\vectorsym{s}}$ to score set.}
\EndIf
\EndWhile
\State {$\overline{\F_{\mathrm{G}}}\brackets{\p}=\frac{1}{m}\sum_{\hat{\vectorsym{s}}\in\mathrm{S}}\hat{\vectorsym{s}}\hat{\vectorsym{s}}^T$}.\Comment{Compute eGFIM} 
\State $\overline{\mathrm{GCRB}_{\G}}\brackets{\p}=\overline{\F_{\mathrm{G}}}\brackets{\p}^{-1}$ \Comment{Invert eGFIM to obtain eGCRB}
\end{algorithmic}
\end{algorithm}

The trimming step, in the spirit of standard trimmed mean computation in robust statistics \cite{1457435}, is a kind of outlier removal step, which is a well-studied but also active field of research (cf. \cite{lugosi2021robust,yang2018general} and the references therein). 
We propose a simple heuristic  trimming criterion; a more refined criterion may improve the eGCRB accuracy when only limited training data is available. 
The trimming process consist of two steps. First, we evaluate the mean $\bar{\vectorsym{r}}$ and an upper bound $B$ on the spread of $\vectorsym{r}$ in the training set $\mathcal{D}$:
\begin{align*}
    \bar{\vectorsym{r}}&=\frac{1}{|\mathcal{D}|}\sum_{i}\vectorsym{r}_i,\\
    B&=\max\limits_{i} \norm{\vectorsym{r}_i-\bar{\vectorsym{r}}}.
\end{align*}
Then, 
the trusted set is defined as 
\begin{equation*}
    \hat{\Upsilon}=\{\g \in \Upsilon:\norm{\g-\bar{\vectorsym{r}}}\leq B\}.
\end{equation*}
This trimming is designed to exclude samples $\vectorsym{\gamma} = \G(\z)$ generated in regions where no training samples were available to train the normalizing flow. This will reduce the requirement of the normalizing flow and generator to extrapolate during inference outside the coverage of the training set. Note that thanks to the adaptivity of $ \hat{\Upsilon}$ to the training set, as the size of the training set $|\mathcal{D}| \rightarrow \infty$, the "unrepresented probability" vanishes: $\epsilon_r \rightarrow 0$.


 \tikzset{doc/.style={document,fill=blue!10,draw,thin,minimum
height=1.2cm,align=center},
pics/.cd,
pack/.style={code={%
\draw[fill=blue!50,opacity=0.2] (0,0) -- (0.5,-0.25) -- (0.5,0.25) -- (0,0.5) -- cycle;
\draw[fill=blue!50,opacity=0.2] (0,0) -- (-0.5,-0.25) -- (-0.5,0.25) -- (0,0.5) -- cycle;
\draw[fill=blue!60,opacity=0.2] (0,0) -- (-0.5,-0.25) -- (0,-0.5) -- (0.5,-0.25) -- cycle;
\draw[fill=blue!60] (0,0) -- (0.25,0.125) -- (0,0.25) -- (-0.25,0.125) -- cycle;
\draw[fill=blue!50] (0,0) -- (0.25,0.125) -- (0.25,-0.125) -- (0,-0.25) -- cycle;
\draw[fill=blue!50] (0,0) -- (-0.25,0.125) -- (-0.25,-0.125) -- (0,-0.25) -- cycle;
\draw[fill=blue!50,opacity=0.2] (0,-0.5) -- (0.5,-0.25) -- (0.5,0.25) -- (0,0) -- cycle;
 \draw[fill=blue!50,opacity=0.2] (0,-0.5) -- (-0.5,-0.25) -- (-0.5,0.25) -- (0,0) -- cycle;
\draw[fill=blue!60,opacity=0.2] (0,0.5) -- (-0.5,0.25) -- (0,0) -- (0.5,0.25) -- cycle;
}}}
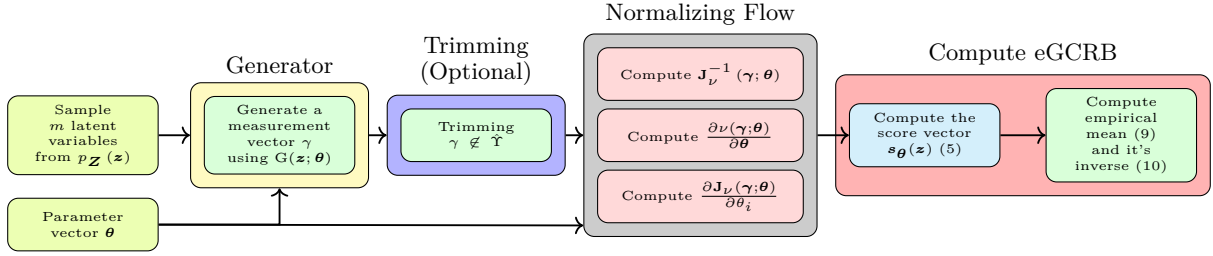
\begin{figure*}
    \centering
    \begin{tikzpicture}[font=\small,every label/.append style={font=\small,align=center}]
        \node (net) [startstop,font=\tiny,text width=1.4cm] {Sample $m$ latent variables from $\prob{\z}$};
        \node (p) [below=0.3cm of net,startstop,font=\tiny,text width=1.4cm] {Parameter vector $\p$};
        \node[right=0.6cm of net,startstop,fill=green!15, text width=1.4cm,font=\tiny,align=center] (bf) {Generate a measurement vector $\gamma$ using 
		$\G(\z;\boldsymbol{\theta})$};
		
		\node[right=0.6cm of bf,startstop,fill=green!15, text width=1.4cm,font=\tiny,align=center] (trimming) {Trimming $\gamma\not\in\hat{\Upsilon}$};

		\node[right=0.6cm of trimming,startstop,fill=red!15, text width=2.5cm,font=\tiny,align=center] (d1) {Compute $\divc{\GI\brackets{\vectorsym{\gamma};\p}}{\p}$};
		\node[above=0.1cm of d1,startstop,fill=red!15, text width=2.5cm,font=\tiny,align=center] (d2) {Compute $\matsym{J}_{\GI}^{-1}\brackets{\vectorsym{\gamma}; \p}$};
		\node[below=0.1cm of d1,startstop,fill=red!15, text width=2.5cm,font=\tiny,align=center] (d3) {Compute $\divc{\matsym{J}_{\GI}\brackets{\vectorsym{\gamma}; \p}}{\theta_i}$};

        \node[right=0.6cm of d1,startstop,fill=cyan!15, text width=1.4cm,font=\tiny,align=center] (score) {Compute the score vector $\vectorsym{s}_{\boldsymbol{\theta} }(\z)$ \eqref{eq:score_vector_hybrid}};
        
        \node[right=0.6cm of score,startstop,fill=green!15, text width=1.4cm,font=\tiny,align=center] (fim) {Compute  empirical  mean  \eqref{eq:gfim} and it's inverse \eqref{eq:inv-GFIM}};
        \begin{pgfonlayer}{background}
         \node[draw,rounded corners,fit=(d1) (d2) (d3),fill=black!20,inner sep=5pt,label={Normalizing Flow}](fit1){};
         \node[draw,rounded corners,fit=(bf),fill=yellow!30,inner sep=5pt,label={Generator}](fit2){};
         \node[draw,rounded corners,fit=(score) (fim),fill=red!30,inner sep=5pt,label={Compute eGCRB}](fit3){};
         \node[draw,rounded corners,fit=(trimming),fill=blue!30,inner sep=5pt,label={Trimming \\ (Optional)}](fit4){};
        \end{pgfonlayer}
        
        \draw[->,line width=0.25mm] (net)  -- (fit2);
        \draw[->,line width=0.25mm] (p)  -| (fit2);
        
        \draw[->,line width=0.25mm] (p)  -- ([yshift=-1.2cm] fit1.west);
        
        \draw[->,line width=0.25mm] (fit2)  -- (fit4);
        \draw[->,line width=0.25mm] (fit4)  -- (fit1);
        \draw[->,line width=0.25mm] (fit1)  -- (score);
        \draw[->,line width=0.25mm] (score)  -- (fim);
        
        
        
        
    \end{tikzpicture}
    
	\caption{Generative Cramer Rao bound using normalizing flow 
	}
	\label{fig:crb_approximation}
\end{figure*}

\section{Theoretical {Properties}}\label{sec:theory}
This section addresses  {three} questions:  {(i)
What are the errors introduced into the GFIM by learning the measurement distribution?} (ii) What is the error introduced by using an empirical mean eGFIM  to estimate the GFIM? (iii) When does the 
 {approximation eGCRB to the CRB computed using the learned generative model in} the proposed approach  {converge to} the correct CRB?
Our key assumption in (iii) will be that the generative model is expressive enough and the training data set has sufficient size and diversity of values of $\p$ and $\R$ that the training is successful, resulting in a generative model that simulates the random mapping $\R\brackets{\p}$. 

\subsection{Learning Error}\label{sec:learning_error}
 {In this part we address the error induced by 
replacing the true measurement distribution $p_R$ by the learned distribution $p_\RGamma$ with trimming of the generator, meaning that $\hat{\Upsilon}$ is a strict subset of $\Upsilon$. 
We account for the deviation between $p_\RGamma$ and $p_R$ on their common support $\hat{\Upsilon} \bigcap \Upsilon $, as well as on  the truncated region $\Upsilon \setminus \hat{\Upsilon}$ where $p_\RGamma=0$.}

Define
\begin{align} \label{eq:hat_f_g}
    \hat{\F}_{\G}\brackets{\p}\triangleq\int_{\hat{\mathcal{Z}} }\vectorsym{s}_{\p}\brackets{\z}\vectorsym{s}_{\p}\brackets{\z}^T \prob{\z} d\z,
\end{align}
as the result of the  GFIM calculation over the trimmed latent variable set $\hat{\mathcal{Z}}=\{\z:\G\brackets{\z;\p}\in\hat{\Upsilon}\}$.  {We begin by 
introducing bounds on the generated and the true score vectors.}

 {\begin{lemma}\label{lemma:bounded_score_vector}
 Let $\vectorsym{s}_{\p}\brackets{\vectorsym{z}}$ be a score vector computed using a trimmed and differentiable $\G \in C^2$ generator $\G$ and it's inverse $\GI$. Then 
$\norm{\vectorsym{s}_{\p}\brackets{\vectorsym{z}}}_2\leq C_s\brackets{\p} < \infty$ { $\forall\p\in\Theta,\forall \vectorsym{z} \in \hat{\mathcal{Z}}$.}
\end{lemma}}
 {
{This result (proved in Sec.~\ref{proof:bound_score_vector}) shows that the score vector is bounded in 2-norm.} 
Next, 
we introduce an additional assumption, that the true measurement distribution too has a bounded score vector. 
\begin{assumption}[Bounded True Score Vector]\label{assum:bound_score_true_no_trim}
\begin{equation}
    \norm{\nabla_{\p}\NLL_{\R}\brackets{\p}}_2\leq C_\R\brackets{\p} < \infty \quad \forall\p\in\Theta,\forall\vectorsym{r}\in\Upsilon.
\end{equation}
\end{assumption}
Note that this assumption is a slightly more restrictive version of Assumption~\ref{sas:derivative}.}

Then we have our main results.
{ 
\begin{theorem}[GFIM Learning Errors]\label{thm:fim_pd}
Let $\G$ be a normalizing flow trained on { $\R \sim p_{\R}$,  
where $p_{\R}$} has a bounded score vector (Assumption \ref{assum:bound_score_true_no_trim}). Then
\begin{subequations} 
\begin{equation}\label{eq:fim_error_bound}
    \norm{\F_{\R}\brackets{\p}-\hat{\F}_{\G}\brackets{\p}}\leq\eta\brackets{\p} \qquad\forall\p\in\Theta,
\end{equation}
\begin{align}\label{eq:eta_expression}
    \eta\brackets{\p}&\triangleq 2 C_{\R}^2\brackets{\p}\mathrm{TV}\brackets{p_{\R},p_{\RGamma};\p} \\
    &+2\brackets{\norm{\hat{\F}_{\G}\brackets{\p}}\mathrm{I}_\mathrm{F}\brackets{p_{\RGamma},p_{\R};\p}}^{1/2}
    + \mathrm{I}_\mathrm{F}\brackets{p_{\RGamma},p_{\R};\p} \nonumber
\end{align}
\end{subequations}
where 
$$\mathrm{TV}\brackets{p_{\RGamma},p_{\R};\p}\triangleq \frac{1}{2} \int_{\Upsilon}\abs{\probt{\vr;\p}{\RGamma}-\probt{\vr;\p}{\R}}d\vr$$
is the total variation distance\cite{tsybakov2004introduction} between the PDFs $p_{\RGamma}$ and $p_{\R}$,  and 
$$\mathrm{I}_\mathrm{F}\brackets{p_{\RGamma},p_{\R};\p}\triangleq\int_{\Upsilon}\probt{\vr;\p}{\RGamma}\norm{\nabla_{\p}\log\brackets{\frac{\probt{\vr;\p}{\RGamma}}{\probt{\vr;\p}{\R}}}}^2_2d\vr$$ 
is the Fisher relative information\cite{hammad1978mesure,zegers2015fisher} between $p_{\RGamma}$ and $p_{\R}$. 

\end{theorem}}



	 {Theorem~\ref{thm:fim_pd} (which is proved in Sec.~\ref{proof:fim_error}) bounds the error in learning the FIM in term of the total variation (TV) distance and the Fisher
relative information between the true and the learned measurement distributions, $p_\R$ and $p_\RGamma$.  The TV distance term captures both the trimming error and sample generation error, whereas the  Fisher
relative information term accounts for the  errors in learning the derivative of $\log p_\R$, namely the score vector. }
 {Both $\mathrm{TV}\brackets{p_{\RGamma},p_{\R};\p}$  and $\mathrm{I}_\mathrm{F}\brackets{p_{\RGamma},p_{\R};\p}$
are non-negative  and vanish if and only if $\probt{\vr;\p}{\RGamma}=\probt{\vr;\p}{\R}\quad\forall\vr\in\Upsilon$. Furthermore, both metrics are bounded; the TV distance by definition, and the Fisher relative information is bounded as a direct consequence of Lemma~\ref{lemma:bounded_score_vector} and Assumption~\ref{assum:bound_score_true_no_trim} }

 {The impact of the learning error on the GCRB is given by the following corollary, where we use Assumption A.3, that the FIM is positive definite, to provide conditions in terms of its strictly positive smallest eigenvalue  $\lambda_{\min} \left(\F_{\R}\brackets{\p}\right) >0$.}
{
\begin{corollary}\label{cor:learning_error}
Suppose that in addition to the assumptions
in Theorem~\ref{thm:fim_pd}, we have $\eta(\p) < \lambda_{\min} (\F_{\R}\brackets{\p}) $. Then
\begin{subequations}\label{eq:CRB_errorV2}
\begin{equation}
\label{eq:GCRB_bound}
    \|\hat{\F}_{\G}\brackets{\p}^{-1} \| \leq \left[\lambda_{\min} (\F_{\R}\brackets{\p}) - \eta(\p) \right]^{-1} \end{equation}
    \begin{align}
    \|\crb_{\R}\brackets{\p} & - \mathrm{GCRB}(\p) \| =
    \| \F_{\R}\brackets{\p}^{-1} - \hat{\F}_{\G}\brackets{\p}^{-1} \| \nonumber \\
    & \leq \|\F_{\R}\brackets{\p}^{-1}\| \cdot \| \hat{\F}_{\G}\brackets{\p}^{-1}\|\cdot \eta(\p) \label{eq:GCRB_error_bound}
\end{align}
\end{subequations}
\end{corollary}
}

 {Note that \eqref{eq:GCRB_bound} in Corollary~\ref{cor:learning_error}(which is proved in Sec.~\ref{proof:well_cond}) is a guarantee that the GFIM is positive definite, i.e., the GCRB is finite, if the condition of the Corollary is satisfied. The second result, bounds the deviation of  the GCRB from the CRB in terms of the FIM learning error.}

 {
To help further interpret Corollary~\ref{cor:learning_error}, consider the relative (normalized) learning error in the FIM, $\frac{\eta(\p)}{\| \F_{\R}\brackets{\p}\|}$.  The condition of the Corollary then becomes { $\kappa(F_{\R}\brackets{\p})\frac{\eta(\p)}{\| \F_{\R}\brackets{\p}\|} < 1$,
} where $\kappa(F_{\R}\brackets{\p}) \triangleq \| F_{\R}\brackets{\p}\| \cdot \| F_{\R}\brackets{\p}^{-1}\|$ is the 2-norm condition number of $F_{\R}\brackets{\p}$ (which is also equal to the condition number of $\crb_{\R}\brackets{\p}$). Hence, the requirement of the corollary on the learning error $\eta(\p)$ is easy to satisfy for a well-conditioned FIM (or CRB), and becomes more demanding with increasing condition number.} 

 {Next, consider the case of small learning error, $\frac{\eta(\p)}{\| \F_{\R}\brackets{\p}\|} \ll 1 $. Then, by standard arguments, \eqref{eq:GCRB_error_bound} yields 
\begin{equation}
\label{eq:GCRB_Rel_error}
    \frac{\|\crb_{\R}\brackets{\p}  - \mathrm{GCRB}(\p)\|}%
    {\|\crb_{\R}\brackets{\p}\|} \leq
    \kappa(F_{\R}\brackets{\p}) \frac{\eta(\p)}{\| \F_{\R}\brackets{\p}\|} +\epsilon_3 
\end{equation}
where 
{$\epsilon_3 = O\left( (\frac{\eta(\p)}{\| \F_{\R}\brackets{\p}\|})^2\right)$, that is, 
the inequality is dominated by the first term on the right hand side, with the remainder $\epsilon_3$ of second order in the relative FIM learning error $\frac{\eta(\p)}{\| \F_{\R}\brackets{\p}\|}$, and hence negligible.} 
By \eqref{eq:GCRB_Rel_error}, the relative error in the GCRB is bounded by the relative GFIM learning error, scaled by the condition number of the FIM. Again, the better the conditioning of the FIM, the lower the sensitivity of the GCRB to the GFIM learning error.}

 {Furthermore, using \eqref{eq:eta_expression} to express the relative error in learning the GFIM for the case of small learning error 
yields the simplified expression
\begin{equation} \label{eq:eta_expression_normalized}
    \frac{\eta\brackets{\p}}{\|\F_{\R}\brackets{\p}\|} \approx
    2 \frac{C_{\R}^2\brackets{\p}}{\|\F_{\R}\brackets{\p}\|} \mathrm{TV}\brackets{p_{\R},p_{\RGamma};\p} \
    +2 \sqrt{\tilde{ \mathrm{I}}_\mathrm{F}\brackets{p_{\RGamma},p_{\R};\p}}
    \end{equation}
    where 
    $\tilde{\mathrm{I}}_\mathrm{F}\brackets{p_{\RGamma},p_{\R};\p} \triangleq  
    \mathrm{I}_\mathrm{F}\brackets{p_{\RGamma},p_{\R};\p}/\|\F_{\R}\brackets{\p}\|$ is the Fisher relative information between $p_{\RGamma}$ and $p_{\R}$ normalized by the Fisher information for $p_{\R}$. Now all terms in \eqref{eq:eta_expression_normalized} are dimensionless and naturally normalized. (Recall that $C_{\R}^2\brackets{\p}$ and $\|\F_{\R}\brackets{\p}\|$ scale similarly with the magnitude of the true score vector.)
    }

\subsection{Sampling Error}
Here, we study the effects of finite number of samples in \eqref{eq:gfim} on the accuracy of the  {estimation of $\hat{\F}_{\G}\brackets{\p}$}, by deriving an upper bound on the error. 

\begin{theorem}[Sampling Error]\label{thm:gcrb_sample_bound}
	
	Let $\overline{\F_{\mathrm{G}}}\brackets{\p}^{-1}$ be the eGCRB computed using a trimmed 
	{generator $\G \in \mathcal{C}^2$ and its inverse $\GI$ -- the corresponding normalizing flow, trained on $\R$.} 
	Assume that Assumptions~\ref{assum:trusted_region},~\ref{assum:bound_score_true_no_trim} hold, and  
	  {$\hat{\F}_{\G}\brackets{\p}\succ0\quad\forall\p\in\Theta$, which implies that  $\norm{\hat{\F}_{\G}\brackets{\p}^{-1}}\leq \mathrm{C_G}\brackets{\p}$.} 
	Then there exist  absolute constants $C_1, C_2>0$ such that provided that $m>C_1\brackets{1+u}\mathrm{C_G}\brackets{\p}^2$, for any $u>0$ we have,
	with probability at least $1-\exp\brackets{-u}$: 
	{ \begin{align}
        \normf{\overline{\mathrm{GCRB}_{\G}}\brackets{\p}-\hat{\F}_{\G}\brackets{\p}^{-1}}\leq\mathrm{B_s}\brackets{\p} ,
	\end{align}}
	where  {$\mathrm{B_s}\brackets{\p}\triangleq C_2\norm{\hat{\F}_{\G}\brackets{\p}^{-1}}^2\mathrm{C_s}\brackets{\p}^2\sqrt{\frac{1+u}{m}}$}.
\end{theorem}

Theorem~\ref{thm:gcrb_sample_bound} (proved in Section~\ref{apx:scb}) is based on a bound for the precision matrix \cite{kereta2021estimating} and properties of the score vector. This result shows that the deviation of the eGCRB from the GCRB is bounded in terms of the norm of the  {generator score vector}  the GCRB itself, and the number of samples.
Importantly, Theorem~\ref{thm:gcrb_sample_bound} shows that the error decreases (at the best possible rate) as the number $m$ of samples increases.

{ 
\subsection{Convergence of eGCRB} To study the convergence of the eGCRB to the true CRB, we first bound the the relative error of
{approximating $\crb_{\R}$ by $\overline{\mathrm{GCRB}_{\G}}\brackets{\p}$} due to both empirical mean and learning errors using  Theorems~\ref{thm:gcrb_sample_bound} and \ref{thm:fim_pd} and Corollary~\ref{cor:learning_error}. Then we discuss the conditions under which the eGCRB convergence to the true CRB. 



{ 

\begin{corollary} \label{cor:total_error}
Suppose  the assumptions
in 
{
Theorem~
\ref{thm:fim_pd} and
Assumptions~\ref{assum:trusted_region},~\ref{assum:bound_score_true_no_trim} hold , and
$\kappa(\crb_{\R}\brackets{\p}) \frac{\eta(\theta)}{\| \F_{\R}\brackets{\p}\|} < 1$.
}
Then 
$\norm{\hat{\F}_{\G}\brackets{\p}^{-1}}\leq \mathrm{C_G}\brackets{\p}$ and there exist absolute constants $C_1, C_2>0$ such that provided that $m>C_1\brackets{1+u}\mathrm{C_G}\brackets{\p}^2$, for any $u>0$ we have, with probability at least $1-\exp\brackets{-u}$: 
    \begin{equation}\label{eq:relative_error}
    \begin{split}
        &\mathrm{RE}\brackets{\p} \triangleq
        \frac{\norm{\overline{\mathrm{GCRB}_{\G}}\brackets{\p}-\crb_{\R}\brackets{\p}}}{\norm{\crb_{\R}\brackets{\p}}}\\& \leq\norm{\hat{\F}_{\G}\brackets{\p}^{-1}}\squareb{\tilde{\mathrm{B_s}}+\eta\brackets{\p}},
    \end{split}
	\end{equation}
	where
	\begin{subequations}
{	\begin{equation}
	    \mathrm{C_G}\brackets{\p}\triangleq
	    \frac{\|\crb_{\R}\brackets{\p}\|}{1-\kappa(\crb_{\R}\brackets{\p}) \frac{\eta(\theta)}{\| \F_{\R}\brackets{\p}\|} },
	    \end{equation}
	    }
	\begin{equation}\label{eq:b_s_tilde}
	    \tilde{\mathrm{B}}_s\brackets{\p}\triangleq C_2\frac{\norm{\hat{\F}_{\G}\brackets{\p}^{-1}}}{\norm{\F_{\R}\brackets{\p}^{-1}}}\mathrm{C_s}\brackets{\p}^2\sqrt{\frac{1+u}{m}}.
	\end{equation}
	\end{subequations}

\end{corollary}


{
In Corollary~\ref{cor:total_error} (proved in Section~\ref{proof:total_error}), 
we observe that the relative error in approximating the CRB using the proposed approach decreases with decreasing $FIM$ learning error and increasing number of samples used to compute the empirical mean in the evaluation of the GFIM. 
Similar to the case of Corollary~\ref{cor:learning_error}, the interpretation of the result is facilitated by considering the case of  normalized learning error  bounded by $\kappa_{\R} \frac{\eta(\theta)}{\| \F_{\R}\brackets{\p}\|} <0.5$, where $\kappa_{\R} \triangleq\kappa(\crb_{\R}\brackets{\p})$. (This is only slightly more stringent than the requirement in Corollary~\ref{cor:total_error}.) Then, as shown in Section~\ref{proof:total_error}, the following exact bound 
holds. 
\begin{equation} \label{eq:Relative_CRB_Error_Simple}
   \mathrm{RE}\brackets{\p}  \leq  \kappa_{\R} \left(4 C_2 \frac{C_{\R}^2\brackets{\p}}{\|\F_{\R}\brackets{\p}\|} \sqrt{\frac{1+u}{m}} + \frac{\eta(\theta)}{\| \F_{\R}\brackets{\p}\|} \right)
\end{equation}
The bound on the relative error in the eGCRB in \eqref{eq:Relative_CRB_Error_Simple} (which is proved in Appendix~\ref{proof:simple_re_crb}), is dimensionless, and shows clearly the effect of the condition number $\kappa_{\R}$ of the CRB, the number of samples used to compute the empirical mean, and the normalized FIM learning error.
}  
}

The eGCRB relative error \eqref{eq:Relative_CRB_Error_Simple} consist of two terms. The first is the sampling error, which can be  made arbitrarily small by using a large enough $m$. The second is term is the learning error, which we address next.
{\begin{assumption}\label{ass:well-trained}[Existence of well-trained generator] Let $\mathcal{G}$ be the set of all generators representable by the chosen architecture of the normalizing flow network,  and define $\G^{*}$ to be an optimal generator in the sense that if $\RGamma^{*}\brackets{\p}=\G^{*}\brackets{\Z;\p}$ then
$\RGamma^{*}\brackets{\p}$
 is distributed the same as the measurement distribution $\R$, that is, $\RGamma^{*}\brackets{\p}\sim\probt{\g;\p}{\R}\quad\forall\p \in \Theta$. Then we assume that:
\begin{equation}\label{eq:generator_exist}
\G^{*}\in\mathcal{G},
\end{equation}
and the dataset $\mathcal{D}$ is rich enough such that the training is successful and results in $
\G=\G^{*}$ which yields:
\begin{subequations}\label{eq:generator_convergne}
\begin{equation}
    \mathrm{TV}\brackets{p_{\R},p_{\RGamma};\p}=0, 
\end{equation}
\begin{equation}
    \mathrm{I}_\mathrm{F}\brackets{p_{\RGamma},p_{\R};\p}=0.
\end{equation}
\end{subequations}
\end{assumption}
}

Two conditions are required for a well-trained generator (Assumption \ref{ass:well-trained}) to be realizable:
(i) the set of generators $\mathcal{G}$ representable by the chosen
architecture of the normalizing flow network contains the optimal generator $\G^{*}$ \eqref{eq:generator_exist}; and (ii) the generator can be trained to achieve this approximation using the training data. 

Assuming that Condition (i) holds, then Condition (ii) can be satisified, i.e.,  a well-trained generator is realizable on a trusted region (Assumption \ref{assum:trusted_region}) in the limit of infinite training data set if Assumptions \ref{sas:support} and \ref{sas:connected_set} on the measurement distribution $\probt{\vectorsym{r};\p}{\R}$ and Assumption \ref{sas:theta_bound} on the training set distribution are satisfied.
}  {Recall that $\mathrm{TV}\brackets{p_{\R},p_{\RGamma};\p}$ includes the trusted region truncation error, which must vanish for $\mathrm{TV}\brackets{p_{\R},p_{\RGamma};\p}=0$. This happens automatically when the true measurement distribution is bounded $\hat{\Upsilon}=\Upsilon$,  or thanks to the  
 proposed adaptive trimming criterion in the limit of infinite training set, $\epsilon_r\xrightarrow{} 0$ as $|\mathcal{D}| \xrightarrow{} \infty$.
}

 {Moreover, Condition (i) can be addressed in several ways. Available prior knowledge of the problem can be incorporated into the chosen
architecture of the normalizing flow (e.g., NoiseFlow\cite{abdelhamed2019noise}, SineFlow\cite{habi2022generative}) to help satisfy  
Condition (i). Because the very notion of parameter estimation requires some modeling of the measurments, such prior knowledge is typically available in parameter estimation problems.
Furthermore, following the standard practice in deep learning, one can increase the representation power of the network by increasing its size and number of trainable parameters. In the extreme case of no domain knowledge, this involves reliance on the ability of NF to provide a universal approximation.} 

However, the question of universal approximations is an active research area, with recent results \cite{papamakarios2021normalizing,lee2021universal,verine2021expressivity,kong2020expressive,kong2021universal} showing that for certain architectural choices and under some additional assumptions, normalizing flows can provide universal approximations with arbitrarily small error. As the currently available universal approximation conditions are sufficient conditions, we expect that ongoing research will result in further relaxation of the conditions and a larger variety of architectural choices.

	 {Finally, we combine {Corollary~\ref{cor:total_error} (equivalently, \eqref{eq:Relative_CRB_Error_Simple})}
	with Assumption~\ref{ass:well-trained} to} 
	 state that if $\G$ is well-trained, then the eGCRB  {converges almost surely} to the data CRB.  
	\begin{theorem}\label{thm:gcrb_conv_no_trim}
	\begin{equation*}
	    \overline{\mathrm{GCRB}_{\G}}\brackets{\p}\xrightarrow{\substack{m\xrightarrow{}\infty}}\crb_{\R}\brackets{\p} \quad \text{a.s}
	\end{equation*}
	\end{theorem}
	\begin{proof}
	 {By Assumption~\ref{ass:well-trained} $\eta(\p)=0$, so that by Corollary~\ref{cor:total_error} the eGCRB converges to the CRB as $m \rightarrow \infty$. To establish the type of convergence, 
	note that
	$\mathbb{E} \left[ \overline{\F_{\mathrm{G}}} \right]
	=  \hat{\F}_{\G}\brackets{\p} 
$
}
	By the strong Law of Large numbers  {$\overline{\F_{\mathrm{G}}}\brackets{\p}$ in \eqref{eq:gfim} converges to its expected value,} $\lim\limits_{m\xrightarrow{}\infty}\overline{\F_{\mathrm{G}}}\brackets{\p}=\hat{\F}_{\G}\brackets{\p} \quad \text{a.s}$. %
 {Finally, by Theorem~\ref{thm:fim_pd} for  $\eta(\p)=0$ we have $\hat{\F}_{\G}\brackets{\p}  
={\F}_{\R}\brackets{\p}$}.
Inverting yields the result. 
	\end{proof}
It follows that if  Assumptions~\ref{assum:bound_score_true_no_trim} and \ref{ass:well-trained} hold then the eGCRB converges to the CRB almost surely as $m\xrightarrow{}\infty$.
\section{Normalizing Flows}\label{sec:nf}
We use a normalizing flow \cite{kobyzev2020normalizing,papamakarios2021normalizing}, a class of (invertiable) neural networks to obtain $\G$ and $\GI$. Here, we give a brief overview of the normalizing flows utilized in this paper. Specifically, we will present conditional normalizing flow (CNF) where the normalizing flow is conditioned on the input parameter $\p$. A CNF 
transforms a random variable with a known distribution (typically Normal) through a sequence of differentiable, invertible mappings. Formally, let $\vectorsym{z}_1,...,\vectorsym{z}_{n_l}$ be a sequence of random variables that are related as  $\vectorsym{z}_i=G_{i}\brackets{\vectorsym{z}_{i-1};\p}$, where for each $\p\in \Ps$ the function $G_{i}\brackets{\cdot;\p}:\mathbb{R}^d\xrightarrow{}\mathbb{R}^d$ is a differentiable and bijective, $n_l$ is the number of flow layers, and $\vectorsym{z}=\vectorsym{z}_0$ a random variable with a known and tractable probability density function $p_{\Z}:\mathbb{R}^d\xrightarrow{}\mathbb{R}$.  Then
defining $\Gamma \triangleq \G\brackets{\vectorsym{z}_0;\p}=G_{n_l}\circ G_{n_l-1} \circ ... \circ G_{1}\brackets{\vectorsym{z}_0;\p}$ as a composition of the $G_i$, the transformation of a random variables formula says that the probability density function for $\Gamma$ is
\begin{equation}
\label{eq:likelihood}
\begin{split}
 \probt{\vectorsym{\gamma};\p}{\Gamma}
 &=\probt{\GI\brackets{\vectorsym{\gamma};\p}}{\Z}\abs{\mathrm{det}\matsym{J}_{\GI}\brackets{\vectorsym{\gamma};\p}}, \\   
 &=\probt{\GI\brackets{\vectorsym{\gamma};\p}}{\Z}\prod_{j=1}^{n_l}\abs{\mathrm{det}\matsym{J}_{j}\brackets{\z_j;\p}},
\end{split}
\end{equation}
where for each fixed $\p$, $\GI=\GI_1\circ \GI_2\circ ...\circ \GI_{n_l}$ and $\GI_i$ are  the inverses of $\G$ and of $G_i$ with respect to their first argument and $\matsym{J}_{j}\brackets{\z_{j-1};\p}
=\frac{\partial \GI_{j}\brackets{\z_{j-1};\p}}{\partial \z_{j-1}}$ is the Jacobian of the $j^{th}$ transformation $\GI_j$ with respect to its input $\z_{j-1}$.  We denote the
value of the $j^{th}$ intermediate flow as $\z_j=\G_j\circ\cdots\circ\G_1\brackets{\z_0;\p}=\GI_{j+1}\circ\cdots\circ\GI_{n_l}\brackets{\g;\p}$ and $\g=\z_{n_l}$.

\bigskip
\noindent
\textbf{Density Learning} A CNF
can be directly used for density learning by finding parameters that minimize the negative log-likelihood (NLL) over a set of samples where the likelihood is given by \eqref{eq:likelihood}. Given a dataset $\mathcal{D}$ (see  Problem \ref{problem:crb}) 
and the transformations $G_1,...,G_{n_l}$ parameterized by $\Omega=(\omega_1,...,\omega_{n_l})$ respectively, the negative log-likelihood is given by:
\begin{equation}\label{eq:nnl_cond}
    \begin{split}
        &L\brackets{\Omega}=-\sum_{i}\log\brackets{\probt{\GI\brackets{\vectorsym{r}_i;\p_i|\Omega}}{\Z}}\\
        &-\sum_{i}\sum_{j}^{n_l}\log\brackets{\abs{\mathrm{det}\matsym{J}_{j}\brackets{ \vectorsym{u}_{ij};\p_i|\Omega}}}.
    \end{split}
\end{equation}
where  $\vectorsym{u}_{ij}=\GI_{j+1}\circ\cdots\circ\GI_{n_l}\brackets{\vectorsym{r}_i;\p}$ denotes the intermediate flow of the $i^{th}$ sample and the $j^{th}$ layer.
Note that the first term is the negative log-likelihood of the sample under the base measure (latent distribution) and the second term is a differential volume correction, which accounts for the change of differential volume induced by the transformations. 

We use a CNF based on the Glow\cite{NEURIPS2018_d139db6a} architecture, which includes the following flow steps: Activation Normalization, Affine Coupling, and so-called 1x1 convolution (an invertible matrix operation). These flow steps transport the base distribution into the target distribution. However, we take the SRFlow approach \cite{lugmayr2020srflow} for the insertion of the conditioning parameter using the Affine Inject flow step that modifies the transformation according to the conditional parameter $\p$. Furthermore, in some cases (e.g., in the non-Gaussian measurement example of Sec.~\ref{sec:multiplcation}),
a more complex modification of the base distribution is required, and this is achieved by replacing the Affine Coupling with a Cubic Spline Coupling flow \cite{durkan2019cubic}. The flow steps mentioned above are detailed in Appendix \ref{apx:flow_layers}. 

\section{Measurements Model Examples}\label{sec:example_models}
First, we present two simple examples in which we can compute both the CRB and GCRB analytically and obtain an optimal generator. Note that by "optimal" we mean that the generator distribution $\probt{\vectorsym{r};\p}{\Gamma}$ is identical to the data distribution $\probt{\vectorsym{r};\p}{\R}$, meaning that 
{Assumption}~\ref{ass:well-trained} holds {with $\G=\G^{*}$}. Then in the second part, we present a real-world measurement model of cameras, which will be used to demonstrate some of the benefits of the GCRB.
\subsection{Simple Measurement Models}\label{sec:example_models_simple}
\subsubsection{Linear Gaussian}\label{sec:linear}
Let 
\begin{equation} \label{eq:linmod}
    \R\brackets{\p}=\matsym{A}\p+\vectorsym{v},
\end{equation}
where matrix $\matsym{A}\in\mathbb{R}^{d\times k}$ with $d>k$ 
 and $\vectorsym{v}\sim \normaldis{0}{\matsym{C}_{vv}}$ is an additive zero-mean Gaussian noise with positive-definite covariance $\matsym{C}_{vv}\in\mathbb{R}^{d\times d}$. 
Then  the measurement is distributed as  $\R\brackets{\p}\sim \normaldis{\matsym{A}\p}{\matsym{C}_{vv}}$, which provides a complete description of the measurement channel. 
The CRB for $\p$ coincides with the expression for the covariance of the linear unbiased minimum variance estimator and is given by  \cite{kay1993fundamentals} : $\mathrm{CRB}_\R\brackets{\p}=\squareb{\matsym{A}^T
    \matsym{C}_{vv}
    ^{-1}\matsym{A}}^{-1}.$

Now we present an optimal generator for this example. Let $\G\brackets{\vectorsym{z};\p}=\matsym{A}\p+\matsym{L}\vectorsym{z}$, where $\vectorsym{z} \sim N(0,\matsym{I} )$ and $\matsym{L}$ is a square root (e.g, the Cholesky) factor of $\matsym{C}_{vv}$, that is, $\matsym{L}\matsym{L}^T=\matsym{C}_{vv}$. Then, as easily verified, $\G$ is
an optimal generator, because  for any $\p$
the distribution $\G\brackets{\vectorsym{z}} \sim \normaldis{\matsym{A}\p}{\matsym{C}_{vv}}$ coincides with that of $\R\brackets{\p}$. The inverse function of $\G$, the normalizing flow, is $\GI\brackets{\g;\p}=\matsym{L}^{-1}\brackets{\g-\matsym{A}\p}$. 
{We compute the score vector of th optimal generator and normalizing flow in Appendix~\ref{sec:linear_score}, which yields }
\begin{equation}\label{eq:score_linear}
    \vectorsym{s}_{\p}\brackets{\vectorsym{z}}=-\matsym{A}^T\brackets{\matsym{L}^{-1}}^T\z,
\end{equation}
and the GFIM corresponding to the optimal generator obtained using  \eqref{eq:generator_fim} is
    $\mathrm{F}_{\G}\brackets{\p}=\expectation{\matsym{A}^T\brackets{\matsym{L}^{-1}}^T\Z\Z^T\matsym{L}^{-1}\matsym{A}}{\Z}.$
Simplifying and taking the inverse results in $\mathrm{F}_{\G}\brackets{\p}^{-1}=\mathrm{GCRB}_{\G}\brackets{\p} =\mathrm{CRB}_R\brackets{\p}$. This confirms, that as expected, an optimal generator will yield the same CRB on the parameter vector $\p$ as the correct distribution.

\subsubsection{Scale Non-Gaussian}\label{sec:multiplcation}
Here, we show a scale model with non-Gaussian distribution. Consider the data model 
\begin{equation}\label{eq:example_mul}
    r=y\theta,
\end{equation}
where $\theta\in\mathbb{R}^+$ is the desired parameter and $y$ is a random variable with the PDF 
\begin{equation} \label{eq:PDFY-Scale}
    \prob{y}=\frac{1}{\sqrt{2\pi\sigma^2}}3y^2\cdot\exp\brackets{-\frac{1}{2\sigma^2}{y}^{6}}.
\end{equation}
Then, as shown in Appendix~\ref{sec:vm_crm}, the FIM of $\theta$ is  $\F_{\R}\brackets{\p}=18\theta^{-2}$ and 
   $ \mathrm{CRB}_R\brackets{\theta}=\frac{\theta^2}{18}$.
We show in Appendix~\ref{apx:scale_opt_gen} that  $\Gamma(\theta) = 
\G\brackets{z;\theta}=\theta z^{\frac{1}{3}}$ is an optimal generator, that is $\Gamma(\theta) \sim \prob{r; \theta}$. 
The inverse function of the optimal generator is the normalizing flow $\GI\brackets{\gamma;\theta}=\brackets{\frac{\gamma}{\theta}}^{3}$. {We compute the score vector of the
optimal generator and normalizing flow in Appendix~\ref{sec:scale_score_vec}, which yields }

\begin{equation}\label{eq:score_scale}
    \vectorsym{s}_{\p}\brackets{\vectorsym{z}}=-\theta^{-1}3\brackets{1-z^2}.
\end{equation} 
Finally, the FIM of the optimal generator obtained using  \eqref{eq:generator_fim} is
    $\mathrm{F}_{\G}\brackets{\p}=\expectation{\brackets{\theta^{-1}3\brackets{1-Z^2}}^2}{Z}=18\theta^{-2},$
where the last equality follows by the Gaussian moment property.
Hence $\mathrm{GCRB}_{\G}\brackets{\theta} =\mathrm{CRB}_R\brackets{\theta}$.
This example demonstrates that an optimal generator yields the correct CRB on the parameter in a non-Gaussian case.

\subsection{Image Processing}
We consider two classical image processing problems, however with a real-world learned measurement model of 4-channel (RGGB) color image sensors using NoiseFlow \cite{abdelhamed2019noise}, a normalizing flow that models camera noise. Using this learned model, we obtain the GCRB for these two problems. 
\subsubsection{Image Denoising}
Image denoising is a well-known problem in signal processing, however modeling camera noise is a challenging task \cite{foi2008practical,zhang2017improved}. Due to the difficulty of noise modeling, it is impossible to compute an analytical lower bound on denoising performance on a realistic model. 

Formally the denoising problem is defined as follows. Denoting by $\matsym{H}$ a clean 4-channel (RGGB) image patch and by $\matsym{V}$ the camera noise, the noisy image tensor is defined as
\begin{equation}\label{eq:denoising}
    \tilde{\matsym{H}}=\matsym{H}+\matsym{V}.
\end{equation}
Our goal is to provide a lower bound on the performance of any unbiased estimator  that estimates the clean image $\matsym{H}$ from the noisy image $\tilde{\matsym{H}}$. Under this model, denoting by $\vecop{\matsym{T}}$ the vectorization of  tensor $\matsym{T}$,  $\R=\vecop{\tilde{\matsym{H}}}$ is the measurement vector  and $\p=\vecop{\matsym{H}}$ is the parameter vector.

\subsubsection{Edge Detection}\label{sec:edge_detection} 
Another interesting image processing task is edge detection. Here we describe the edge model  used in this work. Consider a $c$-channel image of width $h$ pixels, and let {$\matsym{H}_{ijc}=f_{ijc}\brackets{\p}$} be a vertical edge function that maps a continuous-{parameter vector $\p=[\theta_p,\theta_w]$ of }edge position  $\theta_p \in [0, h-1]$ { and width $\theta_w\in\mathbb{R}^{+}$}, to the image color values at horizontal and vertical pixel position $i,j$ and image channel $c$. The edge function is specified in terms of a 
horizontal color scaling function {$s_i\brackets{\p}:([0,h-1],\mathbb{R}^{+})\xrightarrow{}[0,1]$} as
\begin{equation*}
    {f_{ijc}\brackets{\p}=\brackets{p^h_c-p^l_c}\cdot s_{i}\brackets{\p}+p^l_c},
\end{equation*}
where $p^h$ and $p^l$ are the vectors of RGGB pixel values for high and low intensities, respectively. The color scaling function is defined as
\begin{equation*}
    {s_{i}\brackets{\p}=\phi\brackets{\frac{\theta_p-i}{\theta_w}}},
\end{equation*}
where $\phi$ is the Sigmoid function $\phi\brackets{x}=\frac{1}{1+\exp{\brackets{x}}}$. 
Images with  edges of different position and width following the model above are shown in Fig.~\ref{fig:edge_detection}.
\begin{figure}[H]
    \centering
    \includegraphics[width=0.25\textwidth]{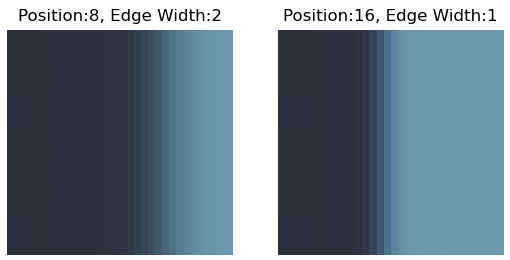}
    \caption{{ Edges in a clean image}
    }
    \label{fig:edge_detection}
\end{figure} 

In this example we want to estimate the edge position $\theta_p$ from a noisy image $\tilde{\matsym{H}}$.  We compare the GCRB with NoiseFlow to the CRB derived for two well-known analytical Gaussian noise models: (i) i.i.d, or white Gaussian noise (WGN); and (ii)  independent noise  with image-dependent intensity - the so-called noise level function (NLF) noise model. The WGN  model ($\matsym{V}_{ijc}\sim\normaldis{0}{\sigma^2}$) with i.i.d noise in each channel of each pixel and variance $\sigma^2$  
has CRB

{\begin{equation}\label{eq:crb_edge_g}
    \mathrm{CRB}_{\mathrm{W}}\brackets{\p}=\frac{\sigma^2 \theta_w^2\brackets{\sum_i   \matsym{M}_i\brackets{\p}}^{-1}}{h\norm{\vectorsym{p}^h-\vectorsym{p}^l}^2_2},
\end{equation}}
{where $\matsym{M}_i\brackets{\p}=s_{i}^2\brackets{\p}\brackets{1-s_{i}\brackets{\p}}^2\begin{bmatrix}
    1& -\frac{\theta_p-i}{\theta_w}\\
     -\frac{\theta_p-i}{\theta_w} &  \frac{\brackets{\theta_p-i}^2}{\theta_w^2}
    \end{bmatrix}.$}


The NLF model with $\matsym{V}_{ijc}\sim\normaldis{0}{\alpha^2f_{ijc}\brackets{{\p}}+\delta^2}$, where $\alpha$ and $\delta$ are the noise parameters, has CRB
{\begin{equation}\label{eq:crb_edge_nlf}
\begin{split}
    &\mathrm{CRB}_{\mathrm{NLF}}\brackets{\p}=\\
    &\brackets{\sum_{i,j,c}\frac{\brackets{p^h_c-p^l_c}^2\matsym{M}_i\brackets{\p}}{\brackets{\alpha^2f_{ijc}\brackets{\p}+\delta^2}^2\theta_w^2}\brackets{\alpha^2f_{ijc}\brackets{\p}+\delta^2+\frac{\alpha^2}{2}}}^{-1}.
\end{split}
\end{equation}}
A detailed calculation of CRB's is given in Appendix \ref{app:crb_edge_detection}.

\subsubsection{Camera Noise Model}\label{sec:camera_noise_model}
 Several recent works \cite{abdelhamed2019noise,chen2018image} have used a data-driven approach to model camera noise. We use NoiseFlow \cite{abdelhamed2019noise} to model   a realistic camera noise $\matsym{V}\sim \probt{\vectorsym{v};\matsym{H}}{\matsym{V}}$ and similarly to model the noisy image $\tilde{\matsym{H}}\sim\probt{\tilde{\vectorsym{h}};\matsym{H}}{\tilde{\matsym{H}}}$. To obtain a noisy image flow, we cascade to NoiseFlow an AdditiveNoise Flow layer corresponding to  \eqref{eq:denoising}, defined as
\begin{equation}\label{eq:additive_noise_flow_step}
    \vectorsym{z}_{n+1}=\matsym{H}+\vectorsym{z}_{n}.
\end{equation}
The inverse of  \eqref{eq:additive_noise_flow_step} is  given by $\vectorsym{z}_{n}=\vectorsym{z}_{n+1}-\matsym{H}$ and the log-determinant term is zero. Note that the ability to incorporate a signal model into the normalizing flow 
is a well-known advantage, which has also been exploited in NoiseFlow \cite{abdelhamed2019noise}.
 
 NoiseFlow is trained  using the Smartphone Image Denoising Dataset (SIDD) \cite{abdelhamed2018high}.  The SSID dataset consists of 150 
 noisy and corresponding clean images captured in ten different scenes, with five smartphone cameras of different brands, under several lighting conditions and ISO (sensitivity) levels. Specifically, NoiseFlow is trained on $h_p=32 \times w_p=32$ pixel RGGB patches  of clean  image and noise $\matsym{H},\matsym{V} \in\mathbb{R}^{h_p\times w_p \times4}$.

 Fig.~\ref{fig:clean_noisy_images} shows examples of clean images and the corresponding noisy images generated by NoiseFlow at different ISO levels and for different camera devices. They illustrate the strong ISO, device, and image dependence of the noise, which cannot be captured by an analytical model, thus precluding traditional calculation of estimation bounds. Instead,  using this learned model, we obtain the GCRB for the two problems of image denoising and edge position detection.
 \begin{figure}
   \centering
    \includegraphics[height=15mm,width=0.49\textwidth]{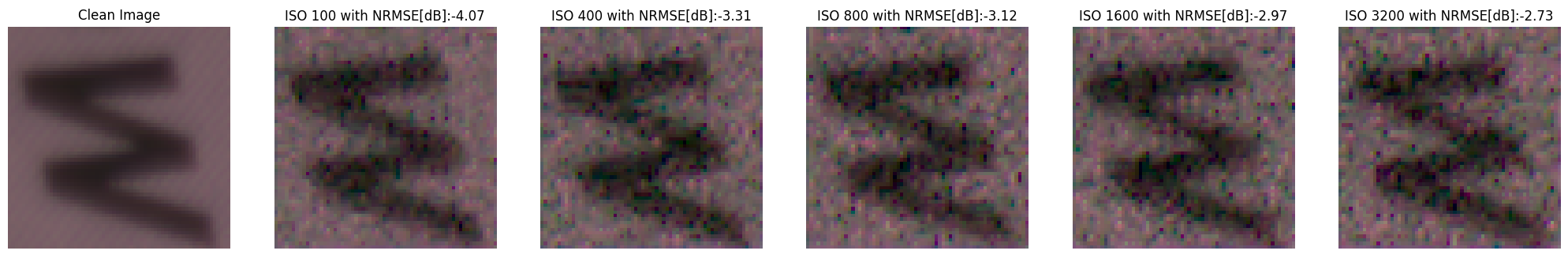}
    \\
    (a) \\
    \includegraphics[height=15mm,width=0.49\textwidth]{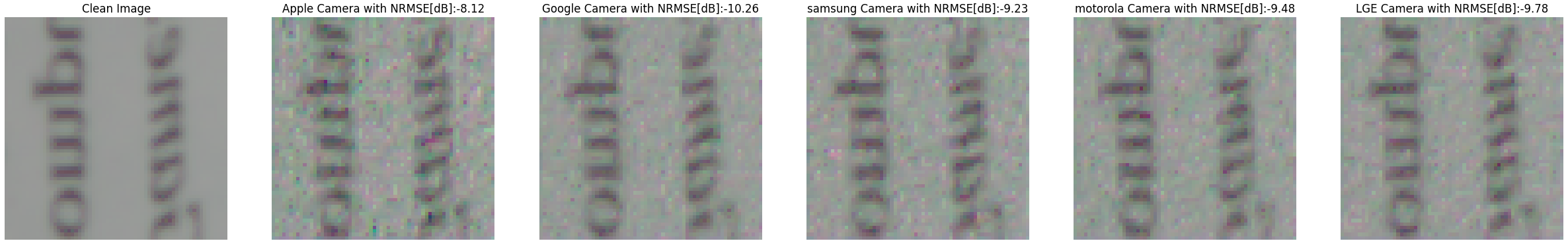}
    \\ (b)
    \caption{NoiseFlow output: clean and noisy images.  (a)~Different ISO levels, for Camera Type=zero (Apple).  (b)~Different cameras at ISO level 100.}
    \label{fig:clean_noisy_images}
\end{figure}

\section{Experimental Results}\label{sec:experimental}
This section presents a set of numerical experiments for assessing, analyzing, and demonstrating the GCRB. In the first set of experiments, we determine the quality of the approximation provided by the GCRB by evaluating the eGCRB on the examples in Section \ref{sec:example_models_simple} and comparing to the true, analytically derived CRB. In the second set of experiments, we study, for the linear estimation problem, the approximation error of the eGCRB due to imperfect training, and due to the use of the sample mean to estimate the expected value. For the last two experiments, we present the usage of GCRB on the real-world examples of image denoising and edge detection in a device-dependent noise. Unless stated otherwise, we evaluate the eGCRB using $m=64K$ generated samples in the sample mean in all experiments. In all experiments, the computation is done using Nvidia 1080Ti GPU running white PyTorch \cite{paszke2019pytorch}. 

\subsection{Approximation Quality}
We evaluate the  accuracy of the approximation to the CRB provided by the GCRB  using two kinds of normalizing flows: (i) the optimal flow, which satisfies the condition of perfectly matched distribution {as $\G^{*}$ in Assumption~\ref{ass:well-trained}} 
; and (ii) a standard/learned normalizing flow (see Sec.~\ref{sec:nf}), which is trained using the dataset $\mathcal{D}$. 

Unless stated otherwise, we use the following parameters in the training process of all experiments
For training a normalizing flow, we use the conditional negative log-likelihood (NLL) of the training set \eqref{eq:nnl_cond} as the loss function. We train each normalizing flow using a dataset of $200k$ samples for 90 epochs with batch size 64. We use the Adam optimizer \cite{kingma2014adam} with learning rate $1e-4$ and parameters $\beta_1=0.9$ and $\beta_2=0.999$. At the end of  training, we obtain the learned normalizing flow $\GI$ and it's inverse (the generator) $\G$ 
and evaluate the eGCRB at several $\p$ values using Algorithm \ref{alg:sampling}. The latent variable is  chosen to be $\z\sim\normaldis{0}{\matsym{I}}$ in all examples. We begin by showing the approximation quality in the two examples presented in Sec.~\ref{sec:example_models_simple}, and then investigate the source of approximation error.

\bigskip
\noindent
\textbf{Linear Example} For the linear model \eqref{eq:linmod} we use  $d=8$, $\sigma_v=2.0$ and $k=2$. Hence $\p \in \mathbb{R}^2$ and $R(\p_i) \in \mathbb{R}^8$.
We generate the training dataset $\mathcal{D}$ in the following manner. First, we generate  matrices $\matsym{A}$ and $\matsym{L}$
using a standard normal distribution,  
and use the same two matrices to generate all the samples in $\mathcal{D}$. For each sample $(\p_i, R(\p_i)) \in \mathcal{D}$, the parameter vector $\p_i \in \mathbb{R}^2$ is drawn i.i.d from a uniform distribution $\p_i\sim U[-2,2]^2$,  $\vectorsym{v}_i \in \mathbb{R}^8$ is drawn i.i.d Normal $\vectorsym{v}_i\sim \normaldis{0}{\matsym{C}_{vv}}$ 
with $\matsym{C}_{\vectorsym{v}\vectorsym{v}}=\matsym{L}\matsym{L}^T$, 
and $R(\p_i) \in \mathbb{R}^8$ is computed using \eqref{eq:linmod}.
Using this dataset, we train a normalizing flow with the architecture  shown in Appendix~\ref{apx_net_linear}, obtaining $\GI$ and  $\G$. 

We chose an architecture with invertible $1x1$ convolution and affine inject since it can represent an optimal generator and satisfies the $\G\in C^2$ condition.

Then, for each value of $\p$ of interest, we use the trained generator to generate samples of the score vector, and compute the eGFIM using \eqref{eq:gfim}, which yields, upon inversion, the eGCRB. For comparison, we repeat the generation of the score vector  using the optimal normalizing flow and generator instead of the learned flow and generator. 

In Fig.~\ref{fig:theta_values_linear}  we display the traces of the two eGCRBs, as well as that of the analytical CRB, for $\p = (\xi, \xi)^T$, with $\xi$ on a uniform grid on the interval [-2,2].
\begin{figure}
    \centering
    \includegraphics[scale=0.38]{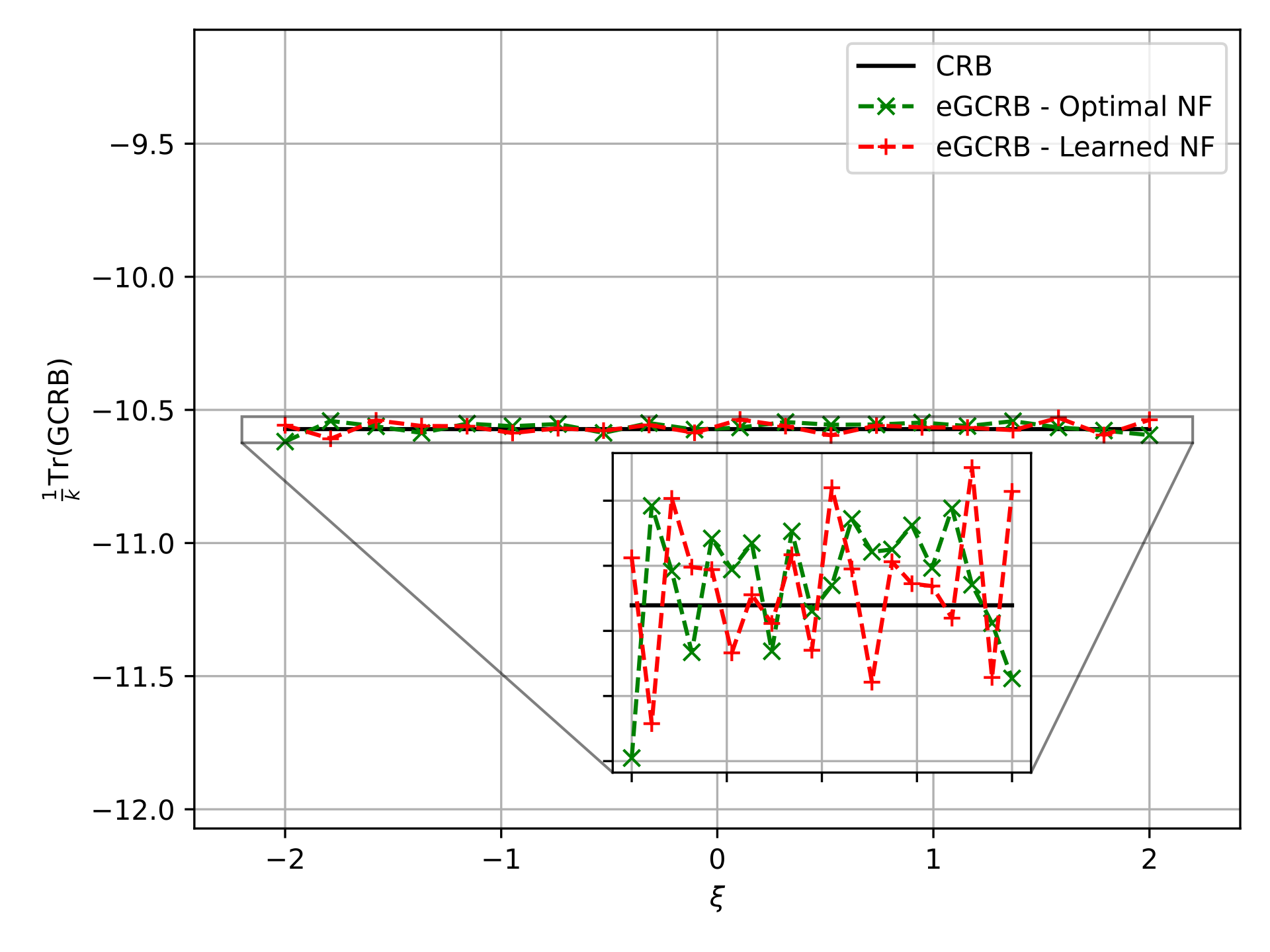}
    \caption{Trace of CRB and eGCRB for the linear measurement model \eqref{eq:linmod}.
    }
    \label{fig:theta_values_linear}
\end{figure} 
Figure \ref{fig:theta_values_linear} shows that a learned normalizing flow can estimate the true CRB to a good accuracy. Because (as we verifed) the specific parameter values $\p$ shown in Fig.~\ref{fig:theta_values_linear} are not present in the training set, 
this also demonstrates that the GCRB works well for unseen examples.
 As expected, on the average the eGCRB using the optimal flow has slightly better accuracy  
than the one using the learned flow, because the former only suffers from the finite sampling error in estimating the GFIM using an empirical mean, whereas the latter is also subject to the imperfectly learned flow model. 
The relative error \eqref{eq:relative_error} of the eGCRB is displayed in Figure \ref{fig:theta_re} for both learned and optimal flows, showing that both have comparable accuracy, of within $\approx 0.5$\% from the true CRB.
\begin{figure}
    \centering
    \includegraphics[scale=0.38]{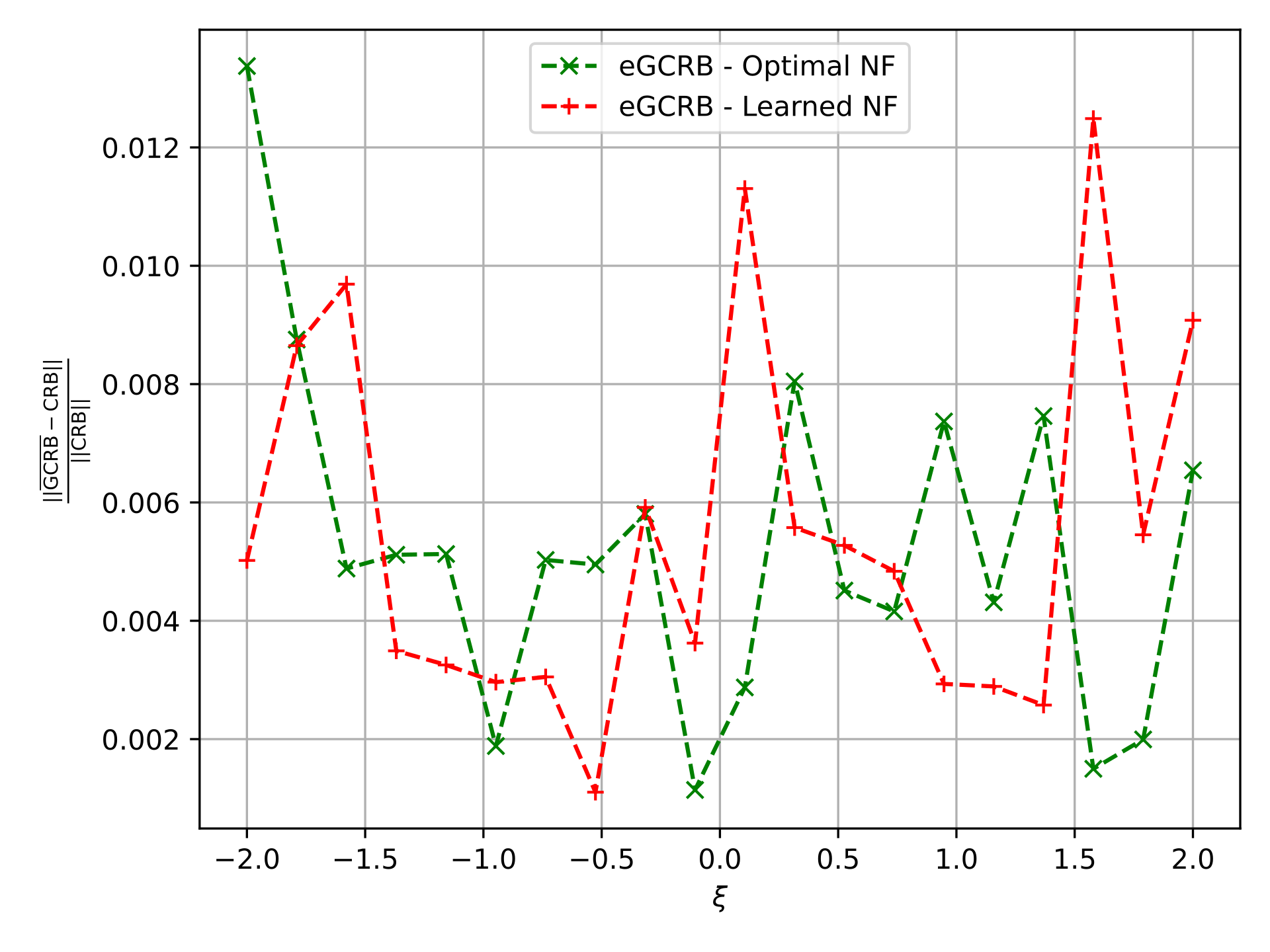}
    \caption{Relative error  between the eGCRB and CRB for the linear measurement model \eqref{eq:linmod} at $\p = (\xi, \xi)^T$ as a function of $\xi$. The eGCRBs obtained using  an optimal and learned flow are compared.
    }
    \label{fig:theta_re}
\end{figure}

\bigskip
\noindent
\textbf{Scale Example} 
In this example we generate the training dataset $\mathcal{D}$ in the following manner. 
For each sample $(\theta_i, R(\theta_i)) \in \mathcal{D}$, the parameter $\theta_i$ is drawn i.i.d from a uniform distribution $\theta_i\sim U[3,6]$,  $y$ is drawn i.i.d $y \sim p_Y$, where $p_Y$ is given by \eqref{eq:PDFY-Scale}  and $R(\theta_i) = y\theta_i$, per \eqref{eq:example_mul}. 
To produce a vector input to the normalizing flow, as needed for the application of affine coupling, 
we define a vector measurement of length 2, composed of two  i.i.d samples with the same parameter $\theta_i$.\footnote{We use this form, rather than padding with an unrelated standard normal random variable, to mitigate issues of exploding condition number\cite{lee2021universal}.} Note that since this vector measurement corresponds to two i.i.d measurements, by the additivity property of the FIM, this only scales the resulting GFIM by a factor of 2. 
 Using this dataset $\mathcal{D}$ we train a normalizing flow with the architecture  shown  in Appendix \ref{apx:net}. We chose an architecture with cubic-spline and affine inject since it can locally represent an optimal generator and satisfies the $\G\in C^2$ condition.
 
 Then, we follow the same procedure as for the linear measurement model to produce the eGCRB using both the learned flow and the optimal flow. 
 The two eGCRB values and the true CRB are compared in
  Fig.~\ref{fig:theta_values}.
\begin{figure}
    \centering
    \includegraphics[scale=0.38]{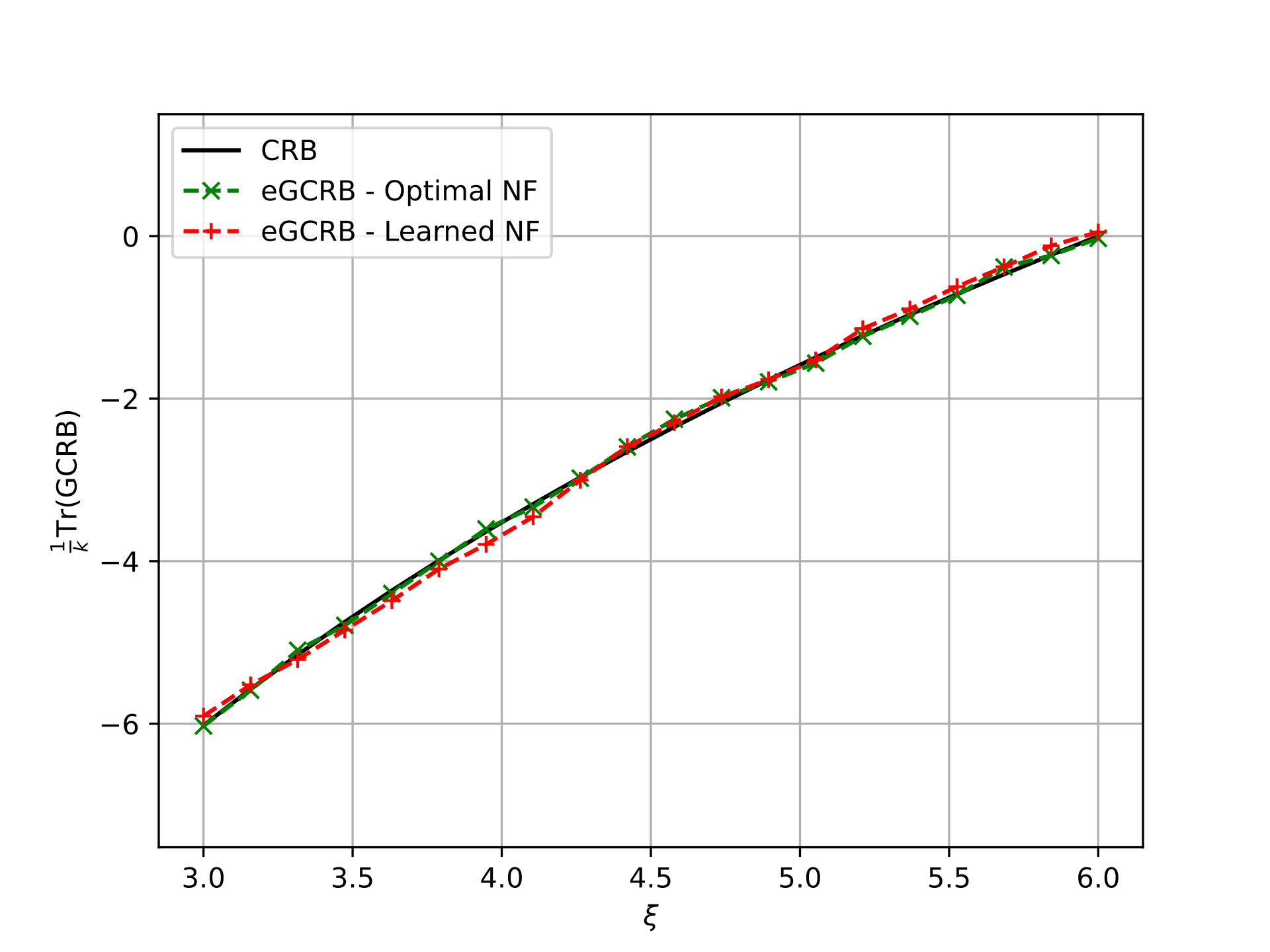}
    \caption{Analytical CRB and eGCRBs for the scale model \eqref{eq:example_mul},\eqref{eq:PDFY-Scale},   using  the optimal flow and a learned flow.}
    \label{fig:theta_values}
\end{figure}
Figure \ref{fig:theta_values} demonstrates that a learned  normalizing flow can estimate the true CRB in the non-Gaussian case, with accuracy comparable to that of the optimal flow. Similar to the linear example, because (as we verified)  the parameter values $\theta$ used to plot  Fig~\ref{fig:theta_values}  are  not present in the training set, this again demonstrates the interpolation capability of the GCRB to provide a good approximation for unseen examples.

To summarize, Fig.~\ref{fig:theta_values} and Fig.~\ref{fig:theta_values_linear} demonstrate Theorem~\ref{thm:fim_pd}, which states that given a well-trained generative model {(Assumption~\ref{ass:well-trained})} the GCRB {approximates well} the CRB. Both figures show random deviations of the eGCRB from the CRB due to two reasons: (i) imperfectly trained generative model; and (ii) a finite number of samples used to calcuate empirical mean, as stated in Theorem~\ref{thm:gcrb_sample_bound}.  Moreover, the eGCRBs in Fig.~\ref{fig:theta_values_linear} and Fig.~\ref{fig:theta_values} are evaluated at points that are not present in the training set, which shows the ability of GCRB to interpolate the CRB values to those points.

\subsection{Error Analysis}
Here, we study further the approximation error due to the empirical mean and imperfect training.  We use two metrics for the error:
\begin{subequations}
\begin{equation*}
    \mathrm{MRE}=\max\limits_{\p\in\Theta_{T}}\mathrm{RE}\brackets{\p},
\end{equation*}
\begin{equation*}
    \overline{\mathrm{MRE}}=\frac{1}{ |\Theta_{T}|}\sum_{\p\in\Theta_{T}}\mathrm{RE}\brackets{\p},
\end{equation*}
\end{subequations}
where $\mathrm{MRE}$ and $\overline{\mathrm{MRE}}$  are the maximal and mean relative norm error, respectively, $\Theta_{T}\subset\Theta$ is the set of $\p$ value used in the validation process, and $
|{\Theta_{T}}|$ is the cardinality of the set $\Theta_{T}$. In this experiment, we verify Theorem \ref{thm:gcrb_sample_bound} using the linear optimal model with the same parameters as above and evaluate eGCRB with different  number of samples $m \in \{64K,128K,256K,512K\}$. We repeat the evaluation for each $m$ 2000 times and and present the histogram of the relative norm error $\mathrm{RE}\brackets{\p}$ in Figure \ref{fig:hist_gfim}. In all the trials we use the same parameter vector $\p=(0.2,0.2)^T$.
\begin{figure}
    \centering
    \includegraphics[scale=0.38]{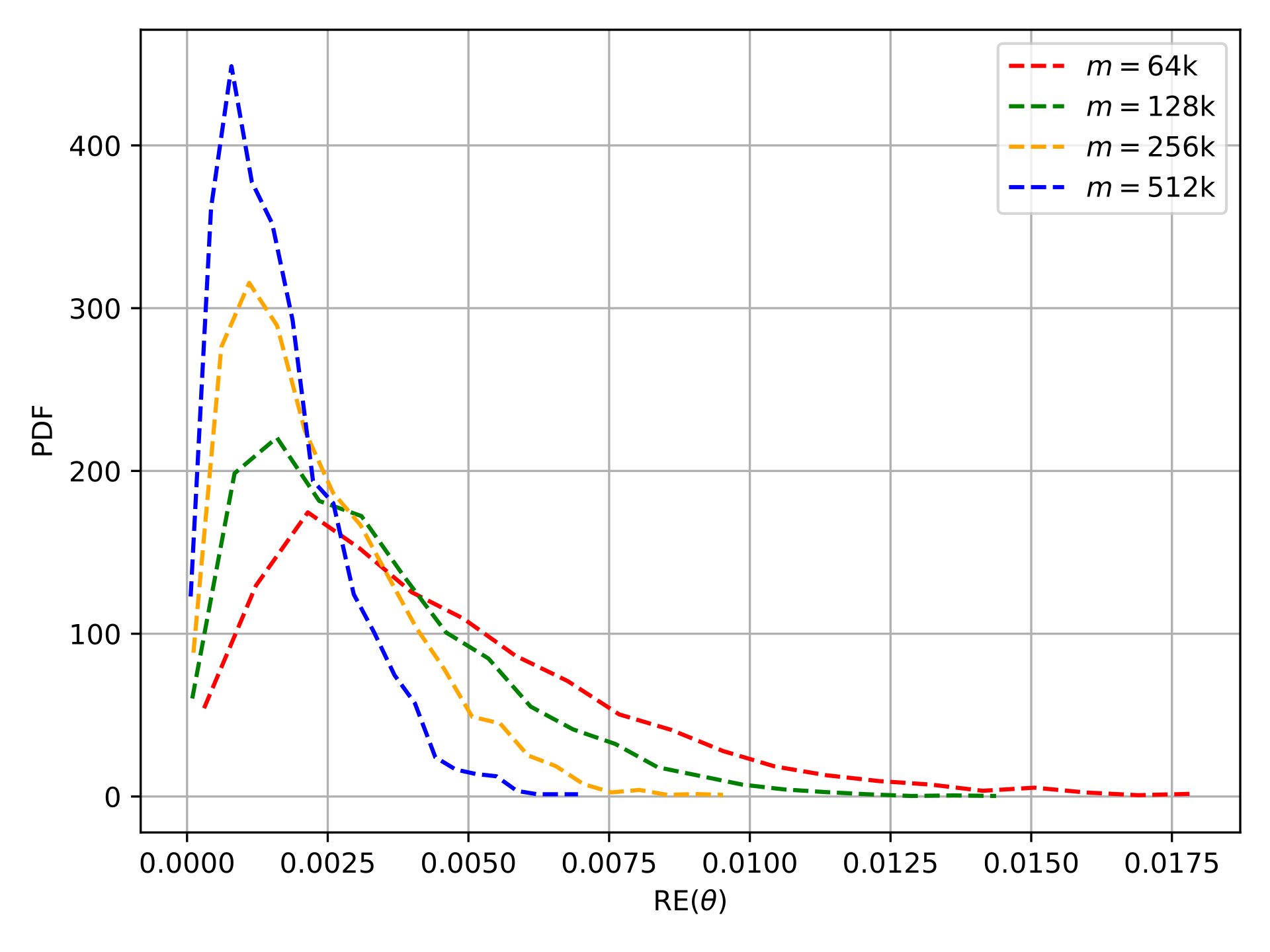}
    \caption{Histogram of  $\mathrm{RE}\brackets{\p}$ for the linear example  using different number of samples   $ m \in \{64k,128k,256k,512k\}$. 
    }
    \label{fig:hist_gfim}
\end{figure}

We see in Fig.~\ref{fig:hist_gfim} the effect of different $m$ values on the distribution of the relative error. This confirms, that as predicted by Theorem~\ref{thm:gcrb_sample_bound}, for the optimal generator, we can make the eGCRB error arbitrary small by increasing the number of samples $m$ to calculate the eGFIM  \eqref{eq:gfim}. This  addresses the error due to sampling assuming a well-trained normalizing flow.

In the next experiment, we address the error due to imperfect training. To focus on this aspect, we set $m=512k$, so that the error due to the empirical mean is negligible.
We train a normalizing flow on the linear problem using various dataset sizes, and report the maximal and mean relative error. To train a normalizing flow with a small dataset size, we adjust the number of epochs to have a constant number of gradient updates, by setting  the number of epochs to $\ceil{90\frac{200e^3}{\abs{\mathcal{D}}}}$ where $\abs{\mathcal{D}}$ is the size of the  dataset and $\ceil{x}$ denotes the ceiling of $x$. In Fig.~\ref{fig:dataset_size_vs_mre} we present the $\mathrm{MRE}$ for different dataset size and validation parameter set $\Theta$  consisting of 20 points $\p = (\xi, \xi)^T$ with $\xi\in [-2 , 2]$ uniformly spaced.
\begin{figure}
    \centering
    \includegraphics[scale=0.38]{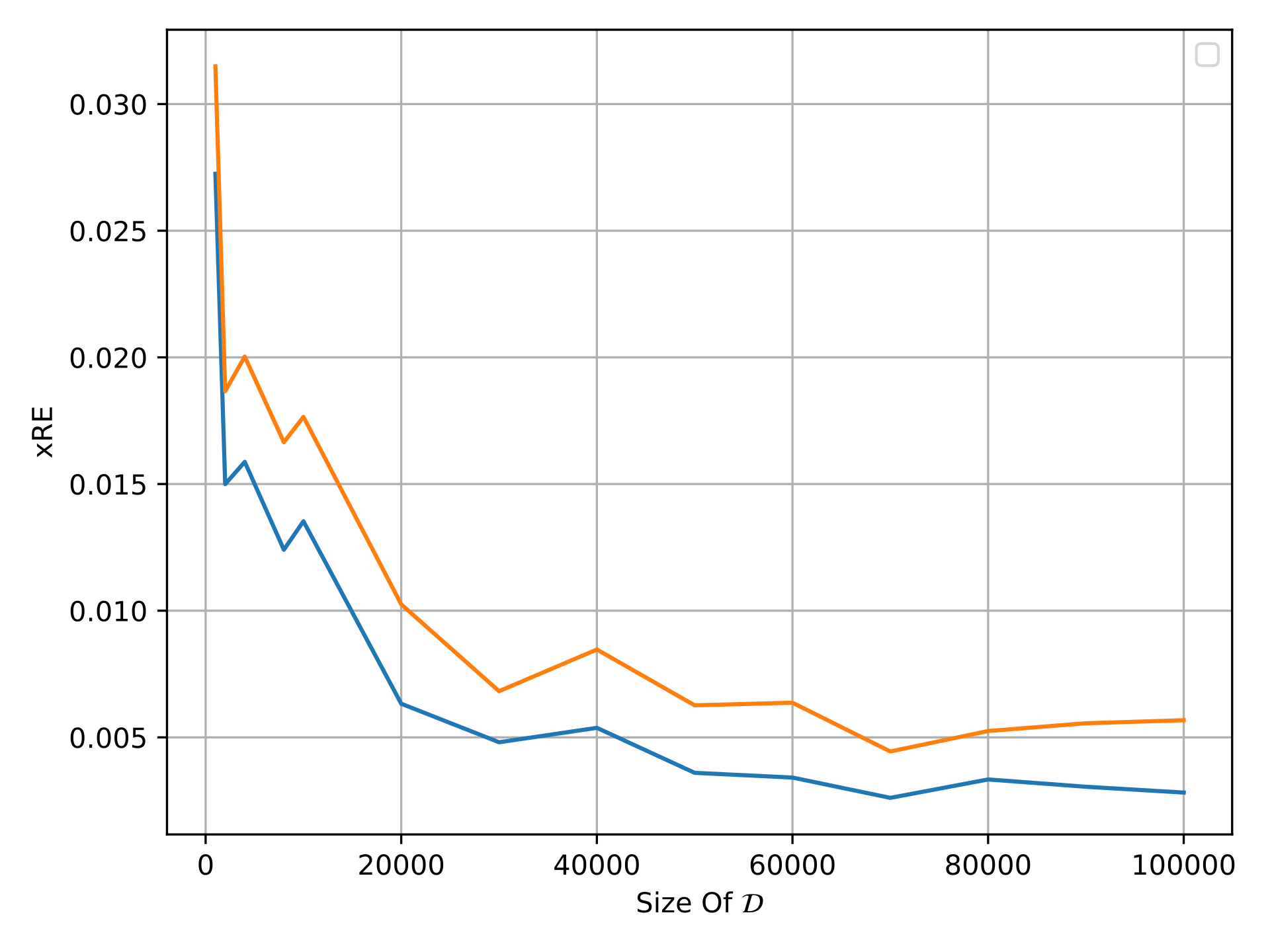}
    \caption{$\mathrm{MRE}$ and $\overline{\mathrm{MRE}}$ for the linear measurement model example  vs. training dataset size.
    }
    \label{fig:dataset_size_vs_mre}
\end{figure}

Fig.~\ref{fig:dataset_size_vs_mre} shows  relative error decreasing with increasing training set size, highlighting the importance of training  to obtain a well-trained generative model. However, increasing the dataset size beyond some point  (in this case, above 50K samples) doesn't improve the results. To interpret this saturation effect, recall that the normalizing flow used in this problem can represent the optimal generator, hence a limitation in representation capacity by the normalizing flow is not the culprit. Instead,   comparison with Fig.~\ref{fig:hist_gfim} suggests that for $|\mathcal{D}| > 5 \cdot 10^4$ the error in this learned generator experiment is  dominated by the empirical mean error - with similar values as when using an optimal generator.


\subsection{Image Denoising}
Here, we study the GCRB for the Image denoising problem,   using the NoiseFlow 
\cite{abdelhamed2019noise} to model the image noise. 
Noise Flow is composed of affine coupling, invertible $1 \times 1$ convolution, gain layer (which is a constant affine transformation), and signal-dependent layer (which is similar to affine inject with a predefined function). 
First, we replace the  ReLU activation function in the original NoiseFlow to Sigmoid Linear Unit (SiLU) \cite{hendrycks2016gaussian},  to  satisfy the requirement that $\G\in C^2$. Then, we train the modified Noise-Flow from scratch using the authors' original code\cite{noise_flow_git_hub}. We obtained NNL of $-3.528$ which is a similarl  to that in the original NoiseFlow.
Then, we add to NoiseFlow an AdditiveNoise layer (see Section \ref{sec:camera_noise_model}), 
which provides a noisy image. 
We generate all the required derivatives using the PyTorch built-in symbolic differentiation invoked using standard PyTorch commands, and utilize our eGCRB computing forumulas to provide an approximation to the CRB. 
We compute the eGCRB using \eqref{eq:gfim} with $m=64k$. 

Note that  for image visualization purposes only, we render RGGB images through a color processing pipeline into sRGB color space. Moreover, to visualize the eGCRB, we extract the diagonal of the eGCRB matrix and present the lower bound for each channels (R, G, G, B) separately. In all experiments we use an image 
size of $h=w=32$  and batch-size 32. 

We present the bounds on the denoising performance in normalized form, by computing the following metrics

\begin{figure}
    \centering
    \includegraphics[width=0.48\textwidth,height=0.46\textheight]{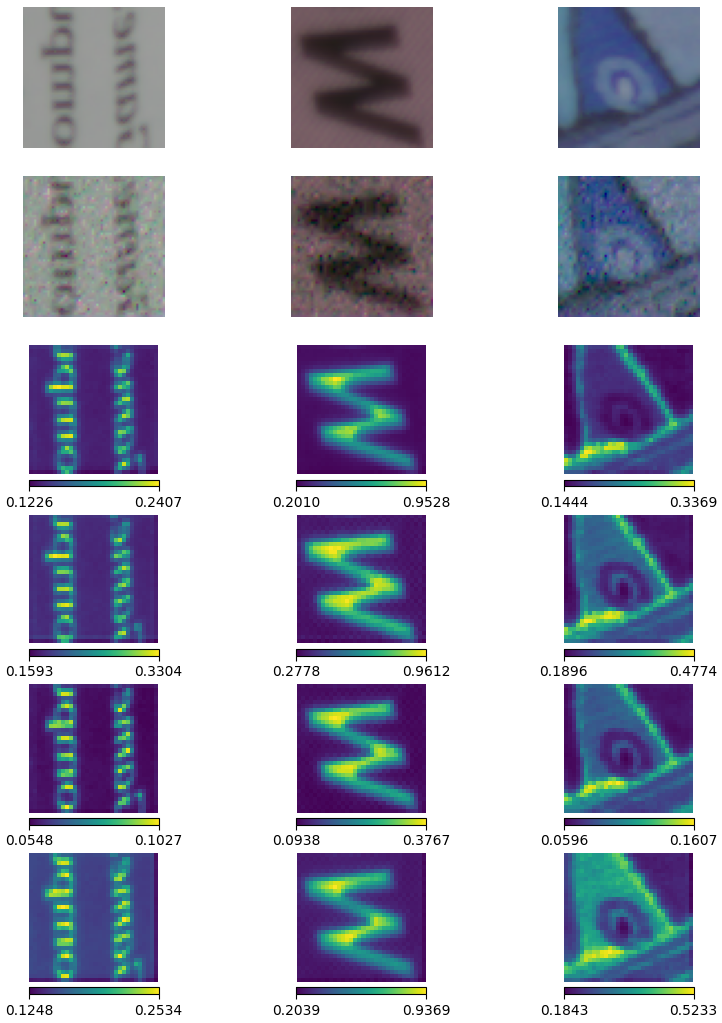}
    \caption{GCRB for the denoising problem on three scenes (one per column). 1st and 2nd rows: clean and noisy image, respectively. Last four last rows: $\mathrm{NPMSE}$ of the R, G, G, B channels.  Camera type zero (Apple).  
}
    \label{fig:gcrb_denoisg_exmaple}
\end{figure}

\begin{subequations}\label{eq:denoising_metrics}
\begin{equation} \label{eq:NPRMSE}
    \mathrm{NPRMSE}_{ijc} \triangleq\frac{\sqrt{g\brackets{\mathrm{diag}\brackets{{\overline{\mathrm{GCRB}}}}}_{ijc}}}{I_{ijc}},
\end{equation}
\begin{equation} \label{eq:NRMSE}
    \mathrm{NRMSE}\triangleq\sqrt{\frac{1}{hwc}\sum_{i,j,c}
    \mathrm{NPMSE}_{ijc}^2}.
\end{equation}
\end{subequations}
 The $\mathrm{NPRMSE}_{ijc}$ is the normalized per pixel bound on the error standard deviation error, 
$\mathrm{diag}\brackets{\matsym{A}}$ denotes the diagonal vector of matrix $\matsym{A}$, $g\brackets{}:\mathbb{R}^{h\cdot w\cdot c}\xrightarrow{}{}\mathbb{R}^{h\times w\times c}$ is the reshaping of vector to RGGB image, and $i,j,c$ are the vertical, horizontal and channel indices, respectively. In turn, $\mathrm{NRMSE}$ is the square of the per pixel NPRMSE, averaged over
 the entire image followed by a square root. 
 
Note that the metrics  in \eqref{eq:denoising_metrics} take into account only the diagonal elements of the eGCRB. However, both the eGFIM and the eGCRB have off-diagonal elements thanks to the ability of NoiseFlow to generate correlated noise that models the sensor. This correlation is evident in the non-zero off-diagonals in the eGFIM, and affects the diagonal elements of the eGCRB.
 

In this study, we perform several experiments. First, we
present several visual examples in Fig.~\ref{fig:gcrb_denoisg_exmaple}. The top row shows three clean images to which we refer, from left to right, as scene one, two, and three. The second row shows the corresponding noisy versions for Camera 0 (Apple) at ISO = 100. 
The next four rows display (as images color-coded by magnitude) the  normalized lower bound on the denoising error standard deviation for each pixel (i.e., the $NPRMSE_{ijc}$): one row for each of the four channels $c=1,2,3,4$. We observe that pixels with different colors have distinct lower bounds, showing the effect of the Signal Depend Layer  in NoiseFlow\cite{abdelhamed2019noise}. An analogous   behavior is seen in the NFL model which scales the noise by the clean image values. 
It is also seen that brighter pixels have a better (smaller) normalized lower bound than darker ones. 

Next, in Fig.\ref{fig:gcrb_denoisg_iso} we plot the normalized lower bound on the denoising performance using Device=0 (Apple) on the same three scenes as in Fig.~\ref{fig:gcrb_denoisg_exmaple}. It is seen that a lower ISO level allows better denoising than a higher one, and that different scenes  have different denoising lower bounds. For an insight as to whether the difference is due to color level, scene structure, or both, we refer to Fig.~\ref{fig:clean_noisy_images}. It reveals that the noise level increases with ISO level, and is relatively higher in darker (lower color level) areas. The first property clearly accounts for the increase in the bound vs. ISO level seen in Fig.\ref{fig:gcrb_denoisg_iso}, whereas the second property explains the relative ranking of the bounds for the three scenes, with increasing normalized bound for darker images.    

\begin{figure}
     \centering
     \begin{subfigure}[b]{0.23\textwidth}
    \centering
    \includegraphics[width=\textwidth]{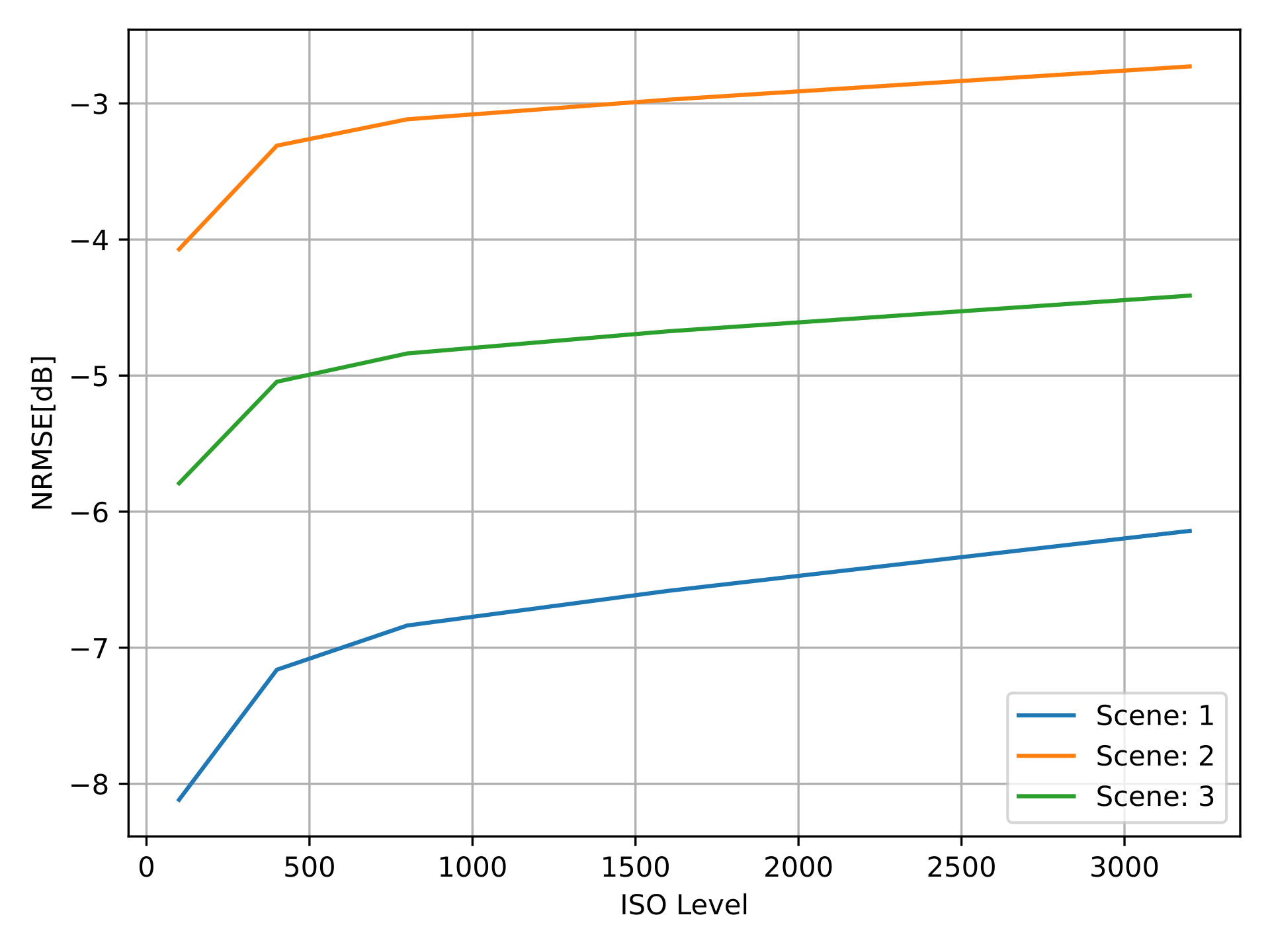}
    \caption{Scene}
    \label{fig:gcrb_denoisg_iso}
    \end{subfigure}
        \begin{subfigure}[b]{0.23\textwidth}
    \centering
    \includegraphics[width=\textwidth]{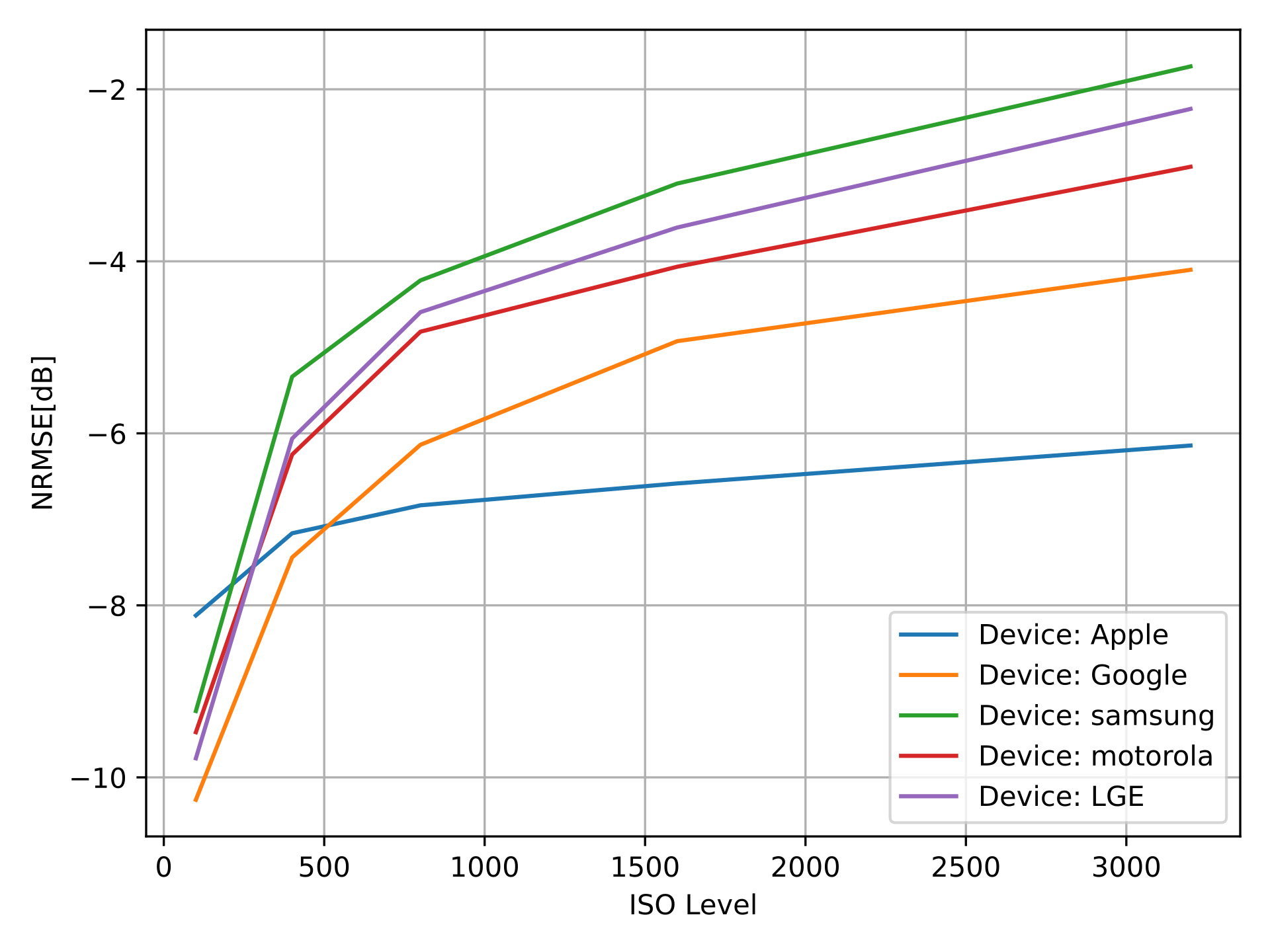}
    \caption{Devices}
    \label{fig:gcrb_denoisg_device}
    \end{subfigure}
    
    \caption{\small {Lower bound on the denoising error vs. ISO levels. In Fig.~\ref{fig:gcrb_denoisg_iso} the lines correspond to Scene 1 - Scene 3 on the first row of Fig.~\ref{fig:gcrb_denoisg_exmaple} and in Fig~\ref{fig:gcrb_denoisg_device} the lines represent different devices in Scene 1 of Fig. \ref{fig:gcrb_denoisg_exmaple}.}}
\end{figure}



Figure \ref{fig:gcrb_denoisg_device} shows, for Scene 1, the effect of different measurement devices. The relative ranking in terms of the denoising bound cannot be inferred from the visual impression of the noise for the different devices in Fig.~\ref{fig:clean_noisy_images}. However, it remains the same for Scene 2 and Scene 3, showing consistency of the bounds for each device.  
These results demonstrates a unique advantage of the  GCRB, which can provide a bound specific to a measurement device.

\subsection{Edge Detection}
We use the same parameters as used in the image denoising problem. First, in Fig.~\ref{fig:edge_position}, we present a lower bound on edge position estimate  vs.  the position of the edge in the image, for several different edge widths, using Device=0 (Apple) and ISO level 100.
\begin{figure}
    \centering
    \includegraphics[scale=0.42]{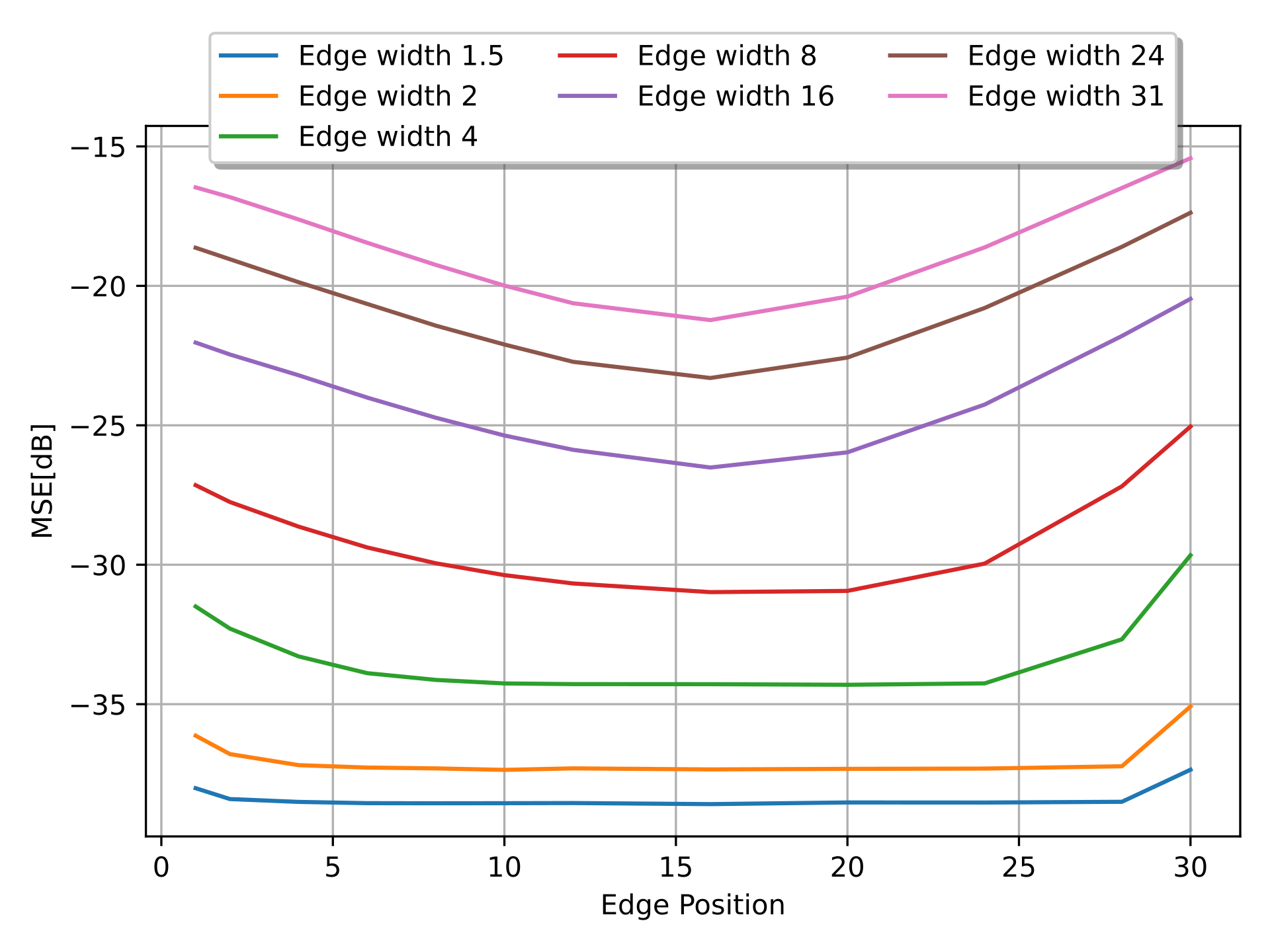}
    \caption{Lower bound on edge position with different edge widths using Device=0 (Apple) at ISO 100. 
    }
    \label{fig:edge_position}
\end{figure}
Fig.~\ref{fig:edge_position} reveals different behavior of the edge localization bound for different edge widths. First, the bound increases with increasing edge width. This is not surprising, since a smooth edge can be expected to be harder localize in the presence of noise than a sharp edge, and  this also agrees with the dependence of the CRB \eqref{eq:crb_edge_g} on edge width for standard Gaussian noise. Second, for small edge width, the bound shows little dependence on position at the center of the image, but increases slightly when the edge approaches the boundaries of the image. This can be attributed to the truncation of some of the edge transition when the edge approaches the image boundaries. Third, for larger edge widths, the bound shows an asymmetric dependence on the edge position relative to the image center. This too can be explained by truncation of one side of the edge transition: however, because the noise is signal-dependent, the effect of truncating the bright side of the edge is opposite to that of truncating the dark side. Moreover, a similar effect of edge position is observed in the analytical CRB \eqref{eq:crb_edge_nlf} for the NLF noise model, which also has signal-dependent noise level. 

To further demonstrate the advantages of the GCRB,   we investigate in the next experiment the ability of a generative model to capture the complex measurements distribution and produce an accurate lower bound. To this end,  we compare, in the context of the edge detection problem,  the three noise models: WGN, NLF, and Noise-Flow. We do so for Device=2 (Samsung) 
(Fig. \ref{fig:clean_noisy_images} at ISO level 100.
For a quantitatively meaningful comparison, we set the  parameters $\sigma^2$, $\alpha$, and $\delta$ of the analytical noise models to the maximum likelihood estimates obtained from
the noisy images that were used to train NoiseFlow\cite{abdelhamed2019noise}. These noisy images are taken from the base SSID dataset\cite{abdelhamed2018high}, and were preprocessed as in  NoiseFlow. 
\begin{figure}
     \centering
     \begin{subfigure}[b]{0.23\textwidth}
    \centering
    \includegraphics[width=\textwidth]{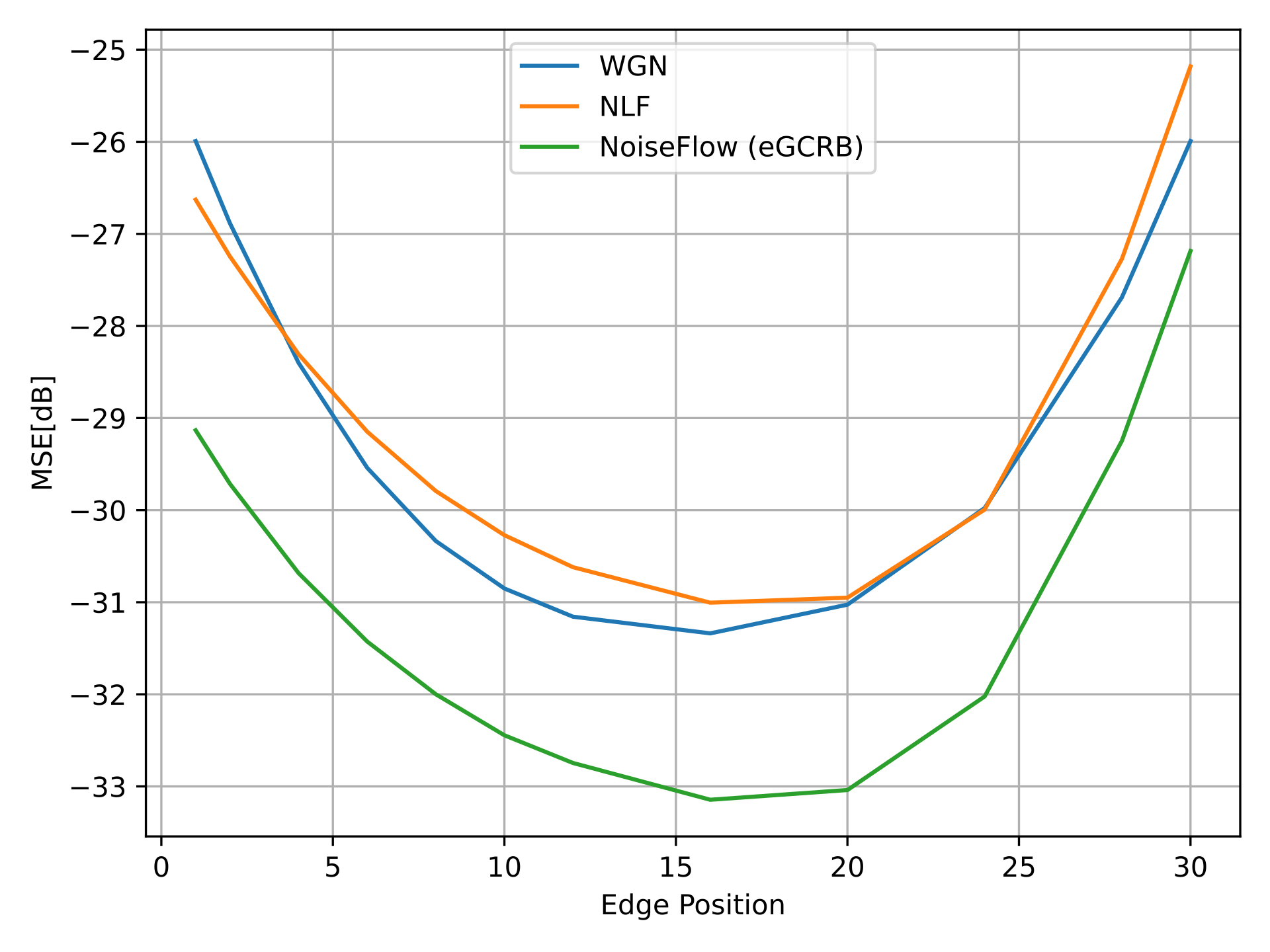}
    \caption{Position}
    \label{fig:edge_position_models_compare}
    \end{subfigure}
        \begin{subfigure}[b]{0.23\textwidth}
    \centering
    \includegraphics[width=\textwidth]{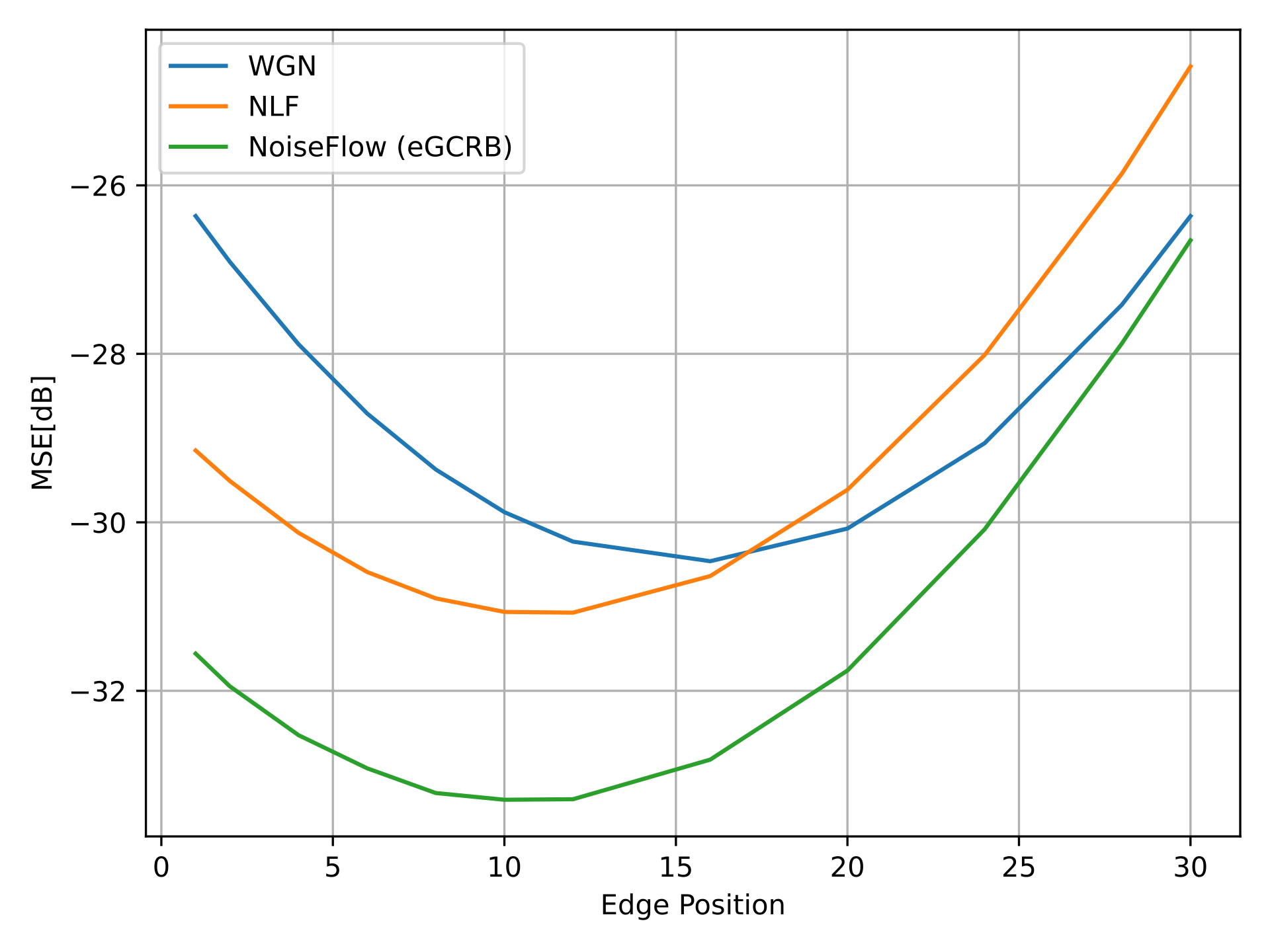}
    \caption{Width}
    \label{fig:edge_width_models_compare}
    \end{subfigure}
    
    \caption{Lower bound on variances of the edge {detection parameters (Position \ref{fig:edge_position_models_compare} and Width \ref{fig:edge_width_models_compare}) over different edge positions,}  using different measurement noise models for Device=2 at ISO 100 and edge width 8 pixels.}
    \label{fig:edge_model_compare}
\end{figure}
 
 Fig.~\ref{fig:edge_model_compare} shows 
 that (i) the WGN model misses altogether the asymmetric behavior of the bound with respect to edge position; and (ii) both Gaussian noise models have  CRBs larger than the eGCRB. Both (i) and (ii) are to be expected, since the WGN model misses the signal-dependence of the noise, and independent Gaussian noise yields the largest CRB  for given noise variance  \cite{stoica2011gaussian}. Finally, note the subtantial difference 
 between the eGCRB for Device=2 (in Fig.~\ref{fig:edge_model_compare}) and the eGCRB for Device=0 in Fig.~\ref{fig:edge_position} for the same edge width of 8 and ISO 100. This again demonstrates the unique ability of the GCRB to provide device-dependent bounds. 
 Overall, these results illustrate the importance of a learned model to capture the complexity of the measurement distribution and obtain an accurate lower bound. 
\begin{figure}
    \centering
    \includegraphics[scale=0.38]{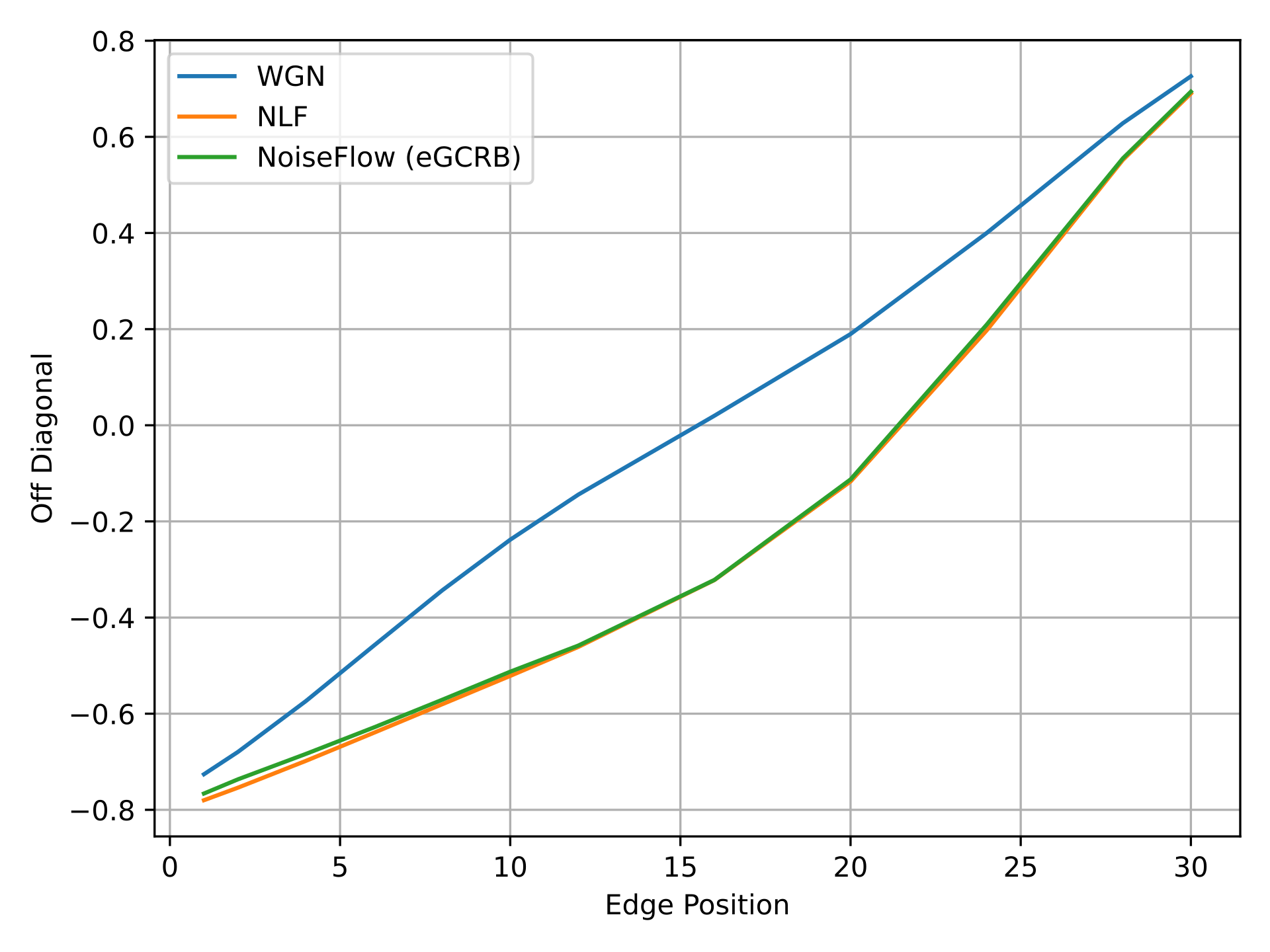}
    \caption{{Normalized off-diagonal elements of the eGCRB (correlation of position and width estimation errors) for the edge detection problem over different edge positions,  using Device=2 at ISO 100.} 
    }
    \label{fig:edge_off_diag}
\end{figure}

{In addition, in Figure \ref{fig:edge_off_diag} we illustrate the ability of the GCRB to study the correlation between estimation errors of different parameters. Specifically, we present the normalized off-diagonal of the eGCRB, namely the Pearson correlation. We observe that whenever the edge is located at the boundaries of the image, $\theta_p=0$ or $\theta_p=31$, there is high correlation between the position and width parameter estimates. This correlation diminishes  for edge position at the center of the image. Moreover, in the center region the NLF and NoiseFlow have a different crossing point, since due to signal depend noise, the point at which the dark and light pixels have the same SNR is shifted .}
\section{Conclusions and Future work}\label{sec:conclusions}
In this paper we use for the first time a generative model to obtain a data-driven estimate of the Cramer-Rao bound, which does not require access to an analytical model of the measurement probability distribution. Specifically, we used a normalizing flow and showed that this generative model provides the same CRB as the measurement distribution if the generator is well-trained. Moreover, we provided an error analysis bounding the inaccuracy due to the use of an empirical mean for the well-trained case, {and the error of the GCRB due to imperfect learning}. We validated the performance of this approach on two simple signal models with known ground-truth CRBs. We also studied the GCRB on two image processing tasks, with a complex learned measurement model. The results demonstrate two advantages of the GCRB: the ability to obtain a highly accurate performance bound for complex measurement distributions without an analytical model; and the ability to obtain a device-specific bound. 

Questions for future research include quantifying the impacts of  limited representation power of the generative model and a limited training data set on the accuracy of the GCRB. {Another direction is to ensure that GCRB is a valid lower bound (rather than a good appromiation to it) by utilizing  methods for error estimation and model selection\cite{aslett2021}. }{ On the practical side, it will be interesting to study some of the many real-world applications that can benefit from this approach, such as direction-of-arrival estimation in sensor arrays with pooly characterized propagation models. 
}
\section{Proofs}\label{sec:proofs}
\subsection{Proof of Lemma~\ref{lemma:bounded_score_vector}}\label{proof:bound_score_vector}
\begin{proof}
  The generated samples $\g = \G(\z; \p) \in \hat{\Upsilon}$ retained 
 after the trimming process correspond to $\z \in \mathcal{Z}$, where the set $\mathcal{Z} \triangleq \G^{-1}(\hat{\Upsilon}; \p)$ is the pre-image of $\hat{\Upsilon}$ under 
 $\G$. Because $\hat{\Upsilon}$ is a compact set in metric space $\mathbb{R}^d$ (Assumption~\ref{sas:bounded_and_connected}) and $\G^{-1}=\GI: \mathbb{R}^d \rightarrow \mathbb{R}^d$ is a continuous mapping, it follows that as the image of a compact set by a continuous mapping, 
  $\mathcal{Z}$ is a compact set. 
  Next, because $\G \in C^2$ and $\G$ is a diffeomorphism wrt to $\z$, it follows that each of the components $\vectorsym{s}_{\p}\brackets{\z}$ is a continuous function of $\z$ on the compact set $\mathcal{Z}$.
  Hence (by pseudocompactness)  $\vectorsym{s}_{\p}\brackets{\vectorsym{z}}$ is bounded componentwise, and thus also in norm: $\norm{\vectorsym{s}_{\p}\brackets{\z}}\leq \mathrm{C_s}\brackets{\p}$.
\end{proof}

 \subsection{Proof of Theorem~\ref{thm:fim_pd}
 } \label{proof:fim_error}
{
We use the following result, proved in Section~\ref{proofa:mcsi} of the Appendix. 
\begin{lemma}[Matrix Cauchy-Schwartz Inequality]\label{lemma:mcsi}
Let $\vectorsym{x},\vectorsym{y}\in\mathbb{R}^n$ be random vectors with correlation matrices $\matsym{R}_{\vectorsym{x}}$ and $\matsym{R}_{\vectorsym{y}}$ and cross correlation $\matsym{R}_{\vectorsym{x}\vectorsym{y}}=\expectation{\vectorsym{x}\vectorsym{y}^T}{}$. Then,
\begin{equation}
    \norm{\matsym{R}_{\vectorsym{x}\vectorsym{y}}}\leq \brackets{\norm{\matsym{R}_{\vectorsym{x}}}\norm{\matsym{R}_{\vectorsym{y}}}}^{1/2}.
\end{equation}
\end{lemma}

For conciseness, in the proof of Theorem~\ref{thm:fim_pd} below we omit the integration variable $\vr$ and the parameter vector $\p$ from integrals. Thus, the PDFs of the true and learned measurement distributions are abbreviated as  $p_{\R}=\probt{\vectorsym{r};\p}{\R}$ and $p_{\Gamma}=\probt{\vectorsym{r};\p}{\Gamma}$, respectively, and the corresponding score vectors
 $\vectorsym{s}_{\R}=\vectorsym{s}_{\R}\brackets{\vr;\p}\triangleq\nabla_{\p}\NLL_{\R}\brackets{\vr;\p}$ and $\vectorsym{s}_{\Gamma}=\vectorsym{s}_{\Gamma}\brackets{\vr;\p}\triangleq\nabla_{\p}\NLL_{\Gamma}\brackets{\vr;\p}$, where  $\NLL_{\R}\brackets{\vr;\p}$  $\NLL_{\Gamma}\brackets{\vr;\p}$ are the corresponding negative log-likelihoods. 
 \begin{proof}
 \begin{align}\label{eq:delta_start}
    \F_{\R}\brackets{\p} & -\hat{\F}_{\G}\brackets{\p} =\F_{\R}\brackets{\p}-\F_{\Gamma}\brackets{\p}\nonumber\\
    & =\int_{\Upsilon\setminus \hat{\Upsilon}} \vectorsym{s}_{\R}\vectorsym{s}_{\R}^Tp_{\R}d\vr +\int_{\hat{\Upsilon}}\vectorsym{s}_{\R}\vectorsym{s}_{\R}^T p_{\R}d\vr-\F_{\Gamma}\brackets{\p}
     \nonumber\\
    &=
    \matsym{P}_1 + \matsym{P}_2 \nonumber \\
     \matsym{P}_1 & \triangleq \int_{\Upsilon\setminus \hat{\Upsilon}} \vectorsym{s}_{\R}\vectorsym{s}_{\R}^T p_{\R}d\vr
    +\int_{\hat{\Upsilon}} \vectorsym{s}_{\R}\vectorsym{s}_{\R}^T \Delta_p d\vr\nonumber \\
    \matsym{P}_2 & \triangleq \int_{\hat{\Upsilon}} \vectorsym{s}_{\R}\vectorsym{s}_{\R}^T p_{\Gamma}d\vr-\F_{\Gamma}\brackets{\p} \\
   \Delta_p & \triangleq p_{\R}-p_{\Gamma}
\end{align}
Because $\probt{\vectorsym{r};\p}{\Gamma}=0\quad\forall\vr\in\Upsilon\setminus \hat{\Upsilon},\p\in\Theta$,
\begin{equation*}
    \matsym{P}_1 
    =\int_{\Upsilon}\vectorsym{s}_{\R}\vectorsym{s}_{\R}^T \Delta_p d\vr+\int_{\Upsilon\setminus \hat{\Upsilon}}\vectorsym{s}_{\R}\vectorsym{s}_{\R}^Tp_{\Gamma}d\vr=\int_{\Upsilon}\vectorsym{s}_{\R}\vectorsym{s}_{\R}^T \Delta_p d\vr
\end{equation*}
which, by Assumption~\ref{assum:bound_score_true_no_trim}, is bounded in terms of the total variation distance as
\begin{equation}\label{eq:p1_final}
    \norm{\matsym{P}_1}\leq 2C_{\R}^2\brackets{\p}\mathrm{TV}\brackets{p_{\R},p_{\Gamma};\p}.
\end{equation}
Turning to $\matsym{P}_2$, we have
\begin{align}\label{eq:part_two}
    \matsym{P}_2
    &=\expectation{\vectorsym{s}_{\R}\vectorsym{s}_{\R}^T-\vectorsym{s}_{\Gamma}\vectorsym{s}_{\Gamma}^T}{\Gamma}\nonumber\\
    &=\expectation{\brackets{\vectorsym{s}_{\Gamma}-\vectorsym{\Delta}_s}\brackets{\vectorsym{s}_{\Gamma}-\vectorsym{\Delta}_s}^T-\vectorsym{s}_{\Gamma}\vectorsym{s}_{\Gamma}^T}{\Gamma}\nonumber\\
    &=\expectation{\vectorsym{\Delta}_s\vectorsym{\Delta}_s^T}{\Gamma}-\expectation{\vectorsym{s}_{\Gamma}\vectorsym{\Delta}_s^T+\vectorsym{\Delta}_s\vectorsym{s}_{\Gamma}^T}{\Gamma},
\end{align}
where $\vectorsym{\Delta}_s\triangleq \vectorsym{s}_{\R}-\vectorsym{s}_{\Gamma}$ is the score difference vector.
Considering the first term in \eqref{eq:part_two}:
\begin{align}\label{eq:p2_part_one}
    &\norm{\expectation{\vectorsym{\Delta}_s\vectorsym{\Delta}_s^T}{\Gamma}}\leq\trace{\expectation{\vectorsym{\Delta}_s\vectorsym{\Delta}_s^T}{\Gamma}}
    =\expectation{\vectorsym{\Delta}_s^T\vectorsym{\Delta}_s}{\Gamma}\nonumber\\&=\expectation{\norm{\vectorsym{\Delta}_s}^2}{\Gamma}\leq\int_{\Upsilon}p_{\Gamma}\norm{\vectorsym{\Delta}_s}^2d\vr\triangleq \mathrm{I}_\mathrm{F}\brackets{p_{\Gamma},p_{\R};\p}.
\end{align}
Next,  applying Lemma \ref{lemma:mcsi} to the norm of the second term in \eqref{eq:part_two}, yields
\begin{align}\label{eq:p2_part_two}
    &\norm{\expectation{\vectorsym{s}_{\Gamma}\vectorsym{\Delta}_s^T+\vectorsym{\Delta}_s\vectorsym{s}_{\Gamma}^T}{\Gamma}}\leq 2\norm{\expectation{\vectorsym{\Delta}_s\vectorsym{s}_{\Gamma}^T}{\Gamma}}\nonumber\\
    &\leq 2\brackets{\norm{\expectation{\vectorsym{s}_{\Gamma}\vectorsym{s}_{\Gamma}^T}{\Gamma}}\norm{\expectation{\vectorsym{\Delta}_s\vectorsym{\Delta}_s^T}{\Gamma}}}^{1/2}\nonumber\\
    &=2\brackets{\norm{\hat{\F}_{\G}\brackets{\p}}\mathrm{I}_\mathrm{F}\brackets{p_{\Gamma},p_{\R};\p}}^{1/2}.
\end{align}
Now combining \eqref{eq:p2_part_one} and \eqref{eq:p2_part_two} yields:

\begin{equation}\label{eq:p2_final}
    \norm{\matsym{P}_2}\leq 2\brackets{\norm{\hat{\F}_{\G}\brackets{\p}}\mathrm{I}_\mathrm{F}\brackets{p_{\Gamma},p_{\R};\p}}^{1/2} +\mathrm{I}_\mathrm{F}\brackets{p_{\Gamma},p_{\R};\p}.
\end{equation}
In the last step we combine \eqref{eq:p1_final} and \eqref{eq:p2_final}, which yields Theorem~\ref{thm:fim_pd}.
\end{proof}}
\subsection{Proof of Corollary~\ref{cor:learning_error}}\label{proof:well_cond}
{\begin{proof}
By definition \eqref{eq:hat_f_g}, we have $\hat{\F}_{\G}\brackets{\p}\succeq 0$. Combining with \eqref{eq:fim_error_bound}, we have $\lambda_{\min} \left( \hat{\F}_{\G}\brackets{\p}\right) \geq \lambda_{\min} \left(\F_{\R}\brackets{\p}\right) - \eta(\p)>0 $, where the positivity is by the assumption of the Corollary . Hence $\hat{\F}_{\G}\brackets{\p} \succ 0 $.
It then follows that
\begin{align*}  
\|\hat{\F}_{\G}\brackets{\p}^{-1} \| &= \lambda_{\max} \left( \hat{\F}_{\G}\brackets{\p}^{-1}\right) = 1/\lambda_{\min} \left( \hat{\F}_{\G}\brackets{\p}\right) \\
 &\leq  \left[\lambda_{\min} (\F_{\R}\brackets{\p}) - \eta(\p) \right]^{-1}
 \end{align*}
which, establishes \eqref{eq:GCRB_bound}.
To establish \eqref{eq:GCRB_error_bound}, we have 
\begin{align*}
   & \| \F_{\R}\brackets{\p}^{-1} - \hat{\F}_{\G}\brackets{\p}^{-1} \| \\
    &= \|\F_{\R}\brackets{\p}^{-1} \left( \F_{\R}\brackets{\p}-\hat{\F}_{\G}\brackets{\p}\right) \hat{\F}_{\G}\brackets{\p}^{-1} \|  \\
    &\leq
\| \F_{\R}\brackets{\p}^{-1}\| \cdot \|\hat{\F}_{\G}\brackets{\p}^{-1} \| \cdot \|  \F_{\R}\brackets{\p}-\hat{\F}_{\G}\brackets{\p} \| 
\end{align*}
and the result follows by applying \eqref{eq:fim_error_bound} to the last factor in the product.
\end{proof}
}
\subsection{Proof of Theorem~\ref{thm:gcrb_sample_bound}}\label{apx:scb}
First we establish that the generated score vector has zero mean. 
\begin{lemma}\label{thm:score_vector_mean}
 Let $\vectorsym{s}_{\p}\brackets{\vectorsym{z}}$ be a score vector computed using a trimmed and differentiable $\G \in C^2$ generator $\G$ and it's inverse $\GI$.
	Then 
	  $ \expectation{\vectorsym{s}_{\p}\brackets{\Z}}{\Z}=0$  .
\end{lemma}
The proof of Lemma~\ref{thm:score_vector_mean} is given in Section~\ref{proofa:score_vector_mean} of the Appendix. Now we present a bound on the estimation of the precision matrix (inverse of a  covariance matrix).
\begin{theorem}\label{thm:inv_conv_bound_zero_mean} (Theorem 13 in \cite{kereta2021estimating}, specialized for $\expectation{\vectorsym{x}}{}=0$.)
	Let $\vectorsym{x}\in\mathbb{R}^{ D}$ be a 
	random vector with $\expectation{\vectorsym{x}}{}=0$ and covariance matrix $\Sigma=\expectation{\vectorsym{x}\vectorsym{x}^T}{}$. Assume $\norm{\matsym{A}\matsym{\Sigma}^{-1}\vectorsym{x}}_2\leq C_{A}$, $\norm{\matsym{B}\matsym{\Sigma}^{-1}\vectorsym{x}}_2\leq C_{B}$, $\norm{\sqrt{\matsym{\Sigma}^{-1}}{\vectorsym{x}}}_2\leq C_{x}$ almost surely, where $\matsym{A}\in\mathbb{R}^{d_1\times D},\matsym{B}\in\mathbb{R}^{d_2\times D}$ are known matrices. Let $\vectorsym{x}_1,..,\vectorsym{x}_m$ be a set of $m$ independent copies of $\vectorsym{x}$ with $\hat{\Sigma}=\frac{1}{m}\sum_{j=1}^m\vectorsym{x}_j\vectorsym{x}_j^T$   the finite sample estimator of $\Sigma$.
	Then there exist  absolute constants $C_1>0$ and $C_2>0$ such that provided $m>C_1\brackets{1+u}C_{x}^2$, we have  with probability at least $1-\exp\brackets{-u}$ for any $u>0$ that 
	\begin{equation*}
	    \norm{\matsym{A}\brackets{\hat{\Sigma}^{-1}-\Sigma^{-1}}\matsym{B}^T}_{F}\leq C_2 C_AC_B\sqrt{\frac{1+u}{m}}.
	\end{equation*}
\end{theorem}
The proof of Theorem~\ref{thm:inv_conv_bound_zero_mean} is given in Section~\ref{proofa:inv_zero} of the Appendix.

{\begin{proof} (of Theorem~\ref{thm:gcrb_sample_bound}
)
{
 By Lemma~\ref{lemma:bounded_score_vector}  $\norm{\vectorsym{s}_{\p}\brackets{\vectorsym{z}}}\leq \mathrm{C_s}\brackets{\p}$
 and} by Lemma~\ref{thm:score_vector_mean} we have $\expectation{\vectorsym{s}_{\p}\brackets{\vectorsym{z}}}{\vectorsym{z}}=0$. It follows that 
  Theorem \ref{thm:inv_conv_bound_zero_mean} is applicable to the score vector $\vectorsym{x}=\vectorsym{s}_{\p}\brackets{\vectorsym{z}}$ satisfying 
 $\expectation{\vectorsym{x}}{} =0$ and $\norm{\vectorsym{x}} \leq \mathrm{C_s}\brackets{\p}$. Making the identifications 
 $\Sigma=\hat{\F}_{\G}$, and $\hat{\Sigma} = \overline{\F_{\mathrm{G}}}\brackets{\p}$
    and setting $\matsym{A}=\matsym{B}=\matsym{I}$, and $C_{A}=C_{B}=\norm{\matsym{\Sigma}^{-1}}\mathrm{C_s}\brackets{\p}$ results in:
    \begin{equation*}
    \begin{split}
        \norm{\overline{\F_{\mathrm{G}}}\brackets{\p}^{-1}-\hat{\F}_{\G}\brackets{\p}^{-1}}_F\leq
        C_2\norm{\hat{\F}_{\G}\brackets{\p}^{-1}}^2 \mathrm{C_s}^2\brackets{\p}\sqrt{\frac{1+u}{m}}  .
    \end{split}
	\end{equation*}
\end{proof}}

 \subsection{Proof of Corollary~\ref{cor:total_error}
 } \label{proof:total_error}
{
\begin{proof}
\begin{align}\label{eq:bound_both}
    \mathrm{E}\brackets{\p}&\triangleq\norm{\overline{\F}_{\G}\brackets{\p}^{-1}-\F_{\R}\brackets{\p}^{-1}}\nonumber\\
    &\leq\norm{\overline{\F}_{\G}\brackets{\p}^{-1}-\hat{\F}_{\G}\brackets{\p}^{-1}}+\norm{\hat{\F}_{\G}\brackets{\p}^{-1}-\F_{\R}\brackets{\p}^{-1}}
    \nonumber \\
    & \leq \mathrm{B_s}\brackets{\p} + \norm{\F_{\R}\brackets{\p}^{-1}}\norm{\hat{\F}_{\G}\brackets{\p}^{-1}}\eta\brackets{\p} 
\end{align}
The first step follows by the triangle inequality, and the second by applying  Theorem~\ref{thm:gcrb_sample_bound} to the first term and upperbounding the spectral norm by the Frobenius norm, and applying  Corollary~\ref{cor:learning_error} to the second term
on the second line in \eqref{eq:bound_both}.
Finally, dividing \eqref{eq:bound_both} by $\norm{\F_{\R}\brackets{\p}^{-1}}$ yields the  Corollary.
 \end{proof}}

\bibliographystyle{IEEEtran}
\bibliography{ref}
\endgroup
\newpage  

\appendix 
\subsection{Score Vector Derivation}\label{apx:eq_prof}
\subsubsection{Score Vector Hybrid Version}\label{apx:score_vector_hybrid}
We show a detailed derivation of the score vector expressed in terms of both $\GI$ and $\G$. We start by splitting \eqref{eq:score_vector_hybrid} into
\begin{align}\label{eq:score_vector_base}
        \vectorsym{s}_{\p}\brackets{\vectorsym{z}}&=\at{\nabla_{\p}\log\probt{\GI\brackets{\vectorsym{\gamma};\p}}{\z}}{\vectorsym{\gamma}=\G\brackets{\z;\p}} \nonumber\\
        &+\at{\nabla_{\p}\log\abs{\mathrm{det}\matsym{J}_{\GI}\brackets{\vectorsym{\gamma}; \p}}}{\vectorsym{\gamma}=\G\brackets{\z;\p}}.
\end{align}
Applying the chain rule to the first term yields
\begin{equation} \label{apx:scorevec1}
    \nabla_{\p}\log\probt{\GI\brackets{\vectorsym{\gamma};\p}}{\z}=\divc{\GI\brackets{\vectorsym{\gamma};\p}}{\p}^{\mathrm{T}} \nabla_{\z}\log\prob{\z}.
\end{equation}
For the second term
\begin{align}
    &\divc{\log\abs{\mathrm{det}\matsym{J}_{\GI}\brackets{\vectorsym{\gamma}; \p}}}{\squareb{\p}_i}=\nonumber\\
    &\frac{1}{2}\divc{\log\mathrm{det}\left[\matsym{J}_{\GI}^T\brackets{\vectorsym{\gamma}; \p}\matsym{J}_{\GI}\brackets{\vectorsym{\gamma}; \p}\right]}{\squareb{\p}_i}=\nonumber\\
    &\trace{\matsym{J}_{\GI}^{-1}\brackets{\vectorsym{\gamma}; \p}\divc{\matsym{J}_{\GI}\brackets{\vectorsym{\gamma}; \p}}{\squareb{\p}_i}}, \label{apx:scorevec2}
\end{align}
where the second equality follows from the identity \cite{petersen2008matrix}
\begin{equation*}
    \frac{\partial \log \det B(t)}{\partial t} = \trace{B^{-1}(t)\frac{\partial B(t)}{\partial t}}
\end{equation*}
for positive-definite matrix $B(t) = A^T(t)A(t)$ where $A(t) \triangleq \matsym{J}_{\GI}^{-1}\brackets{\vectorsym{\gamma}; (\theta_1, \ldots, \theta_{i-1}, t, \theta_{i+1}, \theta_k)}$.
Substituting \eqref{apx:scorevec1} and  \eqref{apx:scorevec2} into \eqref{eq:score_vector_base} yields \eqref{eq:score_vector}.

\subsubsection{Score Vector Generator Version}\label{apx:generator_score_vector}
We derive an alternative expression for the score vector, in terms of  $\G$ only.  We begin with the defining identity
\begin{equation} \label{apx:GIG}
    \GI\brackets{\G\brackets{\z;\p};\p}=\z.
\end{equation}
Taking a derivative w.r.t $\p$ results in
\begin{align*}
    &0=\divc{\GI\brackets{\G\brackets{\z;\p};\p}}{\p}=\nonumber\\&\at{\divc{\GI\brackets{\g;\p}}{\p}}{\g=\G\brackets{\z;\p}}+\matsym{J}_{\GI}\brackets{\G\brackets{\z;\p}; \p}\divc{\G\brackets{\z;\p}}{\p}\nonumber\\
    &=\at{\divc{\GI\brackets{\g;\p}}{\p}}{\g=\G\brackets{\z;\p}}+\matsym{J}_{\G}^{-1}\brackets{\z; \p}\divc{\G\brackets{\z;\p}}{\p}.
\end{align*}
This provides the useful identity
\begin{equation}\label{eq:g_inv_t_g_comp}
    \at{\divc{\GI\brackets{\g;\p}}{\p}}{\g=\G\brackets{\z;\p}}=-\matsym{J}_{\G}^{-1}\brackets{\z; \p}\divc{\G\brackets{\z;\p}}{\p}.
\end{equation}
We use \eqref{eq:g_inv_t_g_comp}  directly to replace the first term in \eqref{eq:score_vector}, eliminating its dependence on $\GI$. 

Consider now the second term in \eqref{eq:score_vector_base}. 
\begin{align}\label{eq:vol_derivative_s}
    &\squareb{\vectorsym{w}\brackets{\z;\p}}_i \triangleq\divc{\log\abs{\mathrm{det}\matsym{J}_{\GI}\brackets{\vectorsym{\gamma}; \p}}}{\squareb{\p}_i}=\nonumber\\
    &\divc{\log\abs{\mathrm{det}\matsym{J}_{\G}\brackets{\GI\brackets{\g;\p}; \p}}^{-1}}{\squareb{\p}_i}=\nonumber\\
    &-\divc{\log\abs{\mathrm{det}\matsym{J}_{\G}\brackets{\GI\brackets{\g;\p}; \p}}}{\squareb{\p}_i}= \nonumber \\
    &-\trace{\matsym{J}_{\G}^{-1}\brackets{\GI\brackets{\g;\p}; \p}\divc{\matsym{J}_{\G}\brackets{\GI\brackets{\g;\p}; \p}}{\squareb{\p}_i}},
\end{align}
where the first equality follows directly from 
the inverse function theorem $\matsym{J}_{\GI}\brackets{\vectorsym{\gamma}; \p}=\matsym{J}_{\G}^{-1}\brackets{\GI\brackets{\g;\p}; \p}$
 and the third follows from \eqref{apx:scorevec2} upon replacing $\GI$ by $\G$.
The second factor under the trace in \eqref{eq:vol_derivative_s} is
\begin{align}\label{eq:jacob_div}
    \divc{\matsym{J}_{\G}\brackets{\GI\brackets{\g;\p}; \p}}{\squareb{\p}_i} 
    &=\at{\divc{\matsym{J}_{\G}\brackets{\vectorsym{\xi};\p}}{\squareb{\p}_i}+\matsym{M}_i\brackets{\vectorsym{\xi}}}{\vectorsym{\xi}=\GI\brackets{\g;\p}},
\end{align}
where the $kl$ element 
of $\matsym{M}_i$ is
\begin{align*} 
    \squareb{\matsym{M}_i\brackets{\vectorsym{\xi}}}_{kl}=\divc{\squareb{\matsym{J}_{\G}\brackets{\vectorsym{\xi};\p}}_{kl}}{\vectorsym{\xi}}^T\at{\divc{\GI\brackets{\g;\p}}{\squareb{\p}_i}}{\g=\G\brackets{\vectorsym{\xi};\p}}.
\end{align*}
Then using \eqref{eq:g_inv_t_g_comp} yields 
\begin{equation}\label{eq:m}
    \squareb{\matsym{M}_i\brackets{\vectorsym{\xi}}}_{kl}=-\divc{\squareb{\matsym{J}_{\G}\brackets{\vectorsym{\xi};\p}}_{kl}}{\vectorsym{\xi}}^T\matsym{J}_{\G}^{-1}\brackets{\vectorsym{\xi}; \p}\divc{\G\brackets{\vectorsym{\xi};\p}}{\p}.
\end{equation}
Next, we combine \eqref{eq:vol_derivative_s}, \eqref{eq:jacob_div} and \eqref{eq:m} to obtain 
\begin{align*}
    &\squareb{\vectorsym{w}\brackets{\z;\p}}_i \triangleq \at{\divc{\log\abs{\mathrm{det}\matsym{J}_{\GI}\brackets{\g; \p}}}{\squareb{\p}_i}}{\g=\G\brackets{\z;\p}}=\nonumber\\
    &\at{-\trace{\matsym{J}_{\G}^{-1}\brackets{\GI\brackets{\g;\p}; \p}\divc{\matsym{J}_{\G}\brackets{\GI\brackets{\g;\p}; \p}}{\squareb{\p}_i}}}{\g=\G\brackets{\z;\p}}\nonumber\\
    &=-\trace{\matsym{J}_{\G}^{-1}\brackets{\z; \p}\at{\divc{\matsym{J}_{\G}\brackets{\GI\brackets{\g;\p}; \p}}{\squareb{\p}_i}}{\g=\G\brackets{\z;\p}}}\nonumber\\
    &=-\trace{\matsym{J}_{\G}^{-1}\brackets{\z; \p}\brackets{\divc{\matsym{J}_{\G}\brackets{\z;\p}}{\squareb{\p}_i}+\matsym{M}_i\brackets{\z}}}.
\end{align*}
Finally, the score vector 
is given by
\begin{subequations}\label{eq:score_vector_generator_only}
\begin{align}
    \vectorsym{s}_{\p}\brackets{\vectorsym{z}}&=-\divc{\G\brackets{\z;\p}}{\p}^T\matsym{J}_{\G}^{-T}\brackets{\z; \p} \divc{\log\prob{\z}}{\z}\nonumber\\
    &-\vectorsym{w}\brackets{\z;\p}.
\end{align}
\begin{align}
    \squareb{\vectorsym{w}\brackets{\z;\p}}_i&=\trace{\matsym{J}_{\G}^{-1}\brackets{\z; \p}\divc{\matsym{J}_{\G}\brackets{\z;\p}}{\squareb{\p}_i}}\nonumber\\
    &+\trace{\matsym{J}_{\G}^{-1}\brackets{\z; \p}\matsym{M}_i\brackets{\z}}.
\end{align}
\begin{equation}
    \squareb{\matsym{M}_i\brackets{\z}}_{kl}=-\divc{\squareb{\matsym{J}_{\G}\brackets{\z;\p}}_{kl}}{\z}\matsym{J}_{\G}^{-1}\brackets{\z; \p}\divc{\G\brackets{\z;\p}}{\p}.
\end{equation}
\end{subequations}
Examining the various derivatives appearing in \eqref{eq:score_vector_generator_only}, it follows that for the generator score vector \eqref{eq:score_vector_generator_only} to be well-defined, it suffices to require that  $\G\in C^2$. Thanks to the defining relation \eqref{apx:GIG} between $\G$ and $\GI$, which induces the equivalence between \eqref{eq:score_vector_generator_only} and \eqref{eq:score_vector}, it follows that the same condition, $\G\in C^2$, suffices for \eqref{eq:score_vector} to be well-defined too.
\subsection{Flow Layers Overview}\label{apx:flow_layers}

\textbf{Affine Coupling} The affine coupling flow layer  \cite{dinh2016density}  is a 
powerful transformation that enables efficient computation of the forward function, the inverse function and the log-determinant. 
Let $\vectorsym{z}_{n}=(\vectorsym{z}_{n}^{A},\vectorsym{z}_{n}^{B})$ be a partition of the components of the vector $\vectorsym{z}_{n}$. Then the affine coupling layer  is defined as follows:
\begin{equation}\label{eq:coupling}
    \vectorsym{z}_{n+1}=\brackets{\exp\brackets{f^n_{\omega_n,s}\brackets{\vectorsym{z}_{n}^{B}}}\vectorsym{z}_{n}^{A}+f^n_{\omega_n,b}\brackets{\vectorsym{z}_{n}^{B}},\vectorsym{z}_{n}^{B}},
\end{equation}
where $f^n_{\omega_n,s}\brackets{\vectorsym{z}_{n}^{B}}$ and $f^n_{\omega_n,b}\brackets{\vectorsym{z}_{n}^{B}}$ are the scale and bias neural networks of the $n^{th}$ transformation, respectively, each parametrized by the learnable parameter vector $\omega_n$.
Note that the affine coupling layer applies an affine transformation to one of the disjoint blocks of $\z_n$, whereas the second block is  simply passed forward to the next flow layer. The inverse of \eqref{eq:coupling} is  given by
\begin{equation*}
    \vectorsym{z}_{n}^A=\frac{\vectorsym{z}_{n+1}^A-f^n_{\omega_n,b}\brackets{\vectorsym{z}_{n+1}^{B}}}{\exp\brackets{f^n_{\omega_n,s}\brackets{\vectorsym{z}_{n+1}^{B}}}}, \quad \vectorsym{z}_{n}^B=\vectorsym{z}_{n+1}^{B},
\end{equation*}
and the log-determinant term is $\sum_i \squareb{f^n_{\omega_n,s}\brackets{\vectorsym{z}_{n}^{B}}}_i$. 
Note that $f^n_{\omega_n,b}$ and $f^n_{\omega_n,s}$  can be an arbitrary complex neural networks, and the affine coupling layer is invertible thanks to its structure. 

\bigskip
\noindent
\textbf{Affine Inject} The Affine Injector conditional flow layer \cite{lugmayr2020srflow} enables direct information transfer from the conditioning parameter vector to the flow branch 
that directly affects the entire input vector $\z_n$.  This is achieved by controlling  the scaling and bias using only the  conditioning parameter vector $\p$:
\begin{equation}\label{eq:inject}
    \vectorsym{z}_{n+1}=\exp\brackets{f^n_{\omega_n,s}\brackets{\p}}\vectorsym{z}_{n}+f^n_{\omega_n,b}\brackets{\p},
\end{equation}
where $f^n_{\omega_n,b}$ and $f^n_{\omega_n,s}$ are the scale and bias neural networks of the $n^{th}$ transformation, respectively. The inverse and log-determinant of  \eqref{eq:inject} are  given by $\vectorsym{z}_{n}=\exp\brackets{-f^n_{\omega_n,s}\brackets{\p}}\brackets{\vectorsym{z}_{n+1}-f^n_{\omega_n,b}\brackets{\p}}$ and  $\sum_i f^n_{\omega_n,s}\brackets{\p}_i$. Note that $f^n_{\omega_n,b}$ and $f^n_{\omega_n,s}$  can be an arbitrary complex neural networks. 

\bigskip
\noindent
\textbf{Spline Flow}
The spline flow  \cite{durkan2019cubic} 
uses monotonic cubic 
splines to extend the affine coupling layer \cite{dinh2016density}. Specifically, a spline flow is defined as
\begin{equation*}
    \z_{n+1}^B=g\brackets{\z_{n}^B,f^n_{\omega_n}\brackets{\z_{n}^A}},\quad \z_{n+1}^A=\z_{n}^A,
\end{equation*}
where $\z_{n}=[\z_{n}^A,\z_{n}^B]$ 
is the splitting of vector $\z_n$ 
into two parts, $g$ is a  monotonic cubic 
spline and $f^n_{\omega_n}$ is an arbitrary neural network that generates the spline parameters. \\\\
\textbf{Invertible Matrix Product}
The  so-called $1\times 1$ convolution has been proposed as a flow layer \cite{NEURIPS2018_d139db6a}, with an LU decomposition of the corresponding matrix to reduce the cost of computing the log-determinant. The same can be done 
for matrix multiplication:
\begin{equation}\label{eq:mm}
    \vectorsym{z}_{n+1}=\matsym{W}\vectorsym{z}_{n},
\end{equation}
where $\matsym{W}=\matsym{P}\matsym{L}\brackets{\matsym{U}+\mathrm{diag}\brackets{\vectorsym{s}}}$, $\matsym{P}$ is a fixed permutation matrix, $\matsym{L}$, $\matsym{U}$ are trainable lower and upper triangular matrices with ones and zeros on the diagonal, respectively, and $\vectorsym{s}$ is a trainable vector with non-zero entries. 
The inverse and log-determinant term for the layer of \eqref{eq:mm} are  given by $\vectorsym{z}_{n}=\matsym{W}^{-1}\vectorsym{z}_{n+1}$ and  $\sum_i \vectorsym{s}_i$, respectively.

\bigskip
\noindent
\textbf{Activation Normalization} The activation normalization flow layer \cite{NEURIPS2018_d139db6a} perform an affine transformation of the channel using a scale and bias. The scale and bias are initialized using the statics of the first training batch such that each output of this flow step will have zero mean and unit variance given
an initial minibatch of data. Then the scale and bias are treated as regular trainable parameters.  Denoting the scale and bias vectors by $\vectorsym{s}$ and $\vectorsym{b}$, respectively, the  activation normalization is defined as
\begin{equation}\label{eq:actnorm}
    \vectorsym{z}_{n+1}=\vectorsym{s} \odot \vectorsym{z}_{n}+\vectorsym{b}.
\end{equation}
The inverse of \eqref{eq:actnorm} is  given by $\vectorsym{z}_{n}=\frac{1}{\vectorsym{s}}\odot\brackets{\z_{n+1}-\vectorsym{b}}$
and the log-determinant term is $\sum_i \squareb{\vectorsym{s}}_i$.
\subsection{Flow Parameters and Training Details}\label{apx:net}

\subsubsection{Flow Step}\label{apx:basic_block}

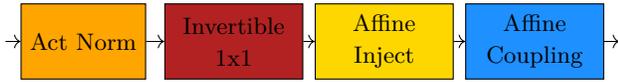
\begin{figure}
     \centering
         \centering
            \begin{tikzpicture}[node distance=2.0cm]
    \node (ry) {};
    
    \node[conv,minimum width=1.6cm ,right=0.2cm of ry] (ac)  {\centering{\small Act Norm}};
    \node[pool,text width=1.6cm,right of=ac,align=center] (inv)  {\centering{\small Invertible 1x1}};
    \node[bn,text width=1.6cm,right of=inv,align=center] (ai)  {\small Affine Inject};
    \node[fc,text width=1.6cm,right of=ai,align=center] (coupling)  {\small Affine Coupling};
    \node[right=0.2cm of coupling] (out) {};
    
    \draw[->] (ry) -- (ac);
    \draw[->] (ac) -- (inv);
    \draw[->] (inv) -- (ai);
    \draw[->] (ai) -- (coupling);
    \draw[->] (coupling) -- (out);
  \end{tikzpicture}
            \caption{Basic Flow Block
            }
        \label{fig:basic_flow_step}
   \end{figure}

We use a CNF based on the Glow\cite{NEURIPS2018_d139db6a} architecture (see Section \ref{sec:nf}). The architecture uses basic block (Fig.~\ref{fig:basic_flow_step}) that consists of the following consecutive layers: activation Normalization\cite{NEURIPS2018_d139db6a}, 1x1 convolution\cite{NEURIPS2018_d139db6a},  affine inject layer, and coupling layer\cite{dinh2016density, durkan2019cubic}. In both coupling and affine inject layers the parameter network generation is MLP (Section \ref{apx_mlp}) with different architectural choices for each problem.
\subsubsection{Multilayer Perceptron (MLP)}\label{apx_mlp}
\begin{figure}
         \centering
  \begin{tikzpicture}[node distance=2cm]
    \node (ry) {};
    \node[fc,text width=1.6cm,minimum width=1.5cm,right =0.2cm of ry,align=center] (fc1)  {\small $\mathrm{FC}(n_{i},n_{h})$ NL};
    \node[fc,text width=1.6cm,minimum width=1.5cm,right of=fc1,align=center] (fc2)  {\small $\mathrm{FC}(n_{h},n_{h})$  NL};

    \node[pool,text width=1.6cm,minimum width=1.5cm,right=1cm of fc2] (fc_n)  {\small $\mathrm{FC}(n_{h},n_{o})$};
    \node[right =0.2cm of fc_n] (out) {};
    \draw[->] (ry) -- (fc1);
    \draw[->] (fc1) -- (fc2);
    \draw[dashed] (fc2) -- (fc_n);
    \draw[->] (fc_n) -- (out);
  \end{tikzpicture}
  \caption{
  Multilayer Perceptron
  }
  \label{fig:mlp}
\end{figure}
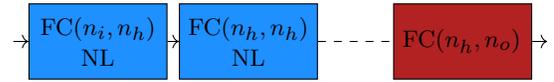

Several CNF layers (affine coupling, spline coupling, and affine inject) used throughout this work require a function to map the conditioning input to the layer parameters. 
We choose to implement these functions as a standard  Multilayer Perceptron (MLP) \cite{goodfellow2016deep} (Fig.~\ref{fig:mlp}), which is a sequence of fully connected layers with a non-linearity in between. Each fully connected layer consists of a matrix and a  bias vector, which are optimized during training.
The MLP  we use is defined by five architectural choices:  the input and output vector sizes $n_{o}$ and $n_{i}$, which are chosen equal and defined by problem dimensions and flow layer; the number of hidden neurons $n_h$;  the number of layers $n_{layers}$, which is chosen differently for each of our examples; and the non-linear function. For the latter we chose the Sigmoid Linear Unit (SiLU) activation function\cite{hendrycks2016gaussian}, allowing the resulting normalizing flow and generator to satisfy our differentiability assumptions.\footnote{We observed essentially identical results using a ReLu for the nonlinearity, which violates the differentiability assumptions at its "corner" at zero. The insensitivity of our scheme to this may be attributed to the fact that the ReLU is differentiable almost everywhere, so that sampling from continuous probability distributions, the likelihood of landing on the "corner" of any of the ReLUs in the network is zero.}
Note that the output layer of the MLP does not include a nonlinearity and that if $n_{layers}=1$, the MLP will be degenerate to a single fully connected layer.
\subsubsection{Linear Gaussian Flow}\label{apx_net_linear}
In the linear example, we use 
one flow block with affine coupling\cite{dinh2016density} (Appendix \ref{apx:basic_block}). The MLP network (Appendix \ref{apx_mlp}) used in the flow step consists of one layer 
($n_{layer}=1$). We use the normalizing flow above because it has sufficient expressive power for this example - in particular, it can represent an optimal generator for a linear Gaussian example.  Since  $\z\sim\normaldis{0}{\matsym{I}}$,
we need to transform  $\z$ to $\z_1\sim\normaldis{0}{\matsym{C}_{vv}}$, this is achievable using an invertible 1x1 convolution with weights  $W=L$.
Then,  the effect of the parameter vector $\p$ can be represented using the Affine Inject with single layer MLP 
$\z_{n+1}=\matsym{A}\p+\z_n$.  Note that we use a more complex
normalizing flow than the optimal  $\G\brackets{\vectorsym{z};\p}=\matsym{A}\vectorsym{\theta}+\matsym{L}\vectorsym{z}$, in the sense that we have additional layers (affine coupling and act norm) and parameters, which may require more data for training.
\subsubsection{Scale Non-Gaussian Flow}\label{apx_scale_linear}
In the non-Gaussian scale example, we use two flow blocks with a cubic spline coupling\cite{durkan2019cubic}
We utilized the cubic spline coupling for its ability to locally model non-linear functions such as $x^3$, where in this example it would need to model $x^{1/3}$. Then the affine inject layer can scale the transformation by $\theta$, which simulates locally the required transformations.   The MLP network (Appendix \ref{apx_mlp}) used in the flow step consists of five layers 
($n_{layer}=5$) and 64 hidden neurons ($n_h=64$).




\subsection{Auxiliary Proofs}\label{apx:aux_proof}
\subsubsection{Proof of Equation~\ref{eq:Relative_CRB_Error_Simple}}\label{proof:simple_re_crb}
{
\begin{proof}
{Consider matrices $\matsym{A}$, $\matsym{B}$, and $\matsym{\Delta} \triangleq \matsym{B}-\matsym{A} $, and assume that $\kappa(\matsym{A}) \frac{\|\matsym{\Delta}\|}{\|\matsym{A}\|} < 0.5$.
Then by \cite{horn2012matrix}(5.8.4)
\begin{equation*}
\frac{\|\matsym{B}^{-1} - \matsym{A}^{-1} \|}{\|\matsym{A}^{-1} \|}  \leq
\frac{\kappa(\matsym{A})}{1-\kappa(\matsym{A}) (\|\matsym{\Delta}\|/\|\matsym{A}\|)}
 \frac{\|\matsym{\Delta} \|}{\|\matsym{A}\|}  
\leq 2 \kappa(\matsym{A}) \frac{\|\matsym{\Delta} \|}{\|\matsym{A}\|} 
 \end{equation*}
 Now, by the triangle inequality
 \begin{align}
     \|\matsym{B}^{-1}\| &\leq \|\matsym{A}^{-1} \| + \|\matsym{B}^{-1} - \matsym{A}^{-1} \| \nonumber\\
    &\leq \|\matsym{A}^{-1}\|+ 2\kappa(A) \frac{\|\matsym{\Delta} \|}{\|\matsym{A}\|}{\|\matsym{A}^{-1}\|} \nonumber\\
     \frac{\|\matsym{B}^{-1}\|}{\|\matsym{A}^{-1}\|} & \leq 1 + 
    2\kappa(\matsym{A}) \frac{\|\matsym{\Delta} \|}{\|\matsym{A}\|}  \leq 2 \nonumber\\
        \left(\frac{\|\matsym{B}^{-1}\|}{\|\matsym{A}^{-1}\|}\right)^2 &\leq 4
        \implies \frac{\|\matsym{B}^{-1}\|^2}{\|\matsym{A}^{-1}\|}  \|\matsym{A}\| \leq 4 \kappa(\matsym{A}) 
    \label{eq:pert_Bound2}
\end{align}

Now setting $\matsym{A} \triangleq\F_{\R}\brackets{\p}$,  $\matsym{B}\triangleq \hat{\F}_{\G}\brackets{\p}$, and $\kappa(\matsym{A})=\kappa_{\R}\triangleq \kappa (\F_{\R}\brackets{\p}) = \kappa(\crb_{\R}\brackets{\p}) $, and
applying \eqref{eq:pert_Bound2} to the first term on the right hand side of \eqref{eq:relative_error} yields
\[
\norm{\hat{\F}_{\G}\brackets{\p}^{-1}}\tilde{\mathrm{B_s}} \leq 4\kappa_{\R} 
    C_2 \frac{C_{\R}^2\brackets{\p}}{\|\F_{\R}\brackets{\p}\|} \sqrt{\frac{1+u}{m}}
    \]
    The second term  on the right hand side of \eqref{eq:relative_error} is bounded by the right hand side of  \eqref{eq:GCRB_Rel_error}. Combining the results yields \eqref{eq:Relative_CRB_Error_Simple}.

}
\end{proof}
}
\subsubsection{Proof of Lemma~\ref{thm:score_vector_mean}}\label{proofa:score_vector_mean}
\begin{proof}
    By definition, 
    \begin{equation}
    \expectation{\vectorsym{s}_{\p}\brackets{\Z}}{\Z}=\expectation{\at{\divc{\log\probt{\Gamma;\p}{\Gamma}}{\p}}{\vectorsym{\Gamma}=\G\brackets{\Z;\p}}}{\Z}.    
    \end{equation}
    First using LOTUS  we replace the expected value w.r.t $\Z$ by that with respect to $\Gamma$.
    \begin{align*}
        \expectation{\vectorsym{s}_{\p}\brackets{\Z}}{\vectorsym{Z}}&=\expectation{\nabla_{\p}\log\probt{\Gamma;\p}{\Gamma}}{\vectorsym{\Gamma}}=\int_{\hat{\Upsilon}}\nabla_{\p}\probt{\g;\p}{\Gamma}d\g,\nonumber\\
        &=\nabla_{\p}\int_{\hat{\Upsilon}}\probt{\g;\p}{\Gamma}d\g=0.
    \end{align*}
    We take the derivative of the log in the second step and use the definition of expectation. In the last step, we use the linearity of integral and that $\hat{\Upsilon}$ doesn't depend on $\p$.
\end{proof}
\subsubsection{Proof of Theorem~\ref{thm:inv_conv_bound_zero_mean}}\label{proofa:inv_zero}
\begin{proof}(Sketch
)
  The proof of Theorem 13 in \cite{kereta2021estimating} starts with
  \begin{equation} \label{apx:original_thm}
      \norm{\matsym{A}_{\Sigma}\brackets{\hat{\Sigma}-\Sigma}\matsym{B}^T_{\Sigma}}\leq \norm{\matsym{A}_{\Sigma}\brackets{\tilde{\Sigma}-\Sigma}\matsym{B}^T_{\Sigma}}+\epsilon,
  \end{equation}
  where $\tilde{\Sigma}=\frac{1}{m}\sum_{j=1}^m\brackets{\vectorsym{x}_j-\expectation{\vectorsym{x}}{}}\brackets{\vectorsym{x}_j-\expectation{\vectorsym{x}}{}}^T$ and $\epsilon$ is some high order terms. With our definitions of $\hat{\Sigma}$ and $\expectation{\vectorsym{x}}{}=0$ we have that $\hat{\Sigma}= \tilde{\Sigma}$ and therefore \eqref{apx:original_thm} holds with equality for $\epsilon =0$.
 The rest of the proof follows \cite{kereta2021estimating}.
\end{proof}
\subsubsection{Proof of Lemma~\ref{lemma:mcsi}}\label{proofa:mcsi}
\begin{proof}
{\begin{align}
     0\preceq\matsym{W}&\triangleq\expectation{\brackets{\vectorsym{x}-\matsym{A}\vectorsym{y}}\brackets{\vectorsym{x}-\matsym{A}\vectorsym{y}}^T}{}\nonumber\\
     &=\matsym{R}_{\vectorsym{x}}
     +\matsym{A}\matsym{R}_{\vectorsym{y}}\matsym{A}^T
     -\matsym{A}\matsym{R}_{\vectorsym{x}\vectorsym{y}}^T
     -\matsym{R}_{\vectorsym{x}\vectorsym{y}}\matsym{A}^T.
 \end{align}
Let $\matsym{A}=
\matsym{R}_{\vectorsym{x}\vectorsym{y}}\brackets{\matsym{R}_{\vectorsym{y}}
+\alpha\matsym{I}_n}^{-1}$, with $\alpha>0$. Then
\begin{align}
    \matsym{W}&=\matsym{R}_{\vectorsym{x}}+
    \matsym{R}_{\vectorsym{x}\vectorsym{y}}
    \brackets{\matsym{R}_{\vectorsym{y}}+\alpha\matsym{I}_n}^{-1}
    \matsym{R}_{\vectorsym{y}}\brackets{\matsym{R}_{\vectorsym{y}}+\alpha\matsym{I}_n}^{-1}\matsym{R}_{\vectorsym{x}\vectorsym{y}}^T\nonumber\\
    &-2\matsym{R}_{\vectorsym{x}\vectorsym{y}}\brackets{\matsym{R}_{\vectorsym{y}}+\alpha\matsym{I}_n}^{-1}\matsym{R}_{\vectorsym{x}\vectorsym{y}}^T \label{eq:W_expand}
\end{align}
Now, we use a standard inequality that holds for any matrices $\matsym{B}, \matsym{C}, \matsym{D}$ with $\matsym{B}\preceq\matsym{C}$: $0\preceq\matsym{D}\brackets{\matsym{C}-\matsym{B}}\matsym{D}^T$, and the easily established fact that 
$\brackets{\matsym{R}_{\vectorsym{y}}+\alpha\matsym{I}_n}^{-1}
\matsym{R}_{\vectorsym{y}}
\brackets{\matsym{R}_{\vectorsym{y}}+\alpha\matsym{I}_n}^{-1}
\preceq\brackets{\matsym{R}_{\vectorsym{y}}+\alpha\matsym{I}_n}^{-1}$ 
to upper bound the second term in \eqref{eq:W_expand} by $\matsym{R}_{\vectorsym{x}\vectorsym{y}}\brackets{\matsym{R}_{\vectorsym{y}}+\alpha\matsym{I}_n}^{-1}\matsym{R}_{\vectorsym{x}\vectorsym{y}}^T$. Substituting into \eqref{eq:W_expand} yields
\begin{align}
    0\preceq\matsym{W}\preceq\matsym{R}_{\vectorsym{x}}-\matsym{R}_{\vectorsym{x}\vectorsym{y}}\brackets{\matsym{R}_{\vectorsym{y}}+\alpha\matsym{I}_n}^{-1}\matsym{R}_{\vectorsym{x}\vectorsym{y}}^T.
\end{align}
It follows that
\begin{align}\label{eq:lem_last_ineq}
    \norm{\matsym{R}_{\vectorsym{x}}}&\geq
    \norm{\matsym{R}_{\vectorsym{x}\vectorsym{y}}\brackets{\matsym{R}_{\vectorsym{y}}+\alpha\matsym{I}_n}^{-1}
    \matsym{R}_{\vectorsym{x}\vectorsym{y}}^T}\nonumber\\
    &\geq\lambda_{min}\brackets{\brackets{\matsym{R}_{\vectorsym{y}}+\alpha\matsym{I}_n}^{-1}}\norm{\matsym{R}_{\vectorsym{x}\vectorsym{y}}}^2\nonumber\\
    &=\brackets{\lambda_{max}\brackets{\brackets{\matsym{R}_{\vectorsym{y}}+\alpha\matsym{I}_n}}}^{-1}\norm{\matsym{R}_{\vectorsym{x}\vectorsym{y}}}^2\nonumber\\
    &=\norm{\matsym{R}_{\vectorsym{y}}+\alpha\matsym{I}_n}^{-1}\norm{\matsym{R}_{\vectorsym{x}\vectorsym{y}}}^2 \nonumber \\
    \norm{\matsym{R}_{\vectorsym{x}}} \norm{\matsym{R}_{\vectorsym{y}}+\alpha\matsym{I}_n} &\geq \norm{\matsym{R}_{\vectorsym{x}\vectorsym{y}}}^2
\end{align}
Finally, the inequality in \eqref{eq:lem_last_ineq} is true for any value $\alpha>0$, and by the continuity of the spectral norm it also holds in the limit of $\alpha\xrightarrow{}0$ which, upon taking the square root, results in Lemma \ref{lemma:mcsi}}
\end{proof}

\subsection{Linear Model Score Vector}\label{sec:linear_score}
To compute the score vector, we need the derivatives of the base distribution and of 
the normalizing flow:
\begin{align} 
    \divc{\log\prob{\z}}{\z}&=-\z \nonumber \\
     \at{\divc{\GI\brackets{\vectorsym{\gamma};\p}}{\p}^{\mathrm{T}}}{\vectorsym{\gamma}=\G\brackets{\z;\p}} &=\matsym{L}^{-1}\matsym{A} \label{eq:linear_flow} \\
      \matsym{J}_{\GI}^{-1}\brackets{\g;\p} &=\matsym{L} \nonumber
\end{align}
Then, substituting Equations \eqref{eq:linear_flow} into \eqref{eq:score_vector}, we compute
the score vector corresponding to the optimal normalizing flow in \eqref{eq:score_linear}.
\subsection{CRB for the Scale Non-Gaussian Example}\label{sec:vm_crm}
Let $r=\theta y$ then using transformation of variables $\probt{r}{R}=\probt{\frac{r}{\theta}}{y}\theta^{-1}$ which results in:
\begin{equation*}
    \probt{r}{R}=\frac{1}{\sqrt{2\pi\sigma^2}}\frac{3r^2}{\theta^3}\cdot\exp\brackets{-\frac{1}{2\sigma^2}
   \left( \frac{r}{\theta}\right)^6 }
\end{equation*}
Then the NNL function of $\probt{r}{R}$ is given by:
\begin{equation*}
    \mathrm{L}_{R}\brackets{\theta}=c+3\log\brackets{\theta}+\frac{1}{2\sigma^2}\left( \frac{r}{\theta}\right)^6 
\end{equation*}
where $c$ is some constant. Taking a derivative wrt $\theta$:
\begin{equation*}
    \frac{\partial \mathrm{L}_{R}\brackets{\theta}}{\partial\theta}
                =\frac{3}{\theta}\brackets{1-\frac{1}{\sigma^2} \left( \frac{r}{\theta}\right)^6 }
\end{equation*}
The FIM is then given by:
\begin{equation*}\label{eq:nnl_mul}
\begin{split}
    &\F_{\R}\brackets{\theta}=\expectation{\brackets{\frac{\partial\mathrm{L}_{R}\brackets{\theta}}{\partial\theta}}^2}{R}
             =\frac{9}{\theta^{2}}\expectation{\brackets{1-\frac{1}{\sigma^2}\left( \frac{R}{\theta}\right)^6 }^2}{R}\\
                    &=\frac{9}{\theta^{2}}\brackets{1-\frac{2}{\sigma^2}
                    \expectation{{\left( \frac{R}{\theta}\right)^6}}{R}+
                    \frac{1}{\sigma^4}
                    \expectation{\left( \frac{R}{\theta}\right)^{12}}{R}}
                    =18\theta^{-2}.
\end{split}
\end{equation*}
\subsubsection{Scale Model Optimal Generator Compare}\label{apx:scale_opt_gen}
We show that $\Gamma(\theta)=\G\brackets{z}=\theta z^{1/3}$ is an optimal generator that produces $\probt{r}{R}$: 
\begin{align*}
    \prob{\gamma}&=\probt{\brackets{\frac{\gamma}{\theta}}^3}{Z}3\frac{\gamma^2}{\theta}\nonumber\\
    &=\frac{1}{\sqrt{2\pi\sigma^2}}\frac{3\gamma^2}{\theta^3}\cdot\exp\brackets{-\frac{1}{2\sigma^2}\abs{\gamma}^{6}\theta^{-6}}=p_{R}(\gamma).
\end{align*}
where $\z\sim\normaldis{0}{1}$ and $\GI\brackets{\gamma}=\brackets{\frac{\gamma}{\theta}}^3$ is the inverse of the optimal generator. 
\subsubsection{Scale Model Score Vector}\label{sec:scale_score_vec}
To compute the score vector we need the derivative and Jacobian of the
normalizing flow:
\begin{align}\label{eq:scale_flow}
    \at{\divc{\GI\brackets{\gamma;\theta}}{\theta}^{\mathrm{T}}}{\gamma=\G\brackets{z;\theta}}&=-\frac{3z}{\theta}\nonumber\\
    \matsym{J}_{\GI}\brackets{\gamma;\theta}&=3\frac{\gamma^2}{\theta^3}\\
    \divc{\matsym{J}_{\GI}\brackets{\gamma;\theta}}{\theta}&=-9\frac{\gamma^2}{\theta^4}\nonumber\\
    \at{\matsym{J}_{\GI}^{-1}\brackets{\gamma;\theta}\divc{\matsym{J}_{\GI}\brackets{\gamma;\theta}}{\theta}}{\gamma=\G\brackets{z;\theta}}&=-\frac{3}{\theta}\nonumber
\end{align}
Then using \eqref{eq:scale_flow} we compute the score vector \eqref{eq:score_vector} of the optimal generator and normalizing flow in \eqref{eq:score_scale}.
\subsection{Edge Detection FIM and CRB}\label{app:crb_edge_detection}
%
%
In this section, we provide the FIM and CRB for the edge position estimation problem defined in Sec.~\ref{sec:edge_detection} with two noise models: WGN and NLF. 
\subsubsection{WGN}
Because $\matsym{V}_{ijc}\sim\normaldis{0}{\sigma^2}$ is {i.i.d }Gaussian Noise, the CRB for general Gaussian noise \cite{kay1993fundamentals} is applicable. The FIM is given by:
\begin{equation}\label{eq:fim_edge_gaus}
    {\matsym{F}\brackets{\p}=\frac{1}{\sigma^2}\sum_{i,j,c}\nabla_{\p} f_{ijc}\brackets{\p} \nabla_{\p} f_{ijc}\brackets{\p}^T},
\end{equation}
{\begin{align*}
    {\nabla_{\p} f_{ijc}\brackets{\p}=\frac{\brackets{p^h_c-p^l_c}}{\theta_w}s_{i}\brackets{\p}\brackets{1-s_{i}\brackets{\p}}\begin{bmatrix}
    1\\
    -\frac{\theta_p-i}{\theta_w}
    \end{bmatrix}}.
\end{align*}}
Substituting into \eqref{eq:fim_edge_gaus} yields
\begin{align*}
    \matsym{F}\brackets{\p}&=\sum_{i,j,c}\frac{\brackets{p^h_c-p^l_c}^2}{\sigma^2\theta_w^2}\matsym{M}_i\brackets{\p}=\frac{h\norm{\vectorsym{p}^h-\vectorsym{p}^l}^2_2}{\sigma^2\theta_w^2 }\sum_i \matsym{M}_i\brackets{\p},
\end{align*}
where 
    ${\matsym{M}_i\brackets{\p}=s_{i}^2\brackets{\p}\brackets{1-s_{i}\brackets{\p}}^2\begin{bmatrix}
    1& -\frac{\theta_p-i}{\theta_w}\\
     -\frac{\theta_p-i}{\theta_w} &  \frac{\brackets{\theta_p-i}^2}{\theta_w^2}
    \end{bmatrix},}$

which provides upon inversion the CRB {yields \eqref{eq:crb_edge_g}}.


\subsubsection{Noise Level Function}
Using the  noise model $\matsym{V}_{ijc}\sim\normaldis{0}{\alpha^2f_{ijc}\brackets{\theta_p}+\delta^2}$, the CRB formula for general Gaussian noise {with independent samples} is again applicable with  FIM

\begin{equation}\label{eq:nfl_edge_fim_base}
\begin{split}
    \matsym{F}\brackets{\p}&=\sum_{i,j,c}\frac{1}{C_{ijc}\brackets{\p}}\nabla_{\p} f_{ijc}\brackets{\p}\nabla_{\p} f_{ijc}\brackets{\p}^T\\
    &+\frac{1}{2 C_{ijc}\brackets{\p}^2}\nabla_{\p} C_{ijc}\brackets{\p}\nabla_{\p} C_{ijc}\brackets{\p}^T, \\
\end{split}
\end{equation}

where  {$C_{ijc}\brackets{\p}=\alpha^2f_{ijc}\brackets{\p}+\delta^2$}. 
The first term of \eqref{eq:nfl_edge_fim_base} is
{\begin{align}
    \matsym{T}_1&\triangleq\sum_{i,j,c}\frac{1}{C_{ijc}\brackets{\p}}\nabla_{\p} f_{ijc}\brackets{\p}\nabla_{\p} f_{ijc}\brackets{\p}^T\nonumber\\
    &=\sum_{i,j,c}\frac{\brackets{p^h_c-p^l_c}^2}{\theta_w^2 C_{ijc}\brackets{\p}}\matsym{M}_i\brackets{\p}
\end{align}}
The second term is given by:
{color{blue}\begin{align}
    \matsym{T}_2 &\triangleq \sum_{i,j,c}\frac{1}{2 C_{ijc}\brackets{\p}^2}\nabla_{\p} C_{ijc}\brackets{\p}\nabla_{\p} C_{ijc}\brackets{\p}^T,\nonumber\\
    &=\sum_{i,j,c}\frac{\alpha^2}{2 C_{ijc}\brackets{\p}^2}\frac{\brackets{p^h_c-p^l_c}^2}{\theta_w^2}\matsym{M}_i\brackets{\p}
\end{align}}
Combining $T_1$ and $T_2$ yields
{\begin{align*}
    \matsym{F}\brackets{\p}&=\sum_{i,j,c}\frac{\brackets{p^h_c-p^l_c}^2\matsym{M}_i\brackets{\p}}{\brackets{\alpha^2f_{ijc}\brackets{\p}+\delta^2}^2\theta_w^2}\brackets{\alpha^2f_{ijc}\brackets{\p}+\delta^2+\frac{\alpha^2}{2}},
\end{align*}}
yielding upon inversion the CRB  \eqref{eq:crb_edge_nlf}.


\end{document}